\documentclass[journal]{IEEEtran}
\usepackage{lscape}
\usepackage{times}
\usepackage{graphicx} 
\usepackage[margin=1in]{geometry}
\usepackage{subfigure} 
\usepackage{algpseudocode}
\usepackage{algorithm}
\usepackage{hyperref}

\newcommand{\rec}{\mathrm{rec}}
\newcommand{\nn}{\nonumber}
\usepackage{url}
\usepackage{graphics}
\usepackage{amsmath}
\usepackage{color}
\usepackage{fancyhdr}
\usepackage{datetime}
\usepackage{amssymb}
\usepackage{amsthm}
\graphicspath{{./Figures/}}
\begin{document}
\title{Distributed Compressive Sensing: A Deep Learning Approach}


\author{Hamid~Palangi, Rabab~Ward, Li~Deng
\thanks{Copyright (c) 2015 IEEE. Personal use of this material is permitted. However, permission to use this material for any other purposes must be obtained from the IEEE by sending a request to pubs-permissions@ieee.org.}
\thanks{H. Palangi and R. Ward  are with the Department
of Electrical and Computer Engineering, University of British Columbia, Vancouver,
BC, V6T 1Z4 Canada (e-mail: \{hamidp,rababw\}@ece.ubc.ca)}
\thanks{L. Deng is with Microsoft Research, Redmond, WA 98052 USA (e-mail: \{deng\}@microsoft.com)}}

\maketitle
\begin{abstract}
Various studies that address the compressed sensing problem with Multiple Measurement Vectors (MMVs) have been recently carried. These studies assume the vectors of the different channels to be jointly sparse. In this paper, we relax this condition. Instead we assume that these  sparse vectors depend on each other but that this dependency is unknown. We capture this dependency by computing the conditional probability of each entry in each vector being non-zero, given  the ``\textit{residuals}'' of all previous vectors. To estimate these probabilities, we propose the use of the Long Short-Term Memory (LSTM) \cite{lstm}, a data driven model for sequence modelling that is deep in time. To calculate the model parameters, we minimize a cross entropy cost function. To reconstruct the sparse vectors at the decoder, we propose a greedy solver that uses the above model to estimate the conditional probabilities. By performing extensive experiments on two real world datasets, we show that the proposed method significantly outperforms the general MMV solver (the Simultaneous Orthogonal Matching Pursuit (SOMP)) and  a number of the model-based Bayesian methods. The proposed method does not add any complexity to the general compressive sensing encoder. The trained model is used just at the decoder. As the proposed method is a data driven method, it is only applicable when training data is available. In many applications however, training data is indeed available, e.g. in recorded images and videos. 
\end{abstract}
\IEEEpeerreviewmaketitle

\section{Introduction}
\label{sec:intro}
\IEEEPARstart{C}{ompressive} Sensing (CS) \cite{CSdonoho},\cite{Candes},\cite{CSrich} is an effective approach for acquiring sparse signals where both sensing and compression are performed at the same time. Since there are numerous examples of natural and artificial signals that are sparse in the time, spatial or a transform domain, CS has found numerous applications. These include medical imaging, geophysical data analysis, computational biology, remote sensing and communications.

In the general CS framework, instead of acquiring $N$ samples of a signal $\mathbf{x}\in\Re^{N\times 1}$, $M$ random measurements are acquired where $M<N$. This is expressed by the underdetermined system of linear equations: 
\begin{equation}
\label{eq:Intro1}
\mathbf{y}=\mathbf{\Phi} \mathbf{x}
\end{equation}
where $\mathbf{y}\in\Re^{M\times 1}$ is the known measured vector and $\mathbf{\Phi} \in\Re^{M\times N}$ is a random measurement matrix. To uniquely recover $\mathbf{x}$ given $\mathbf{y}$ and $\mathbf{\Phi}$, $\mathbf{x}$ must be sparse in a given basis $\mathbf{\Psi}$. This means that
\begin{equation}
\label{eq:Intro2}
\mathbf{x}=\mathbf{\Psi} \mathbf{s}
\end{equation}
where $\mathbf{s}$ is $K-sparse$, i.e., $\mathbf{s}$ has at most $K$ non-zero elements. The basis $\mathbf{\Psi}$ can be complete; i.e., $\mathbf{\Psi}\in\Re^{N\times N}$, or over-complete; i.e., $\mathbf{\Psi}\in\Re^{N\times N_{1}}$ where $N<N_{1}$ (compressed sensing for over-complete dictionaries is introduced in \cite{OvercompleteCS}). From \eqref{eq:Intro1} and \eqref{eq:Intro2}:
\begin{equation}
\label{eq:Intro3}
\mathbf{y}=\mathbf{A} \mathbf{s}
\end{equation}
where $\mathbf{A} = \mathbf{\Phi\Psi}$. Since there is only one measurement vector, the above problem is usually called the Single Measurement Vector (SMV) problem in compressive sensing. 

In distributed compressive sensing , also known as the Multiple Measurement Vectors (MMV) problem, a set of $L$ sparse vectors $\lbrace{\mathbf{s}_i}\rbrace_{i=1,2,\dots,L}$ is to be jointly recovered from a set of $L$ measurement vectors $\lbrace{\mathbf{y}_i}\rbrace_{i=1,2,\dots,L}$. Some application areas of MMV include magnetoencephalography, array processing, equalization of sparse communication channels and cognitive radio \cite{Eldar_SCS}. 

Suppose that the $L$ sparse vectors and the $L$ measurement vectors are arranged as columns of matrices $\mathbf{S}=[\mathbf{s}_{1},\mathbf{s}_{2},\dots,\mathbf{s}_{L}]$ and $\mathbf{Y}=[\mathbf{y}_{1},\mathbf{y}_{2},\dots, \mathbf{y}_{L}]$ respectively. In the MMV problem, $\mathbf{S}$ is to be reconstructed given $\mathbf{Y}$:
\begin{equation}
\label{eq:Intro4}
\mathbf{Y}=\mathbf{A} \mathbf{S}
\end{equation} 
In \eqref{eq:Intro4}, $\mathbf{S}$ is assumed to be jointly sparse, i.e., non-zero entries of each vector occur at the same locations as those of other vectors, which means that the sparse vectors have the same support. Assume that $\mathbf{S}$ is jointly sparse. Then, the necessary and sufficient condition to obtain a unique $\mathbf{S}$ given $\mathbf{Y}$ is \cite{RankAware}: 
\begin{equation}
\label{eqTheoremMMV}
\vert supp(\mathbf{S})\vert<\frac{spark(\mathbf{A})-1+rank(\mathbf{S})}{2}
\end{equation}
where $\vert supp(\mathbf{S})\vert$ is the number of rows in $\mathbf{S}$ with non-zero energy and $spark$ of a given matrix is the smallest possible number of linearly dependent columns of that matrix. $spark$ gives a measure of linear dependency in the system modelled by a given matrix. In the SMV problem, no rank information exists. In the MMV problem, the rank information exists and affects the uniqueness bounds. 
Generally, solving the MMV problem jointly can lead to better uniqueness guarantees than solving the SMV problem for each vector independently \cite{AverageCaseAnalysisMMV}. 

In the current MMV literature, a jointly sparse matrix is recovered typically by one of the following methods: 1) greedy methods \cite{Tropp1} like Simultaneous Orthogonal Matching Pursuit (SOMP) which performs non-optimal subset selection, 2) relaxed mixed norm minimization methods \cite{Tropp2}, or 3) Bayesian methods like \cite{Rao1,Carin2,Rao3} where a posterior density function for the values of $\mathbf{S}$ is created, assuming a prior belief, e.g., $\mathbf{Y}$ is observed and $\mathbf{S}$ should be sparse in basis $\mathbf{\Psi}$. The selection of one of the above methods depends on the requirements imposed by the specific application.

\subsection{Problem Statement}
\label{intro:problem}
The MMV reconstruction methods stated above do not rely on the use of training data. However, for many applications, a large amount of data similar to the data to be compressed by CS is available. Examples are camera recordings of the same environment, images of the same class (e.g., flowers, buildings, ....), electroencephalogram (EEG) of different parts of the brain, etc. In this paper, we address the following questions in the MMV problem when training data is available:
\begin{enumerate}
\item Can we learn the structure of the sparse vectors in $\mathbf{S}$ by a data driven bottom up approach using the already available training data? If yes, then how can we exploit this structure in the MMV problem to design a better reconstruction method? 
\item Most of the reconstruction algorithms for the MMV problem rely on the joint sparsity of $\mathbf{S}$. However, in some practical applications, the sparse vectors in $\mathbf{S}$ are not exactly jointly sparse. This can be due to noise or due to sources that create different sparsity patterns. Examples are images of different scenes captured by different cameras, images of different classes, etc. Although $\mathbf{S}$ is not jointly sparse, there may exist a possible dependency among the columns of $\mathbf{S}$, however, due to lack of joint sparsity, the above methods will not give satisfactory performance. The question is, can we design the aforementioned data driven method in a way that it captures the dependencies among the sparse vectors in $\mathbf{S}$? The type of such dependencies may not be necessarily that of joint sparsity. And then how can we use the learned dependency structure in the reconstruction algorithm at the decoder? 
\end{enumerate}

Please note that we want to address the above questions ``\textit{without adding any complexity or adaptability}'' to the encoder. In other words, our aim is not to design an optimal encoder, i.e., optimal sensing matrix $\mathbf{\Phi}$ or the sparsifying basis $\mathbf{\Psi}$, for the given training data. The encoder would be as simple and general as possible. This is specially important for applications that use sensors having low power consumption due to a limited battery life. However, the decoder in these cases can be much more complex than the encoder. For example, the decoder can be a powerful data processing machine.
\subsection{Proposed Method}
\label{intro:method}
To address the above questions, we propose the use of a two step greedy reconstruction algorithm. In the first step, at each iteration of the reconstruction algorithm, and for each column of $\mathbf{S}$ represented as $\mathbf{s}_i$, we first find the conditional probability of each entry of $\mathbf{s}_i$ being non-zero, given the residuals of all previous sparse vectors (columns) at that iteration.  Then we select the most probable entry and add it to the support of $\mathbf{s}_i$. The definition of the residual matrix at the $j-$th iteration is $\mathbf{R}_j = \mathbf{Y} - \mathbf{A}\mathbf{S}_j$ where $\mathbf{S}_j$ is the estimate of the sparse matrix $\mathbf{S}$ at the $j-$th iteration. Therefore in the first step, we find the locations of the non-zero entries. In the second step we find the values of these non-zero entries. This can be done by solving a least squares problem that finds $\mathbf{s}_i$ given $\mathbf{y}_i$ and $\mathbf{A}_{\Omega_i}$. $\mathbf{A}_{\Omega_i}$ is a matrix that includes only those atoms (columns) of $\mathbf{A}$ that are members of the support of $\mathbf{s}_i$. 

\begin{figure*}[t]
\center
\includegraphics[width=.8\textwidth]{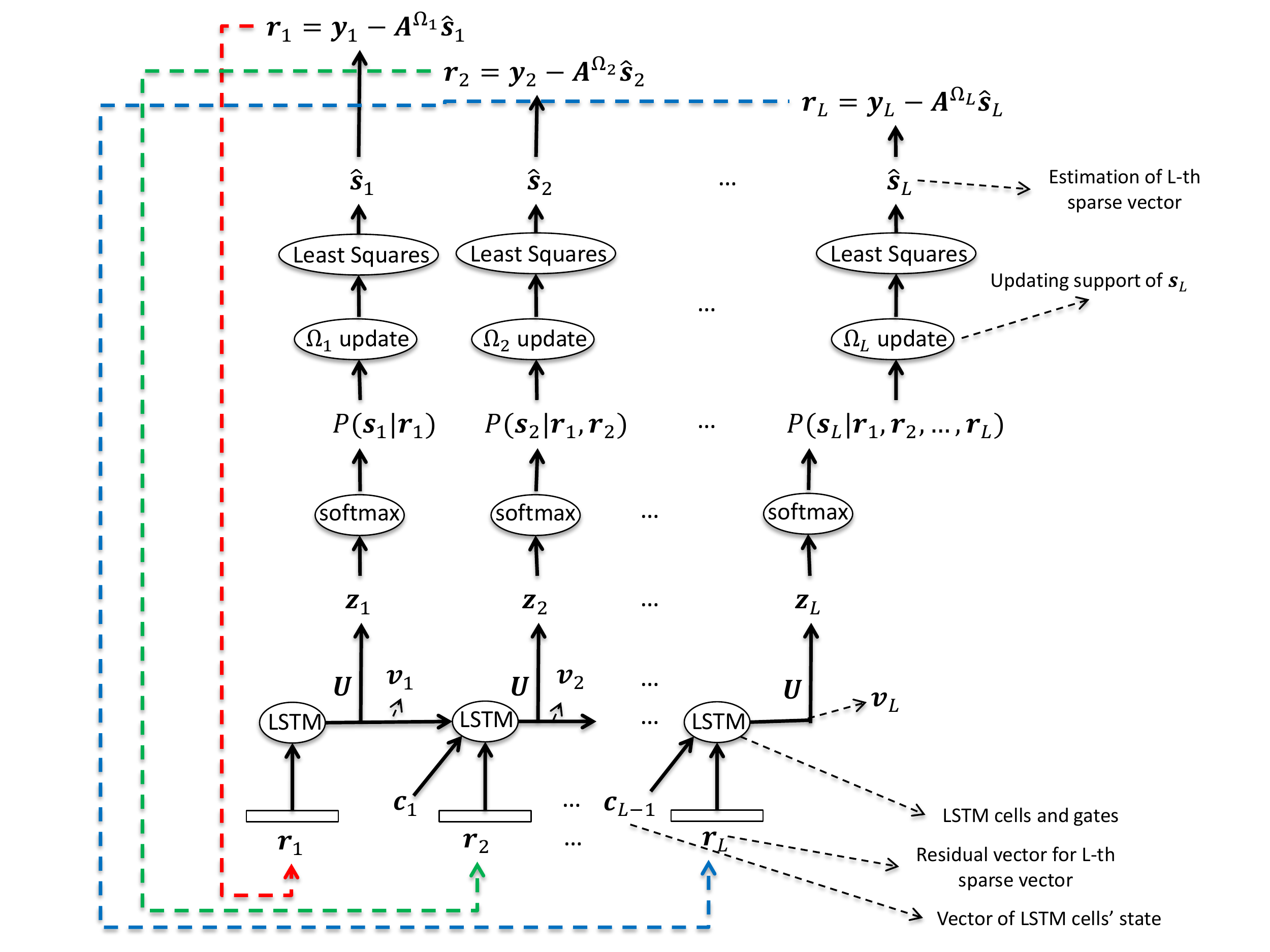}
\caption{Block diagram of the proposed method unfolded over channels.}
\label{Fig:BlockDiagram}
\end{figure*}

\begin{figure}[t]
\center
\includegraphics[width=0.6\textwidth]{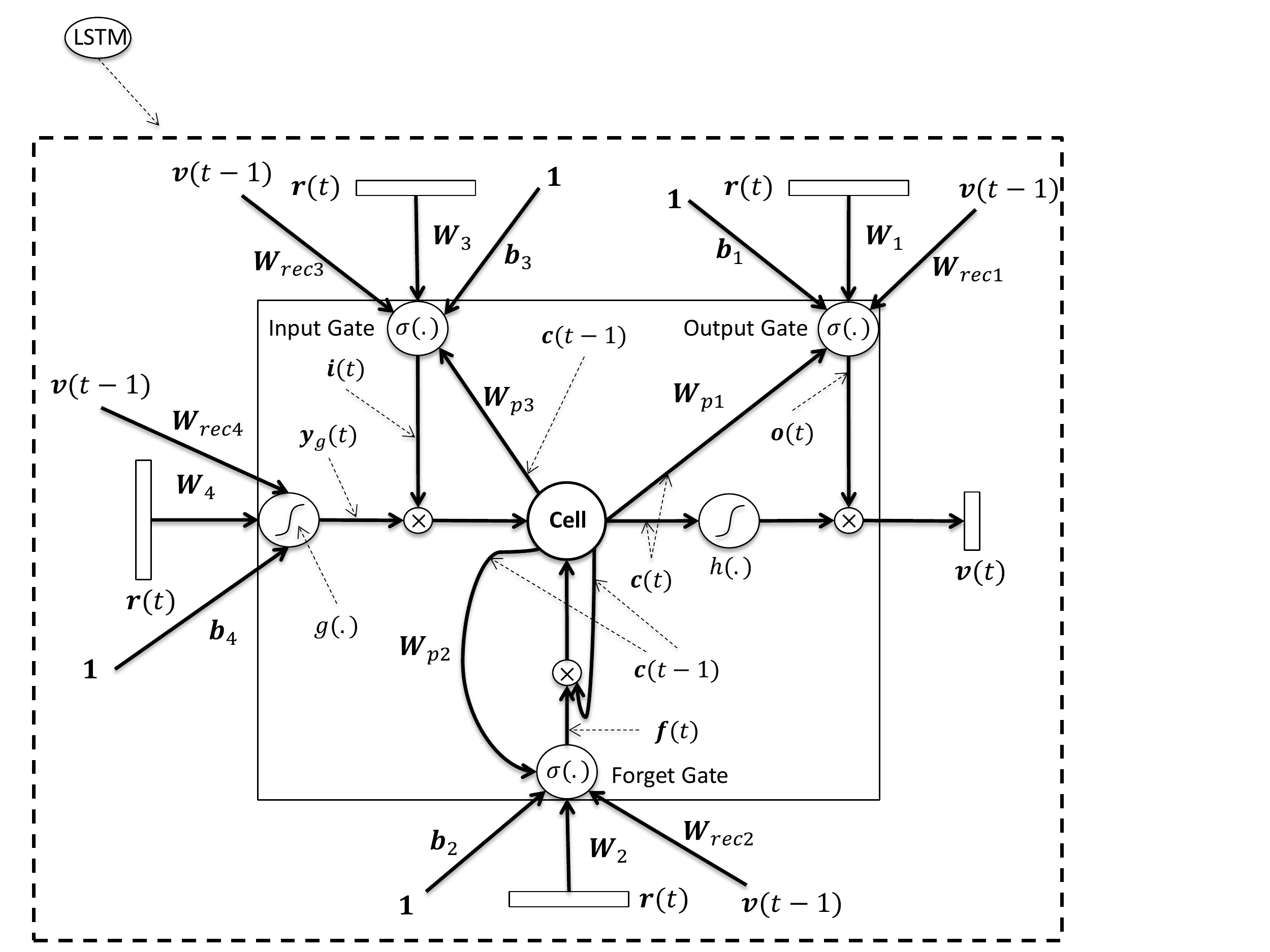}
\caption{Block diagram of the Long Short-Term Memory (LSTM).}
\label{Fig:LSTM}
\end{figure}

To find the conditional probabilities at each iteration, we propose the use of a Recurrent Neural Network (RNN) with Long Short-Term Memory (LSTM) cells and a softmax layer on top of it. To find the model parameters, we minimize a cross entropy cost function between the conditional probabilities given by the model and the known probabilities in the training data. The details on how to generate the training data and the training data probabilities are explained in subsequent sections. Please note that this training is done only once. After that, the resulting model is used in the reconstruction algorithm for any test data that has not been observed by the model before. Therefore, the proposed reconstruction algorithm would be almost as fast as the greedy methods. The block diagram of the proposed method is presented in Fig. \ref{Fig:BlockDiagram} and Fig. \ref{Fig:LSTM}. We will explain these figures in detail in subsequent sections.

To the best of our knowledge, this is the first model-based method in MMV sparse reconstruction that is based on a deep learning bottom up approach. Similar to all deep learning methods, it has the important feature of learning the structure of $\mathbf{S}$ from the raw data automatically. Although it is based on a greedy method that selects subsets that are not necessarily optimal, we experimentally show that by using a properly trained model and only one layer of LSTM, the proposed method significantly outperforms well known MMV baselines (e.g., SOMP) as well as the well known Bayesian methods for the MMV problem (e.g., Multitask Bayesian Compressive Sensing (MT-BCS)\cite{Carin2} and Sparse Bayesian Learning for temporally correlated sources (T-SBL)\cite{Rao3}). We show this on two real world datasets.

We emphasize that the computations carried at the encoder mainly include multiplication by a random matrix. The extra computations are only needed at the decoder. Therefore an important feature of compressive sensing (low power encoding) is preserved. 

\subsection{Related Work}
\label{intro:RelatedWork}
Exploiting data structures besides sparsity for compressive sensing has been extensively studied in the literature \cite{ModelBasedCS,Eldar_SCS,Carin1,Carin2,Rao1,Rao3,NNCS,hpICASSP2013,Baraniuk_SDA_CS,LSCS,AfterRev2,AfterRev3}. In \cite{ModelBasedCS}, it has been theoretically shown that using signal models that exploit these structures will result in a decrease in the number of measurements. In \cite{Eldar_SCS}, a thorough review on CS methods that exploit the structure present in the sparse signal or in the measurements is presented. In \cite{Carin1}, a Bayesian framework for CS is presented. This framework uses a prior information about the sparsity of $\mathbf{s}$ to provide a posterior density function for the entries of $\mathbf{s}$ (assuming $\mathbf{y}$ is observed). It then uses a Relevance Vector Machine (RVM) \cite{rvm} to estimate the entries of the sparse vector. This method is called Bayesian Compressive Sensing (BCS). In \cite{Carin2}, a Bayesian framework is presented for the MMV problem. It assumes that the $L$ ``tasks'' in the MMV problem in \eqref{eq:Intro4}, are not statistically independent. By imposing a shared prior on the $L$ tasks, an empirical method is presented to estimate the hyperparameters and extensions of RVM are used for the inference step. This method is known as Multitask Compressive Sensing (MT-BCS). In \cite{Carin2}, it is experimentally shown that the MT-BCS outperforms the method that applies Orthogonal Matching Pursuit (OMP) on each task, the Simultaneous Orthogonal Matching Pursuit (SOMP) method which is a straightforward extension of OMP for the MMV problem, and the method that applies BCS for each task. In \cite{Rao1}, the Sparse Bayesian Learning (SBL) \cite{rvm,rvm2} is used to solve the MMV problem. It was shown that the global minimum of the proposed method is always the sparsest one. The authors in \cite{Rao3}, address the MMV problem when the entries in each row of $\mathbf{S}$ are correlated. An algorithm based on SBL is proposed and it is shown that the proposed algorithm outperforms the mixed norm ($\ell_{1,2}$) optimization as well as the method proposed in \cite{Rao1}. The proposed method is called T-SBL. In \cite{NNCS}, a greedy algorithm aided by a neural network is proposed to address the SMV problem in \eqref{eq:Intro3}. The neural network parameters are calculated by solving a regression problem and are used to select the appropriate column of $\mathbf{A}$ at each iteration of OMP. The main modification to OMP is replacing the correlation step with a neural network. They experimentally show that the proposed method outperforms OMP and $\ell_1$ optimization. This method is called Neural Network OMP (NNOMP). In \cite{hpICASSP2013}, an extension of \cite{NNCS}  with a hierarchical Deep Stacking Netowork (DSN) \cite{DSN} is proposed for the MMV problem. ``\textit{The joint sparsity of $\mathbf{S}$ is an important assumption in the proposed method}''. To train the DSN model, the Restricted Boltzmann Machine (RBM) \cite{rbm3} is used to pre-train DSN and then fine tuning is performed. It has been experimentally shown that this method outperforms SOMP and $\ell_{1,2}$ in the MMV problem. The proposed methods are called Nonlinear Weighted SOMP (NWSOMP) for the one layer model and DSN-WSOMP for the multilayer model. In \cite{Baraniuk_SDA_CS}, a feedforward neural network is used to solve the SMV problem as a regression task. Similar to \cite{hpICASSP2013} (if we assume that we have only one sparse vector in \cite{hpICASSP2013}), a pre-training phase followed by a fine tuning is used. For pre-training, the authors have used Stacked Denoising Auto-encoder (SDA) \cite{SDA}. Please note that an RBM with Gaussian visible units and binary hidden units (i.e., the one used in \cite{hpICASSP2013}) has the same energy function as an auto-encoder with sigmoid hidden units and real valued observations \cite{SDA2}. Therefore the extension of \cite{Baraniuk_SDA_CS} to the MMV problem will give similar performance as that of \cite{hpICASSP2013}. In \cite{LSCS}, a reconstruction method is proposed for sparse signals whose sparsity patterns change slowly with time. The main idea is to replace Compressive Sensing (CS) on the observation $\mathbf{y}$ with CS on the Least Squares (LS) residuals. LS residuals are calculated using the previous estimation of the support. In \cite{AfterRev2}, a reconstruction method is proposed to recover sparse signals with a sparsity pattern that slowly changes over time. The main idea is to use Sparse Bayesian Learning (SBL) framework. Similar to SBL, a set of hyperparameters are defined to control the sparsity of signals. The main difference is that the prior for each coefficient also involves the coefficients of the adjacent temporal observations. In \cite{AfterRev3}, a CS algorithm is proposed for time-varying sparse signals based on the least-absolute shrinkage and selection operator (Lasso). A dynamic Lasso algorithm is proposed for the signals with time-varying amplitudes and support.

The rest of the paper is organized as follows: In section \ref{sec:LSTM}, the basics of Recurrent Neural Networks (RNN) with Long Short-Term Memory (LSTM) cells are briefly explained. The proposed method and the learning algorithm are presented in section \ref{sec:ProposedMethod}. Experimental results on two real world datasets are presented in section \ref{sec:Experiments}. Conclusions and future work directions are discussed in section \ref{sec:conclusion}. Details of the final gradient expressions for the learning section of the proposed method are presented in Appendix \ref{sec:appGradientExpress}. 

\section{RNN with LSTM cells}
\label{sec:LSTM}
The RNN is a type of deep neural networks \cite{hinton2012deep,Dahl} that are ``deep'' in the temporal dimension. It has been used extensively in time sequence modelling \cite{RNNref,robinson1994application,deng1994analysis,mikolov2010recurrent,graves2012sequence,
bengio2013advances,mesnilinvestigation,hpRDSN,hpESN}. If we look at the sparse vectors (columns) in $\mathbf{S}$ as a sequence, the main idea of using RNN for the MMV problem is to predict the sparsity patterns over different sparse vectors in $\mathbf{S}$. 

Although RNN performs sequence modelling in a principled manner, it is generally difficult to learn the long term dependency within the sequence due to the vanishing gradients problem. One of the effective solutions for this problem in RNNs is to employ memory cells instead of neurons that is originally proposed in \cite{lstm} as Long Short-Term Memory (LSTM). It is further developed in \cite{lstm_forget} and \cite{lstm_peephole} by adding forget gate and peephole connections to the architecture. 

We use the architecture of LSTM illustrated in Fig. \ref{Fig:LSTM} for the proposed sequence modelling method for the MMV problem. In this figure, $\mathbf{i}(t),\;\mathbf{f}(t)\;,\mathbf{o}(t)\;, \mathbf{c}(t)$ are input gate, forget gate, output gate and cell state vector respectively, $\mathbf{W}_{p1},\;\mathbf{W}_{p2}$ and $\mathbf{W}_{p3}$ are peephole connections, $\mathbf{W}_i$, $\mathbf{W}_{reci}$ and  $\mathbf{b}_i$, $i=1,2,3,4$ are input connections, recurrent connections and bias values, respectively, $g(\cdot)$ and $h(\cdot)$ are $\tanh(\cdot)$ function and $\sigma(\cdot)$ is the sigmoid function. We use this architecture to find $\mathbf{v}$ for each channel and then use the proposed method in Fig. \ref{Fig:BlockDiagram} to find the entries that have a higher probability of being non-zero. 
Considering Fig.~\ref{Fig:LSTM}, the forward pass for LSTM model is as follows:
\begin{align}
\label{eq:lstm_forward}
&\mathbf{y}_g(t) = g(\mathbf{W}_4\mathbf{r}(t) + \mathbf{W}_{\rec 4}\mathbf{v}(t-1) + \mathbf{b}_4)\nn\\
&\mathbf{i}(t) = \sigma(\mathbf{W}_3\mathbf{r}(t) + \mathbf{W}_{\rec 3}\mathbf{v}(t-1) + \mathbf{W}_{p3}\mathbf{c}(t-1) + \mathbf{b}_3)\nn\\
&\mathbf{f}(t) = \sigma(\mathbf{W}_2\mathbf{r}(t) + \mathbf{W}_{\rec 2}\mathbf{v}(t-1) + \mathbf{W}_{p2}\mathbf{c}(t-1) + \mathbf{b}_2)\nn\\
&\mathbf{c}(t) = \mathbf{f}(t)\circ\mathbf{c}(t-1) + \mathbf{i}(t)\circ\mathbf{y}_g(t)\nn\\
&\mathbf{o}(t) = \sigma(\mathbf{W}_1\mathbf{r}(t) + \mathbf{W}_{\rec 1}\mathbf{v}(t-1) + \mathbf{W}_{p1}\mathbf{c}(t) + \mathbf{b}_1)\nn\\
&\mathbf{v}(t) = \mathbf{o}(t)\circ h(\mathbf{c}(t))
\end{align}
where $\circ$ denotes the Hadamard (element-wise) product. 

Summary of notations used in Fig. \ref{Fig:LSTM} is as follows:
\begin{itemize}
\item ``$t$'': Stands for the time index in the sequence. For example, if we have 4 residual vectors of four different channels, we can show them as $\mathbf{r}(t),\; t=1,2,3,4$. 
\item ``$1$'': is a scalar
\item ``$\mathbf{W}_{reci}, \; i=1,2,3,4$'': Recurrent weight matrices of dimension $ncell \times ncell$ where $ncell$ is the number of cells in LSTM.
\item ``$\mathbf{W}_{i}, \; i=1,2,3,4$'': Input weight matrices of dimension $M \times ncell$ where $M$ is the number of random measurements in compressive sensing. These matrices map the residual vectors to feature space.
\item ``$\mathbf{b}_{i}, \; i=1,2,3,4$'': Bias vectors of size $ncell \times 1$. 
\item ``$\mathbf{W}_{pi}, \; i=1,2,3$'': Peephole connections of dimension $ncell \times ncell$.
\item ``$\mathbf{v}(t), \; t=1,2,\dots,L$'': Output of the cells. Vector of size $ncell \times 1$. $L$ is the number of channels in the MMV problem.
\item ``$\mathbf{i}(t), \mathbf{o}(t), \mathbf{y}_g(t), \; t=1,2,\dots,L$'': Input gates, output gates and inputs before gating respectively. Vector of size $ncell \times 1$.
\item ``$g(\cdot)$ and $h(\cdot)$'': $\tanh(\cdot)$ function.
\item ``$\sigma(\cdot)$'': Sigmoid function.
\end{itemize}

\section{Proposed Method}
\label{sec:ProposedMethod}
\subsection{High Level Picture}
\label{subsec:highlevel}
The summary of the proposed method is presented in Fig. \ref{Fig:BlockDiagram}. We initialize the residual vector, $\mathbf{r}$, for each channel by the measurement vector, $\mathbf{y}$, of that channel. These residual vectors serve as the input to the LSTM model that captures features of the residual vectors using input weight matrices ($\mathbf{W}_1$,$\mathbf{W}_2$,$\mathbf{W}_3$,$\mathbf{W}_4$) as well as the dependency among the residual vectors using recurrent weight matrices ($\mathbf{W}_{rec1}$,$\mathbf{W}_{rec2}$,$\mathbf{W}_{rec3}$,$\mathbf{W}_{rec4}$) and the central memory unit shown in Fig. \ref{Fig:LSTM}. A transformation matrix $\mathbf{U}$ is then used to transform, $\mathbf{v}\in\Re^{ncell\times 1}$, the output of each memory cell after gating, into the sparse vectors space, i.e., $\mathbf{z}\in\Re^{N\times 1}$. ``$ncell$'' is the number of cells in the LSTM model. Then a softmax layer is used for each channel to find the probability of each entry of each sparse vector being non-zero. For example, for channel 1, the $j$-th output of the softmax layer is:
\begin{equation}
\label{eq:softmax}
P(s_1(j)\vert \mathbf{r}_1) = \frac{e^{z(j)}}{\sum_{k=1}^N e^{z(k)}}
\end{equation} 
Then for each channel, the entry with the maximum probability value is selected and added to the support set of that channel. After that, given the new support set, the following least squares problem is solved to find an estimate of the sparse vector for the $j$-th channel:
\begin{equation}
\label{eq:LS}
\hat{\mathbf{s}}_j = \underset{\mathbf{s}_j}{\operatorname{argmin}} \Vert\mathbf{y}_j - \mathbf{A}^{\Omega_j}\mathbf{s}_j\Vert^2_2
\end{equation}
Using $\hat{\mathbf{s}}_j$, the new residual value for the $j$-th channel is calculated as follows:
\begin{equation}
\label{eq:residual}
\mathbf{r}_j = \mathbf{y}_j - \mathbf{A}^{\Omega_j}\hat{\mathbf{s}}_j
\end{equation}
This residual serves as the input to the LSTM model at the next iteration of the algorithm. The stopping criteria for the algorithm is when the residual values are small enough or when it has performed $N$ iterations where $N$ is the dimension of the sparse vector. Since we have used LSTM cells for the proposed method, we call it LSTM-CS algorithm. The pseudo-code of the proposed method is presented in Algorithm \ref{alg:LSTM-CS}.
\begin{algorithm}
\scriptsize
\caption{Distributed Compressive Sensing using Long Short-Term Memory (LSTM-CS)}
\label{alg:LSTM-CS}
\textbf{Inputs}: CS measurement matrix $\mathbf{A}\in\Re^{M\times N}$; matrix of measurements $\mathbf{Y}\in\Re^{M\times L}$; minimum $\ell_2$ norm of residual matrix ``$resMin$'' as stopping criterion; Trained ``$lstm$'' model\\
\textbf{Output}: Matrix of sparse vectors $\hat{\mathbf{S}}\in\Re^{N\times L}$\\
\textbf{Initialization}: $\hat{\mathbf{S}}=0$; $j=1$; $i=1$;  $\Omega=\emptyset$; 
$\mathbf{R}=\mathbf{Y}$.
\begin{algorithmic}[1]
\Procedure {LSTM-CS}{$\mathbf{A}$,$\mathbf{Y},lstm$}
\While{$i\leq N$ or $\Vert\mathbf{R}\Vert_2 \leq resMin$}
\State $i \leftarrow i+1$
\For {$j=1\rightarrow L$}
\State $\mathbf{R}(:,j)_i \leftarrow \frac{\mathbf{R}(:,j)_{i-1}}{max(\vert\mathbf{R}(:,j)_{i-1}\vert)}$
\State $\mathbf{v}_j \leftarrow lstm(\mathbf{R}(:,j)_i,\mathbf{v}_{j-1},\mathbf{c}_{j-1})$\Comment{$\mbox{LSTM}$}\label{alg:LSTM-CS:vj}
\State $\mathbf{z}_j \leftarrow \mathbf{U}\mathbf{v}_j$
\State $\mathbf{c}\leftarrow softmax(\mathbf{z}_j)$
\State $idx\leftarrow Support(max(\mathbf{c}))$
\State $\Omega_i\leftarrow\Omega_{i-1}\cup idx$
\State $\hat{\mathbf{S}}^{\Omega_i}(:,j)\leftarrow (\mathbf{A}^{\Omega_i})^{\dagger}\mathbf{Y}(:,j)$\Comment{$\mbox{Least Squares}$}\label{alg2LS}
\State $\hat{\mathbf{S}}^{\Omega_i^C}(:,j)\leftarrow 0$\label{alg2Complement}
\State $\mathbf{R}(:,j)_i\leftarrow \mathbf{Y}(:,j)-\mathbf{A}^{\Omega_i}\hat{\mathbf{S}}^{\Omega_i}(:,j)$
\EndFor
\EndWhile
\EndProcedure
\end{algorithmic}
\end{algorithm}

We continue by explaining how the training data is prepared from off-line dataset and then we present the details of the learning method. Please note that all the computations explained in the subsequent two sections are performed only once and they do not affect the run time of the proposed solver in Fig. \ref{Fig:BlockDiagram}. It is almost as fast as greedy algorithms in sparse reconstruction.
\subsection{Training Data Generation}
\label{subsec:traindatagen}
The main idea of the proposed method is to look at the sparse reconstruction problem as a two step task: a classification as the first step and a subsequent least squares as the second step. In the classification step, the aim is to find the atom of the dictionary, i.e., the column of $\mathbf{A}$, that is most relevant to the given residual of the current channel and the residuals of the previous channels. Therefore we need a set of residual vectors and their corresponding sparse vectors for supervised training. Since the training data and $\mathbf{A}$ are given, we can imitate the steps explained in the previous section to generate the residuals. This means that, given a sparse vector $\mathbf{s}$ with $k$ non-zero entries, we calculate $\mathbf{y}$ using \eqref{eq:Intro3}. Then we find the entry that has the maximum value in $\mathbf{s}$ and set it to zero. Assume that the index of this entry is $k_0$. This gives us a new sparse vector with $k-1$ non-zero entries. Then we calculate the residual vector from:
\begin{equation}
\label{eq:trainsample1}
\mathbf{r} = \mathbf{y} - \mathbf{a}_{k_0}  s(k_0)
\end{equation}
Where $\mathbf{a}_{k_0}$ is the $k_0$-th column of $\mathbf{A}$ and $s(k_0)$ is the $k_0$-th entry of $\mathbf{s}$. It is obvious that this residual value is because of not having the remaining $k-1$ non-zero entries of $\mathbf{s}$. From these remaining $k-1$ non-zero entries, the second largest value of $\mathbf{s}$ has the main contribution to $\mathbf{r}$ in \eqref{eq:trainsample1}. Therefore, we use $\mathbf{r}$ to predict the location of the second largest value of $\mathbf{s}$. Assume that the index of the second largest value of $\mathbf{s}$ is $k_1$. We define \textit{$\mathbf{s}_0$ as a one hot vector that has value $1$ at $k_1$-th entry and zero at other entries}. Therefore, the training pair is $(\mathbf{r},\mathbf{s}_0)$. 

Now we set the $k_1$-th entry of $\mathbf{s}$ to zero. This gives us a new sparse vector with $k-2$ non-zero entries. Then we calculate the new residual vector from:
\begin{equation}
\label{eq:trainsample2}
\mathbf{r} = \mathbf{y} - [\mathbf{a}_{k_0},\mathbf{a}_{k_1}]  [s(k_0),s(k_1)]^T
\end{equation}
We use the residual in \eqref{eq:trainsample2} to predict the location of the third largest value in $\mathbf{s}$. Assume that the index of the third largest value of $\mathbf{s}$ is $k_2$. We define \textit{$\mathbf{s}_0$ as a one hot vector that has value $1$ at $k_2$-th entry and zero at other entries}. Therefore, the new training pair is $(\mathbf{r},\mathbf{s}_0)$. 

The above procedure is continued upto the point that $\mathbf{s}$ does not have any non-zero entry. Then the same procedure is used for the next training sample. This gives us training samples for one channel. Then the same procedure is used for the next channel in $\mathbf{S}$. Since the number of non-zero entries, $k$, is not known in advance, we assume a maximum number of non-zero entries per channel for training data generation. 

\subsection{Learning Method}
\label{subsec:LearningMethod}
To calculate the parameters of the proposed model, i.e., $\mathbf{W}_1,\mathbf{W}_2,\mathbf{W}_3,\mathbf{W}_4$, $\mathbf{W}_{rec1},\mathbf{W}_{rec2},\mathbf{W}_{rec3},\mathbf{W}_{rec4}$, $\mathbf{W}_{p1},\mathbf{W}_{p2},\mathbf{W}_{p3}$, $\mathbf{b}_1,\mathbf{b}_2,\mathbf{b}_3,\mathbf{b}_4$ in Fig. \ref{Fig:LSTM} and transformation matrix $\mathbf{U}$ in Fig.\ref{Fig:BlockDiagram}, we minimize a cross entropy cost function over the training data. Assuming $\mathbf{s}$ is the output vector of the softmax layer given by the model in Fig. \ref{Fig:BlockDiagram} (output of the softmax layer is represented as conditional probabilities in Fig. \ref{Fig:BlockDiagram}) and $\mathbf{s}_0$ is the one hot vector explained in the previous section, the following optimization problem is solved:
\begin{align}
\label{eq:cost}
&L(\mathbf{\Lambda}) = \underset{\mathbf{\Lambda}}{\operatorname{min}}\left\{\sum_{i=1}^{nB}\sum_{r=1}^{Bsize}\sum_{\tau =1}^{L}\sum_{j=1}^N L_{r,i,\tau ,j}(\mathbf{\Lambda})\right\} \nn\\
&L_{r,i,\tau ,j}(\mathbf{\Lambda}) = - s_{0,r,i,\tau}(j)log(s_{r,i,\tau}(j))
\end{align}
where $nB$ is the number of mini-batches in the training data, $Bsize$ is the number of training data pairs, $(\mathbf{r},\mathbf{s}_0)$, in each mini-batch, $L$ is the number of channels in the MMV problem, i.e., number of columns of $\mathbf{S}$, and $N$ is the length of vector $\mathbf{s}$ and $\mathbf{s}_0$. $\mathbf{\Lambda}$ denotes the collection of the model parameters that includes $\mathbf{W}_1$, $\mathbf{W}_2$, $\mathbf{W}_3$, $\mathbf{W}_4$, $\mathbf{W}_{rec1}$, $\mathbf{W}_{rec2}$, $\mathbf{W}_{rec3}$, $\mathbf{W}_{rec4}$, $\mathbf{W}_{p1}$, $\mathbf{W}_{p2}$, $\mathbf{W}_{p3}$, $\mathbf{b}_1$, $\mathbf{b}_2$, $\mathbf{b}_3$ and $\mathbf{b}_4$ in Fig. \ref{Fig:LSTM} and $\mathbf{U}$ in Fig. \ref{Fig:BlockDiagram}.

To solve the optimization problem in \eqref{eq:cost}, we use Backpropagation through time (BPTT) with Nesterov method. The update equations for parameter $\mathbf{\Lambda}$ at epoch $k$ are as follows:
\begin{align}
\label{eq:Nesterov}
&\triangle\mathbf{\Lambda} _k = \mathbf{\Lambda} _k - \mathbf{\Lambda} _{k-1}\nn\\
&\triangle\mathbf{\Lambda} _k = \mu_{k-1}\triangle\mathbf{\Lambda} _{k-1} - \epsilon_{k-1}\nabla L(\mathbf{\Lambda} _{k-1} + \mu_{k-1}\triangle\mathbf{\Lambda} _{k-1})
\end{align}
where $\nabla L(\cdot)$ is the gradient of the cost function in \eqref{eq:cost}, $\epsilon$ is the learning rate and $\mu_k$ is a momentum parameter determined by the scheduling scheme used for training. Above equations are equivalent to Nesterov method in \cite{Nestrov1983}. To see why, please refer to appendix A.1 of \cite{RNNhinton2013} where the Nesterov method is derived as a momentum method. The gradient of the cost function, $\nabla L(\mathbf{\Lambda})$, is: 
\begin{equation}
\label{eq:costGrad}
\nabla L(\mathbf{\Lambda}) = \sum_{i=1}^{nB}\underbrace{\sum_{r=1}^{Bsize}\sum_{\tau =1}^{L}\sum_{j=1}^N \frac{\partial L_{r,i,\tau ,j}(\mathbf{\Lambda})}{\partial\mathbf{\Lambda}}}_{\mathrm{one\; large\; update}}
\end{equation}
As it is obvious from \eqref{eq:costGrad}, since we have unfolded the LSTM over channels in $\mathbf{S}$, we fold it back when we want to calculate gradients over the whole sequence of channels.

$\frac{\partial L_{r,i,\tau ,j}(\mathbf{\Lambda})}{\partial\mathbf{\Lambda}}$ in \eqref{eq:costGrad} and error signals for different parameters of the proposed model that are necessary for training are presented in Appendix \ref{sec:appGradientExpress}. Due to lack of space, we omit the presentation of full derivation of the gradients. 

We have used mini-batch training to accelerate training and one large update instead of incremental updates during back propagation through time. To resolve the gradient explosion problem we have used gradient clipping. To accelerate the convergence, we have used Nesterov method \cite{Nestrov1983} and found it effective in training the proposed model for the MMV problem. 

We have used a simple yet effective scheduling for $\mu_k$ in \eqref{eq:Nesterov}, in the first and last 10\% of all parameter updates $\mu_k = 0.9$ and for the other 80\% of all parameter updates $\mu_k = 0.995$. We have used a fixed step size for training LSTM. Please note that since we are using mini-batch training, all parameters are updated for each mini-batch in \eqref{eq:costGrad}.

A summary of training method for LSTM-CS is presented in Algorithm \ref{alg:TrainLSTM-CS}.
\begin{algorithm}[t]
\caption{Training the proposed model for Distributed Compressive Sensing}
\label{alg:TrainLSTM-CS}
\begin{algorithmic}
\scriptsize
\State \textbf{Inputs}: Fixed step size ``$\epsilon$'', Scheduling for ``$\mu$'', Gradient clip threshold ``$th_G$'', Maximum number of Epochs ``$nEpoch$'', Total number of training pairs in each mini-batch ``$Bsize$'', Number of channels for the MMV problem ``$L$''. 
\State \textbf{Outputs}: LSTM-CS trained model for distributed compressive sensing ``$\mathbf{\Lambda}$''.
\State \textbf{Initialization}: Set all parameters in $\mathbf{\Lambda}$ to small random numbers, $i=0$, $k=1$.
\Procedure {LSTM-CS}{$\mathbf{\Lambda}$}
\While {$i \leq nEpoch$}
\For {``first minibatch'' $\rightarrow$ ``last minibatch''}
\State $r \leftarrow 1$
\While{$r \leq Bsize$}
\State Compute $\sum_{\tau=1}^{L}\frac{\partial L_{r,\tau}}{\partial\mathbf{\Lambda}_k}$
\State \Comment{use \eqref{eq:appendix5} to \eqref{eq:LSTM47} in appendix \ref{sec:appGradientExpress}}
\State $r \leftarrow r+1$
\EndWhile
\State Compute $\nabla L(\mathbf{\Lambda}_k) \leftarrow$ ``sum above terms over $r$''
\If{$\nabla L(\mathbf{\Lambda}_k) > th_G$}
\State $\nabla L(\mathbf{\Lambda}_k) \leftarrow th_G$\\ 
\Comment {For each entry of the gradient matrix $\nabla L(\mathbf{\Lambda}_k)$}
\EndIf
\State Compute $\triangle\mathbf{\Lambda} _k$\Comment {use \eqref{eq:Nesterov}}
\State Update: $\mathbf{\Lambda} _k \leftarrow \triangle\mathbf{\Lambda} _k + \mathbf{\Lambda} _{k-1}$
\State $k \leftarrow k+1$
\EndFor
\State $i \leftarrow i+1$
\EndWhile
\EndProcedure
\end{algorithmic}
\end{algorithm}

Although the training method and derivatives in Appendix \ref{sec:appGradientExpress} are presented for all parameters in LSTM, in the implementation ,we have removed peephole connections and forget gates. Since length of each sequence, i.e., the number of columns in $\mathbf{S}$, is known in advance, we set state of each cell to zero in the beginning of a new sequence. Therefore, forget gates are not a great help here. Also, as long as the order of columns in $\mathbf{S}$ is kept, the precise timing in the sequence is not of great concern, therefore, peephole connections are not that important as well. Removing peephole connections and forget gate will also help to have less training time, i.e., less number of parameters need to be tuned during training.

\section{Experimental Results and Discussion}
\label{sec:Experiments}
We have performed the experiments on two real world datasets, the first is the MNIST dataset of handwritten digits \cite{mnist} and the second is three different classes of images from natural image dataset of Microsoft Research in Cambridge \cite{MSRimage}.

In this section, we would like to answer the following questions: (i) How is the performance of different reconstruction algorithms for the MMV problem, including the proposed method, when different channels, i.e., different columns in $\mathbf{S}$, have different sparsity patterns? (ii) Does the proposed method perform well enough when there is correlation among different sparse vectors? E.g., when sparse vectors are DCT or Wavelet transform of different blocks of an image? (iii) How fast is the proposed method compared to other reconstruction algorithms for the MMV problem? (iv) How robust is the proposed method to noise?

 For all the results presented in this section, the reconstruction error is defined as:
\begin{equation}
\label{eq:mse}
NMSE = \frac{\Vert \mathbf{\hat{S}}-\mathbf{S} \Vert}{\Vert \mathbf{S} \Vert}
\end{equation}
where $\mathbf{S}$ is the actual sparse matrix and $\mathbf{\hat{S}}$ is the recovered sparse matrix from random measurements by the reconstruction algorithm. The machine used to perform the experiments has an Intel(R) Core(TM) i7 CPU with clock 2.93 GHz and with 16 GB RAM.
\subsection{MNIST Dataset}
\label{sec:MNIST}
\begin{figure}[t]
\includegraphics[width = 0.5\textwidth]{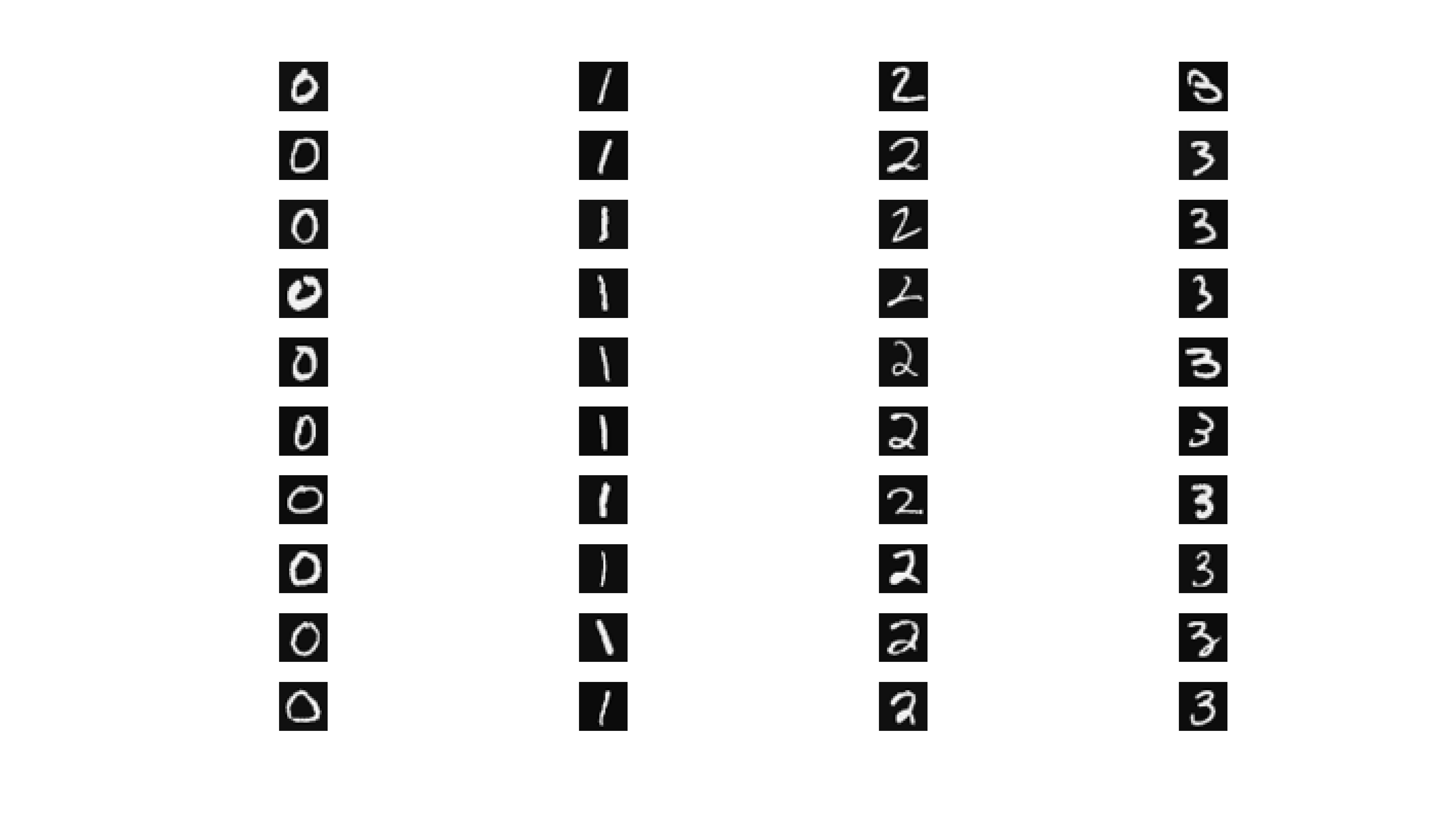}
\caption{Randomly selected images for test from MNIST dataset. The first channel encodes digit zero, the second channel encodes digit one and so on.}
\label{fig:TestImageMNIST}
\end{figure}

MNIST is a dataset of handwritten digits where the images of the digits are normalized in size and centred so that we have fixed size images. The task is to simultaneously encode 4 images each of size $24\times 24$, i.e., we have 4 channels and $L=4$ in \eqref{eq:Intro4}. The encoder is a typical compressive sensing encoder, i.e., a randomly generated matrix $\mathbf{A}$. We have normalized each column of $\mathbf{A}$ to have unit norm. Since the images are already sparse, i.e., have a few number of non-zero pixels, no transform, $\mathbf{\Psi}$ in \eqref{eq:Intro2}, is used. To simulate the measurement noise, we have added a Gaussian noise with standard deviation $0.005$ to the measurement matrix $\mathbf{Y}$ in \eqref{eq:Intro4}. This results in measurements with signal to noise ratio (SNR) of approximately $46 dB$. We have divided each image into four $12\times 12$ blocks. This means that the length of each sparse vector is $N=144$. We have taken $50\%$ random measurements from each sparse vector, i.e., $M=72$. After receiving and reconstructing all blocks at the decoder, we compute the reconstruction error defined in \eqref{eq:mse} for the full image. We have randomly selected 10 images for each digit from the set $\{0,1,2,3\}$, i.e., 40 images in total for the test. This means that the first column of $\mathbf{S}$ is an image of digit $0$, the second column is an image of digit $1$, the third column is an image of digit $2$ and the fourth column is an image of digit $3$. Test images are represented in Fig. \ref{fig:TestImageMNIST}.

We have compared the performance of the proposed reconstruction algorithm (LSTM-CS) with 7 reconstruction methods for the MMV problem. These methods are: 
\begin{itemize}
\item Simultaneous Orthogonal Matching Pursuit (SOMP) which is a well known baseline for the MMV problem.
\item Bayesian Compressive Sensing (BCS)\cite{Carin1} applied independently on each channel. For the BCS method we set the initial noise variance of $i$-th channel to the value suggested by the authors, i.e.,  $std(\mathbf{y}_i)^2/100$ where $i\in \{1,2,3,4\}$ and $std(.)$ calculates the standard deviation. We set the threshold for stopping the algorithm to $10^{-8}$.
\item Multitask Compressive Sensing (MT-BCS) \cite{Carin2} which takes into account the statistical dependency of different channels. For MT-BCS we set the parameters of the Gamma prior on noise variance to $a=100/0.1$ and $b=1$ which are the values suggested by the authors. We set the stopping threshold to $10^{-8}$ as well.
\item Sparse Bayesian Learning for Temporally correlated sources (T-SBL) \cite{Rao3} which exploits correlation among different sources in the MMV problem. For T-SBL, we used the default values proposed by the authors.  
\item Nonlinear Weighted SOMP (NWSOMP) \cite{hpICASSP2013} which solves a regression problem to help the SOMP algorithm with prior knowledge from training data. For NWSOMP, during training, we used one layer, 512 neurons and 25 epochs of parameters update.
\item Compressive Sensing on Least Squares Residual (LSCS) \cite{LSCS} where no explicit joint sparsity assumption is made in the design of the method. For LSCS, we used $sigma0 =  cc* (1/3)*sqrt(Sav/m)$ suggested by the authors where $m$ is the number of measurements and $Sav = 16$ as suggested by the author. We tried a range of different values of $cc$ and got the best results with $cc=0.1$. We also set $sigsys = 1,\: siginit = 3$ and $lambdap = 4$ as suggested by the author. 
\item The method proposed in \cite{AfterRev2,TSPminorRev} and referred to as PCSBL-GAMP where sparse Bayesian learning is used to design the method and no explicit joint sparsity assumption is made. For PCSBL-GAMP, we used $beta = 1$, $Pattern = 2$ because we need the coupling among the sparse vectors, i.e., left and right coupling, maximum number of iterations equal to $maxiter = 400$, and $C = 1e0$ as suggested by the authors for the noisy case.
\end{itemize}

\begin{figure}[t]
\centerline{
\subfigure{\includegraphics[width = 0.5\textwidth]{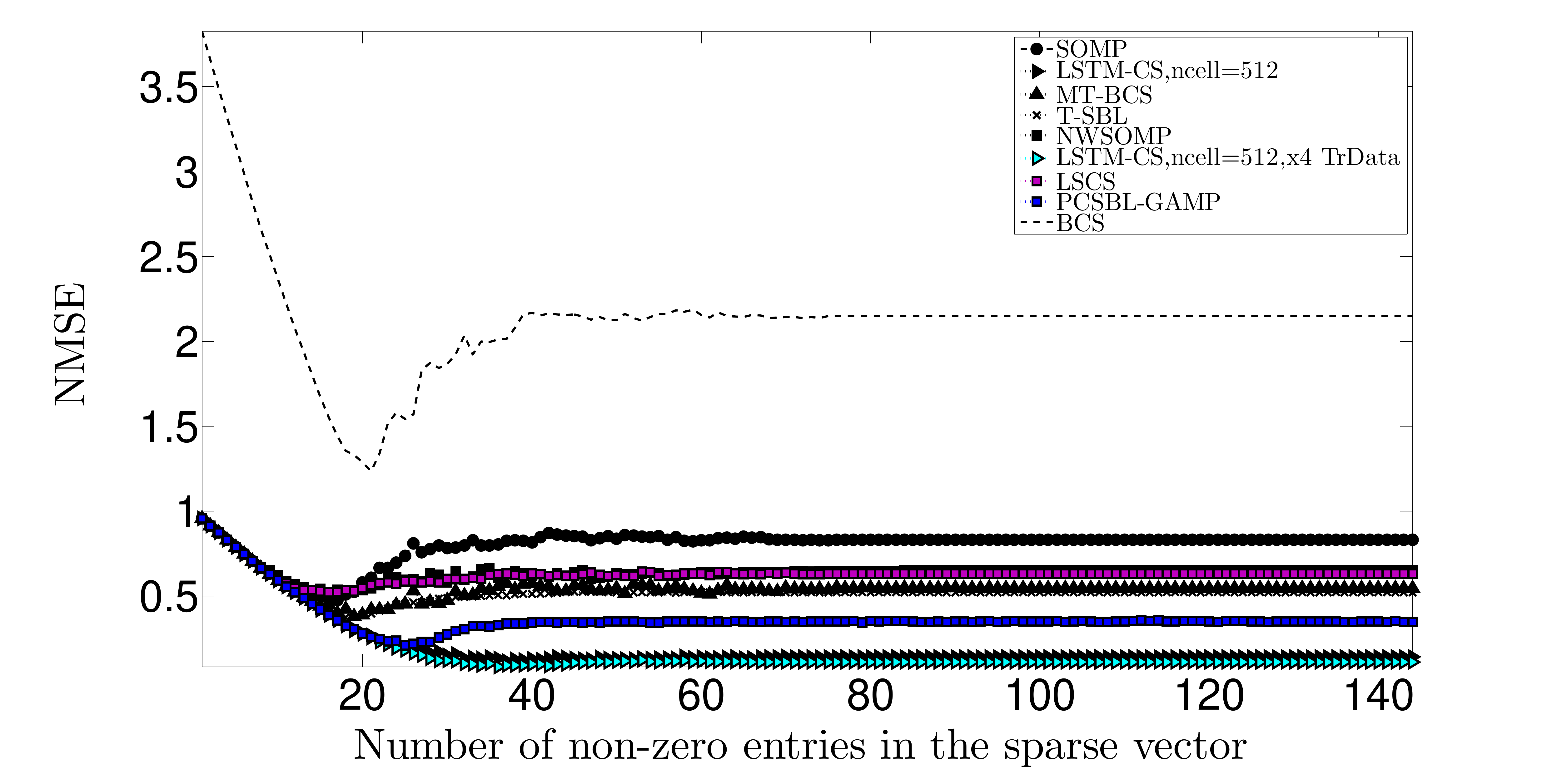}}
}
\centerline{
\subfigure{\includegraphics[width = 0.5\textwidth]{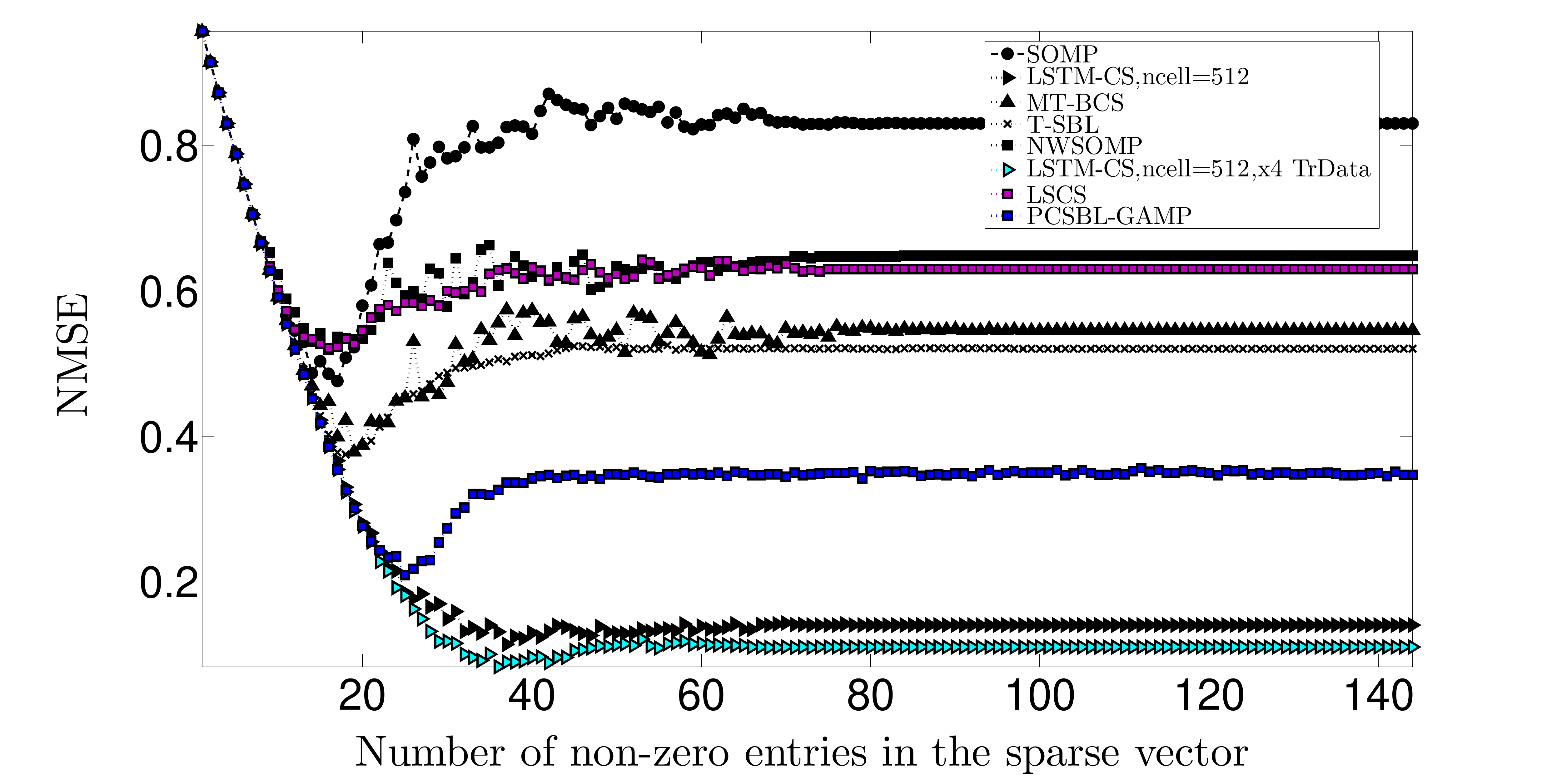}}
}
\caption{Comparison of different MMV reconstruction algorithms for MNIST dataset. Bottom figure is the same as top figure without results of BCS algorithm to make the difference among different algorithms more visible. In this experiment $M=72$ and $N=144$.}
\label{fig:MNISTresultm72}
\end{figure}

For LSTM-CS, during training, we used one layer, 512 cells and 25 epochs of parameter updates. We used only 200 images for the training set. The training set does not include any of the 40 images used for test. To monitor and prevent overfitting, we used 3 images per channel as the validation set and we used early stopping if necessary. Please note that the images used for validation were not used in the training set or in the test set. Results are presented in Fig. \ref{fig:MNISTresultm72}.

In Fig. \ref{fig:MNISTresultm72}, the vertical axis is the $NMSE$ defined in \eqref{eq:mse} and horizontal axis is the number of non-zero entries in the sparse vector. The number of measurements, $M$, is fixed to $72$. Each point on the curves in Fig. \ref{fig:MNISTresultm72} is the average of $NMSE$ over $40$ reconstructed test images at the decoder. 

\begin{figure*}[t]
\includegraphics[width =  \textwidth]{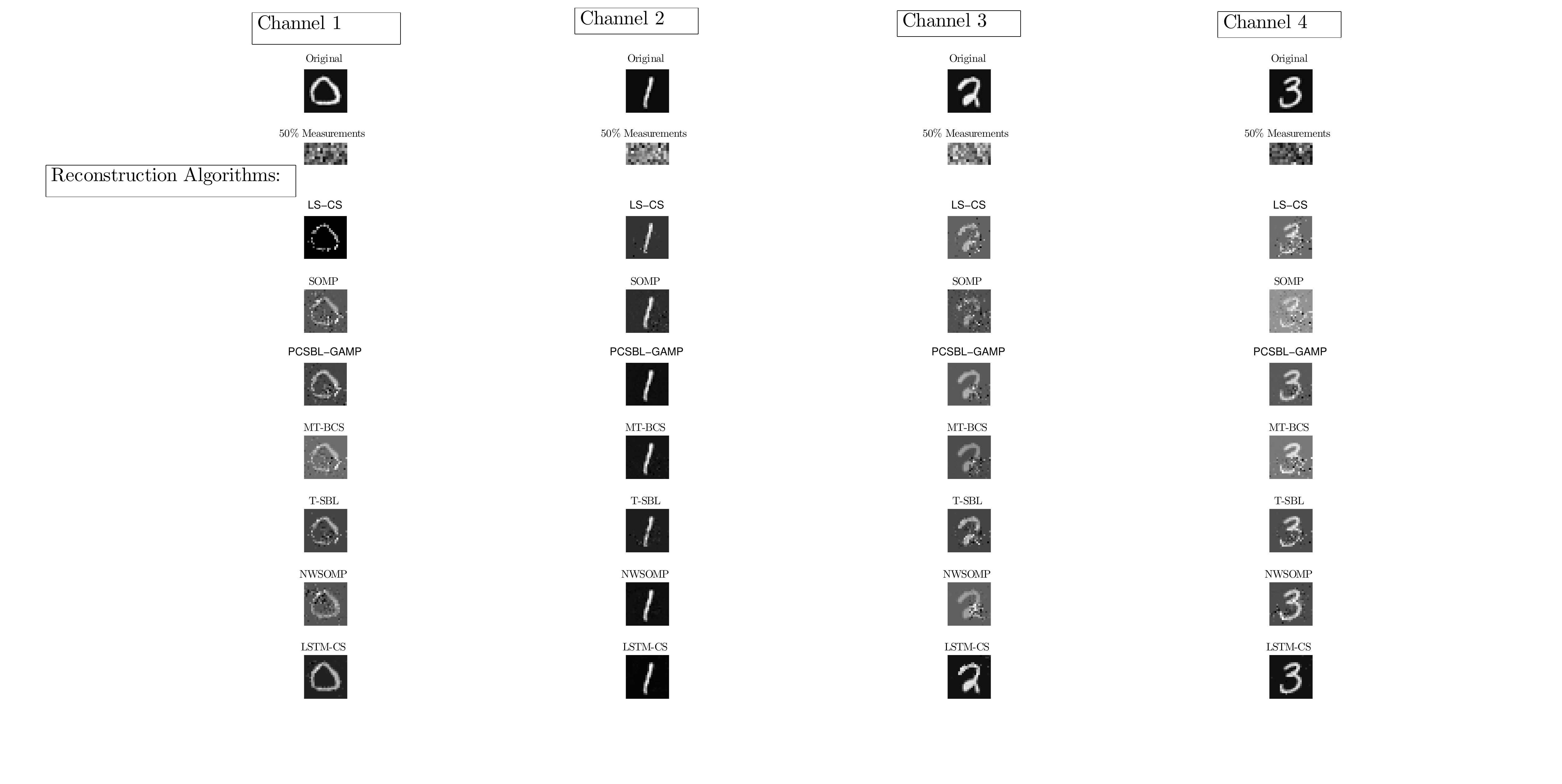}
\caption{Reconstructed images using different MMV reconstruction algorithms for $4$ images of the MNIST dataset. First row are original images, $\mathbf{S}$, second row are measurement matrices, $\mathbf{Y}$, third row are reconstructed images using LS-CS, fourth row are reconstructed images using SOMP, fifth row using PCSBL-GAMP, sixth row using MT-BCS, seventh row using T-SBL, eighth row using NWSOMP and the last row are reconstructed images using the proposed LSTM-CS method.}
\label{fig:MNISTrecImages}
\end{figure*}

For the MNIST dataset, we observe from Fig. \ref{fig:MNISTresultm72} that LSTM-CS significantly outperforms the reconstruction algorithms for the MMV problem discussed in this paper. One important reason for this is that existing MMV solvers rely on the joint sparsity in $\mathbf{S}$, while the proposed method does not rely on this assumption. Another reason is that the structure of each sparse vector is effectively captured by LSTM. The reconstructed images using different MMV reconstruction algorithms for $4$ test images are presented in Fig. \ref{fig:MNISTrecImages}. An interesting observation from Fig. \ref{fig:MNISTrecImages} is that the accuracy of reconstruction depends on the complexity of the sparsity pattern. For example when the sparsity pattern is simple, e.g., image of digit $1$ in Fig. \ref{fig:MNISTrecImages}, all the algorithms perform well. But when the sparsity pattern is more complex, e.g., image of digit $0$ in Fig. \ref{fig:MNISTrecImages}, then their reconstruction accuracy degrades significantly. 

\begin{figure}[t]
\centerline{
\subfigure{\includegraphics[width = 0.5\textwidth]{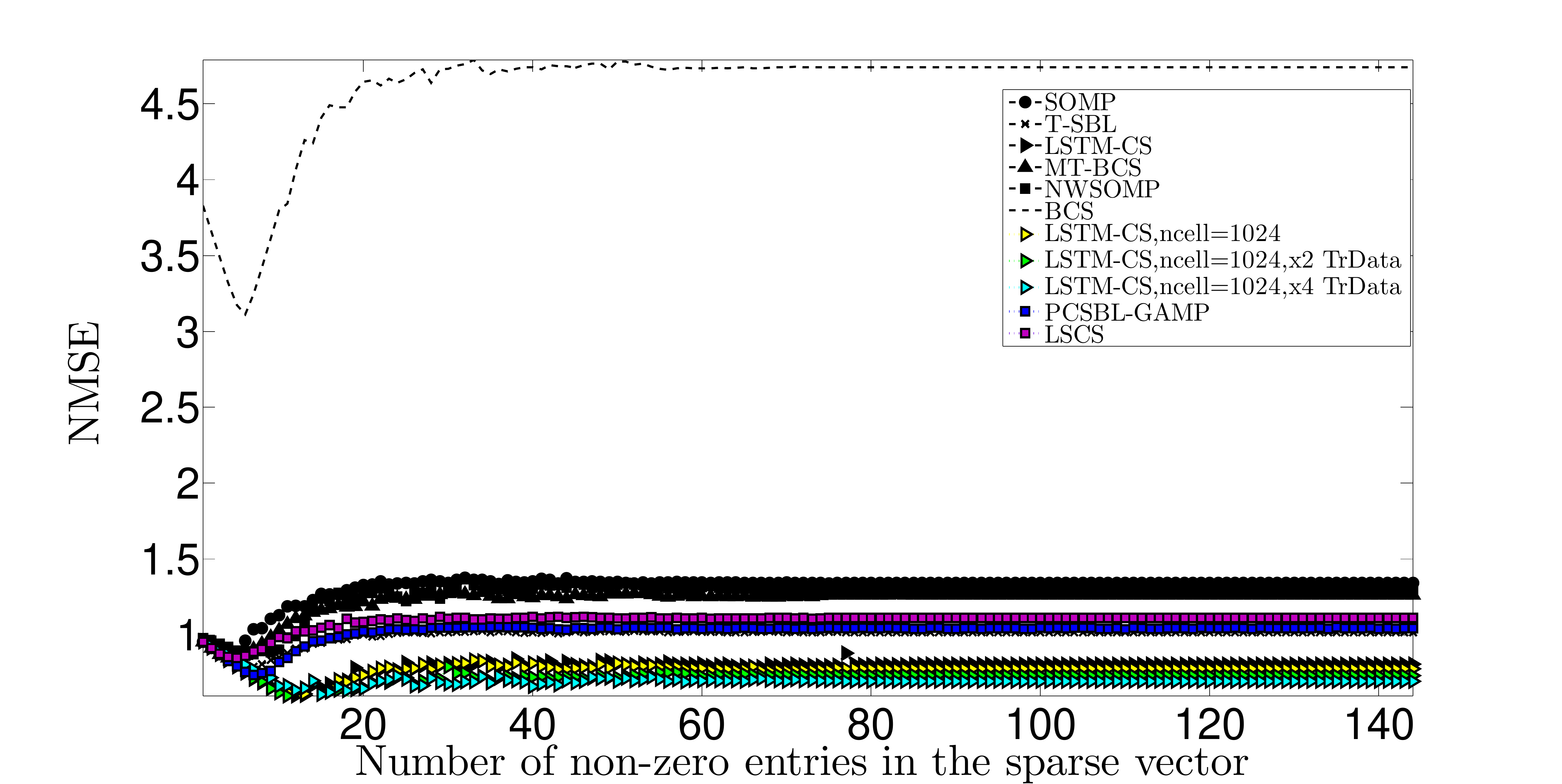}}
}
\centerline{
\subfigure{\includegraphics[width = 0.5\textwidth]{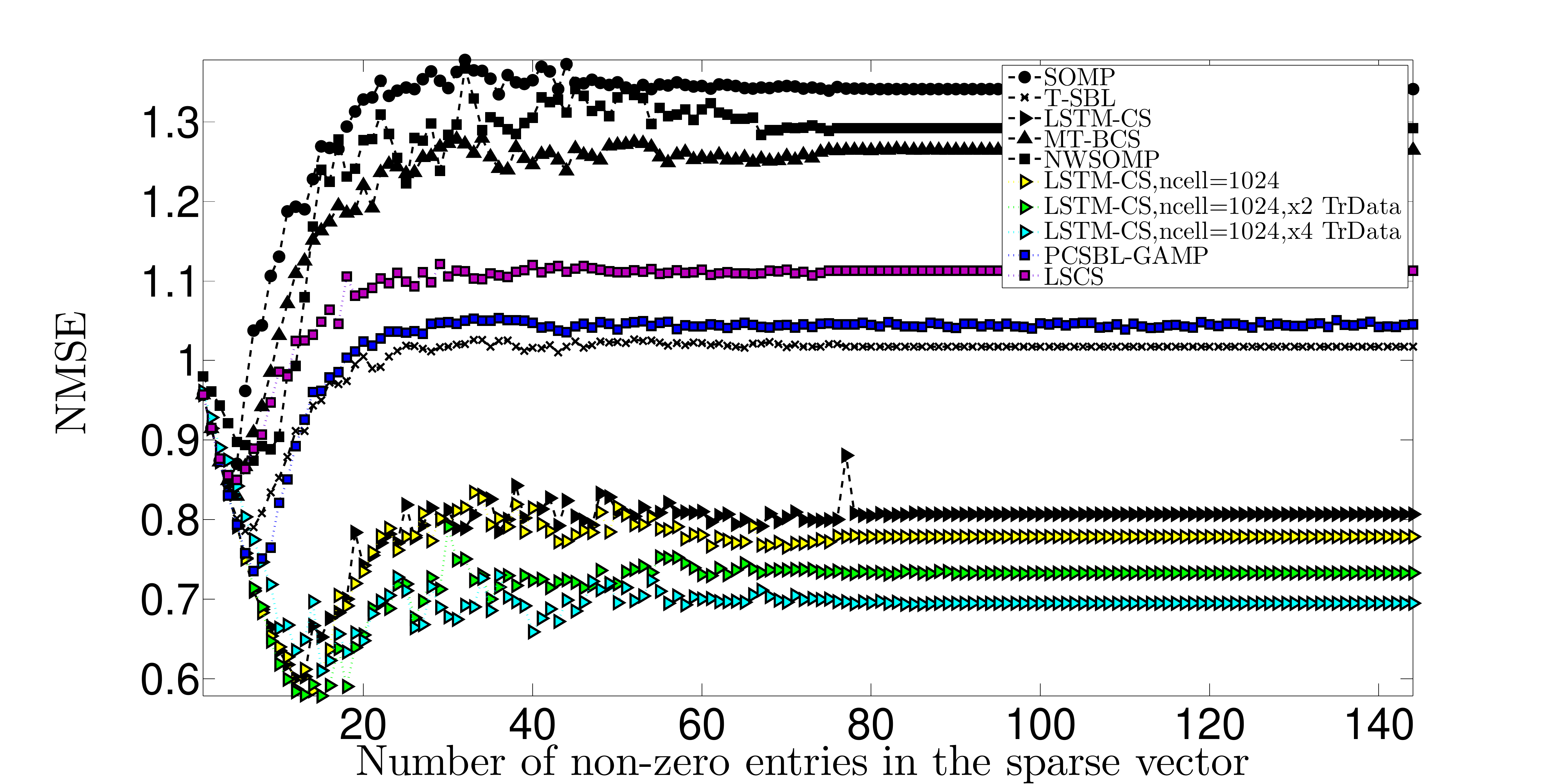}}
}
\caption{Comparison of different MMV reconstruction algorithms for MNIST dataset. Bottom figure is the same as top figure without results of BCS algorithm to make the difference among different algorithms more visible. In this experiment $M=36$ and $N=144$.}
\label{fig:MNISTresultm36}
\end{figure}

We have repeated the experiments on the MNIST dataset with $25\%$ random measurements, i.e., $M=36$. The results are presented in Fig. \ref{fig:MNISTresultm36}. We trained 4 different LSTM models for this experiment. The first one is the same model used for previous experiment ($m=72$). In the second model, we increased the number of cells in the LSTM model from 512 to 1024. In the third and fourth models, we used 2 times and 4 times more training data respectively. The rest of the experiments' settings was similar to the settings described before. As observed from these results, by investing more on training a good LSTM model, LSTM-CS method performs better. 

\begin{figure*}[t]
\centering
\subfigure[Results for all Methods.]{
\includegraphics[width = 0.45\textwidth ,height =0.35 \textwidth]{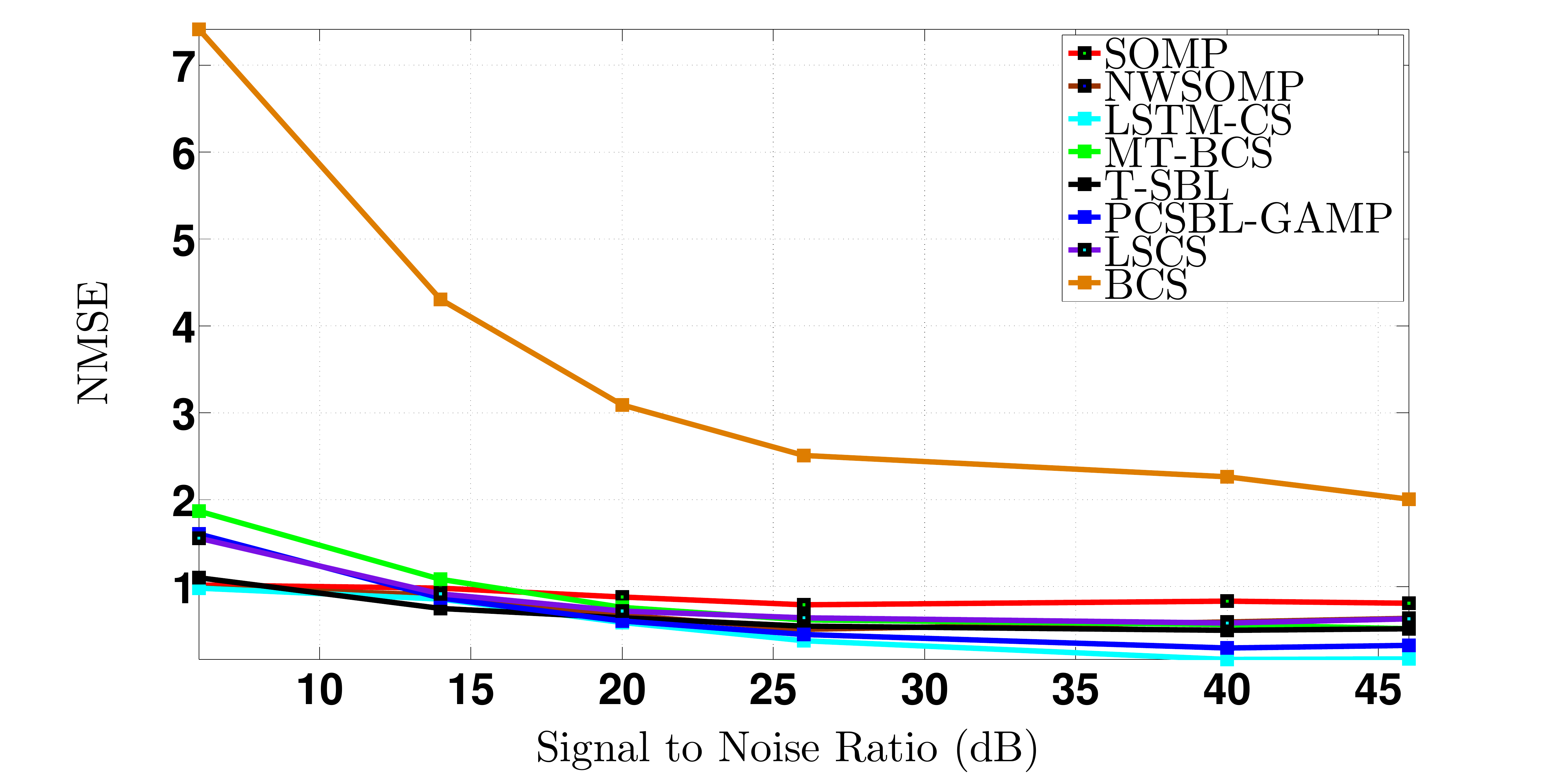}
\label{Fig:Noise_a}}
\quad
\subfigure[Results without BCS method for a more clear visibility.]{
\includegraphics[width = 0.45\textwidth ,height = 0.35\textwidth]{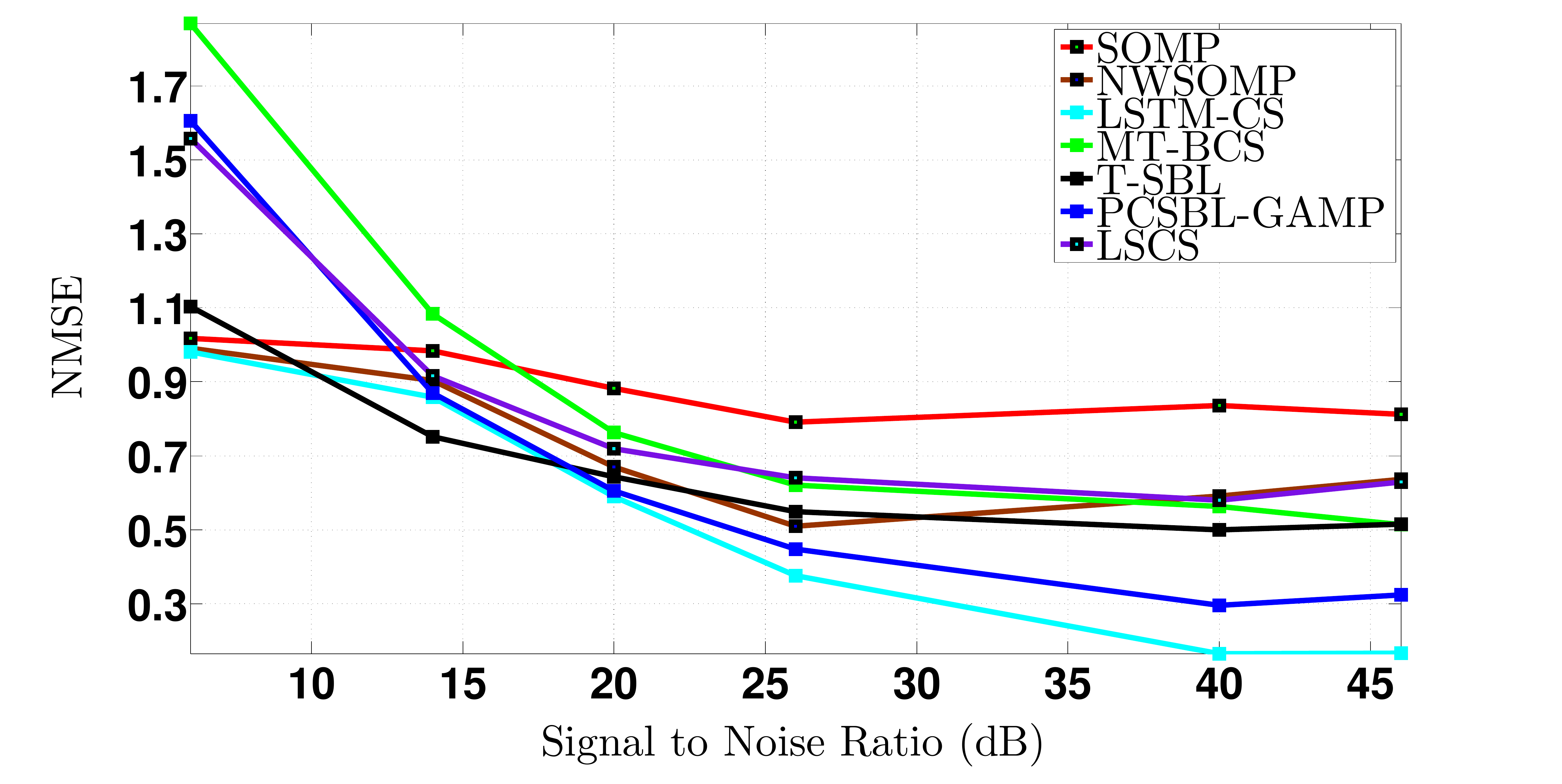}
\label{Fig:Noise_b}}
\caption{Reconstruction performance of the methods discussed in the paper for different noise levels.}
\label{Fig:Noise}
\end{figure*}

All the results presented so far are for noisy measurements where an additive Gaussian noise with standard deviation 0.005 is used ($SNR \simeq 46 dB$). To evaluate the stability of the proposed LSTM-CS method to noise, and compare it with other methods discussed in this paper, an experiment was performed using the following range of noise standard deviations:
\begin{align}
\label{eq:sigma}
\sigma = \{ 0.5, 0.2, 0.1, 0.05, 0.01, 0.005 \}
\end{align}
where $\sigma$ is the standard deviation of noise. This approximately corresponds to:
\begin{align}
\label{eq:snr}
SNR = \{ 6\:dB, 14\:dB, 20\:dB, 26\:dB, 40\:dB, 46\:dB \}
\end{align}
We used the same experimental settings explained above. Results are presented in Fig. \ref{Fig:Noise}.

As observed from the results, in very noisy environment, i.e., SNR = 6 dB, performance of MT-BCS , LSCS  and PCSBL-GAMP degrades significantly while T-SBL , NWSOMP and LSTM-CS (proposed in this paper) methods show less severe degradation. In very low noise environment, i.e., SNR = 46 dB, performance of LSTM-CS, trained with just 512 cells and 200 training images, is better than other methods. In medium noise environment, i.e., SNR = 20 dB and SNR = 26 dB, performance of LSTM-CS, T-SBL and PCSBL-GAMP are close (although LSTM-CS is slightly better). Please note that the performance of LSTM-CS can be further improved by using a better architecture (e.g., more cells, more training data or more layers) as explained previously. 

\begin{figure}[t]
\centering
\includegraphics[width = 0.45\textwidth ,height = 0.35\textwidth]{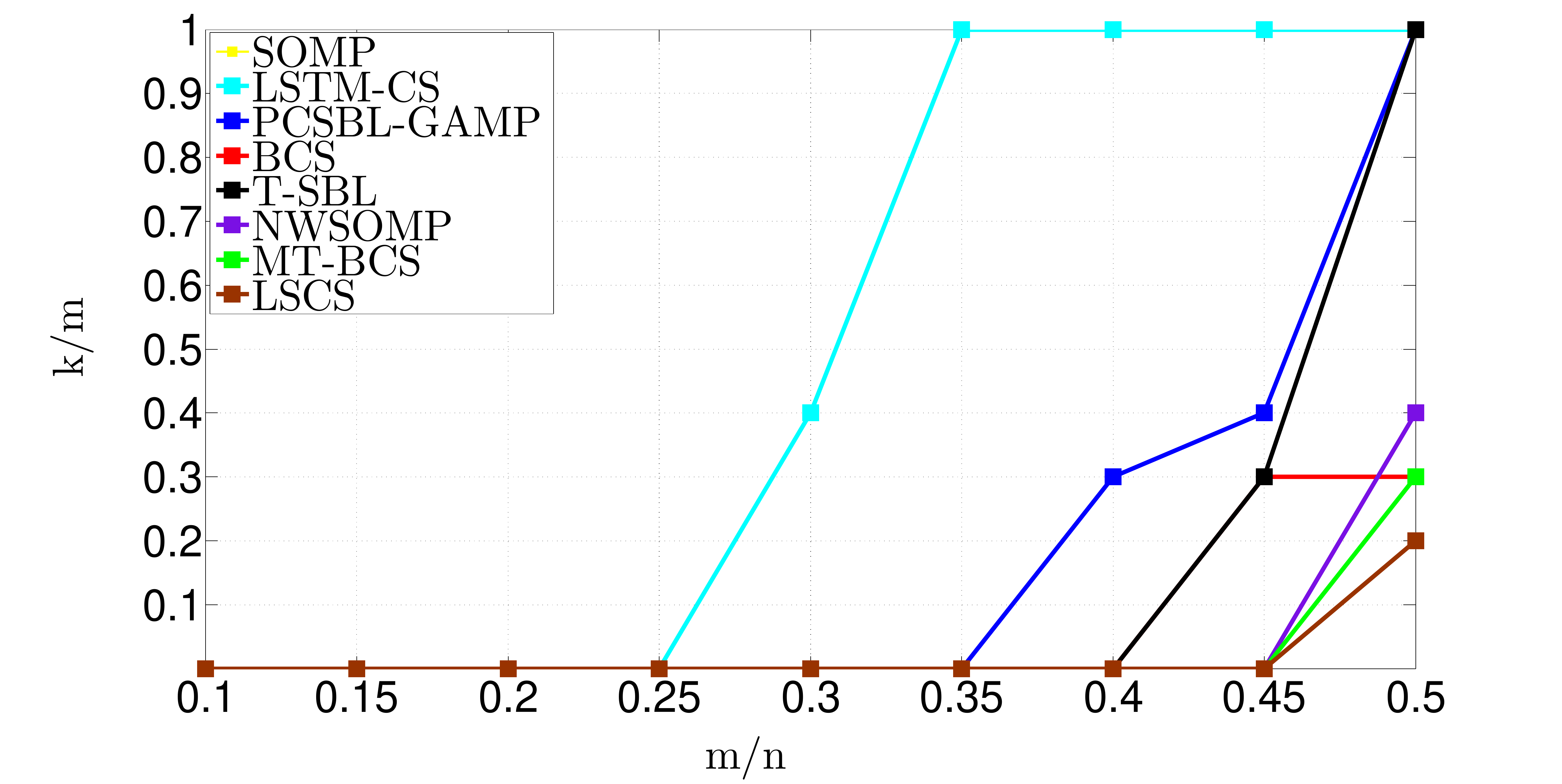}
\caption{Phase transition diagram for different methods on MNIST dataset where $90\%$ perfect recovery is considered. Assuming a perfect recovery condition of NMSE $\leq 0.6$ for this dataset. ``$n$'' is the number of entries in each sparse vector, ``$m$'' is the number of random measurements and ``$k$'' is the number of non-zero entries in each sparse vector.}
\label{Fig:phase}
\end{figure}

To present the phase transition diagram of solvers, we used a simple LSTM-CS solver that uses 512 cells and just 200 training images. The performance was evaluated over the following values of $\frac{m}{n}$ where $n$ is the number of entries in each sparse vector and $m$ is the number of measurements per channel:
\begin{align}
\label{eq:m/n}
\frac{m}{n} = \{0.10,0.15,0.20,0.25,0.30,0.35,0.40,0.45,0.50 \}
\end{align}
For this experiment, we randomly selected $50$ images per channel from MNIST dataset. Since we have $L=4$ channels, and each image is of size $24 \times 24$, and each image has $4$ blocks of $12 \times 12$ pixels, in total we will have $50 \times 4 \times 4 = 800$ sparse vectors. Considering Fig. \ref{fig:MNISTresultm72} of the paper, NMSE of most solvers is about 0.6. Therefore we set the following as the condition for perfect recovery: if more than $90\%$ of test images are reconstructed with an NMSE of 0.6 or less, count that test image as perfectly recovered. We did this for each $\frac{m}{n}$ in \eqref{eq:m/n}. Results are presented in Fig. \ref{Fig:phase}. 
Results presented in Fig.\ref{Fig:phase} shows the reconstruction performance improvement when LSTM-CS method is used.

\begin{figure*}[t]
\centering
\subfigure[Results for all Methods.]{
\includegraphics[width = 0.45\textwidth ,height = 0.35\textwidth]{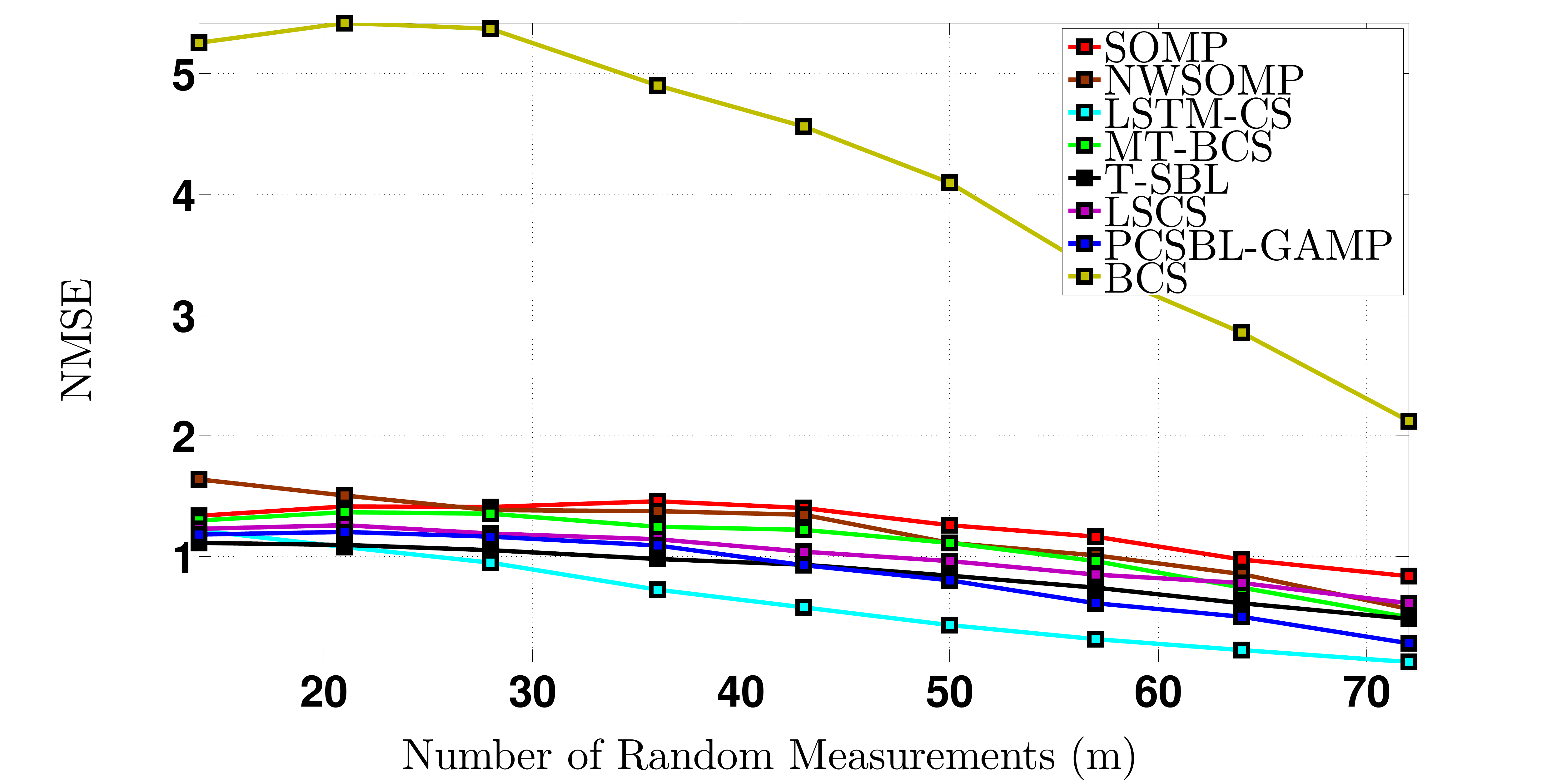}
\label{Fig:vs_m-rev2-a}}
\quad
\subfigure[Results without BCS method for a more clear visibility.]{
\includegraphics[width = 0.45\textwidth ,height = 0.35\textwidth]{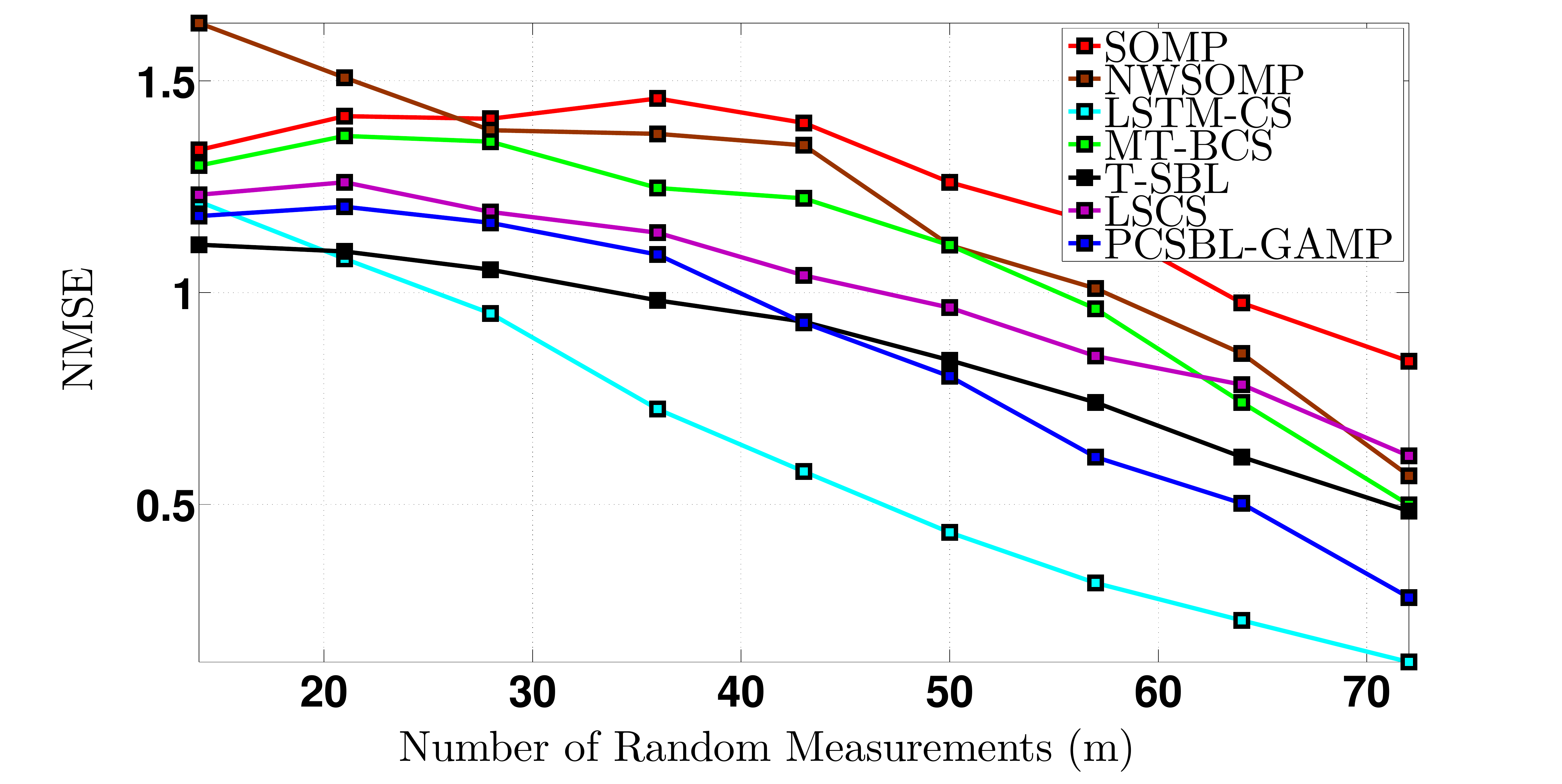}
\label{Fig:vs_m-rev2-b}}
\caption{Comparison of different MMV reconstruction algorithms for different number of random measurements for MNIST dataset. In this experiment n = 144.}
\label{Fig:vs_m-rev2}
\end{figure*}

We also present the performance of LSTM-CS for different number of random measurements. We used the set of random measurements in \eqref{eq:m/n} with $n=144$. We used an LSTM with 512 cells and 400 training images. The settings for all other methods was similar to the one described before. Results are presented in Fig. \ref{Fig:vs_m-rev2}.
As observed from Fig. \ref{Fig:vs_m-rev2}, using LSTM-CS method improves the reconstruction performance compared to other methods discussed in this paper.

\subsection{Natural Images Dataset}
\label{sec:NaturalImage}
\begin{figure*}[t]
\centerline{
\subfigure{\includegraphics[width = 0.1\textwidth]{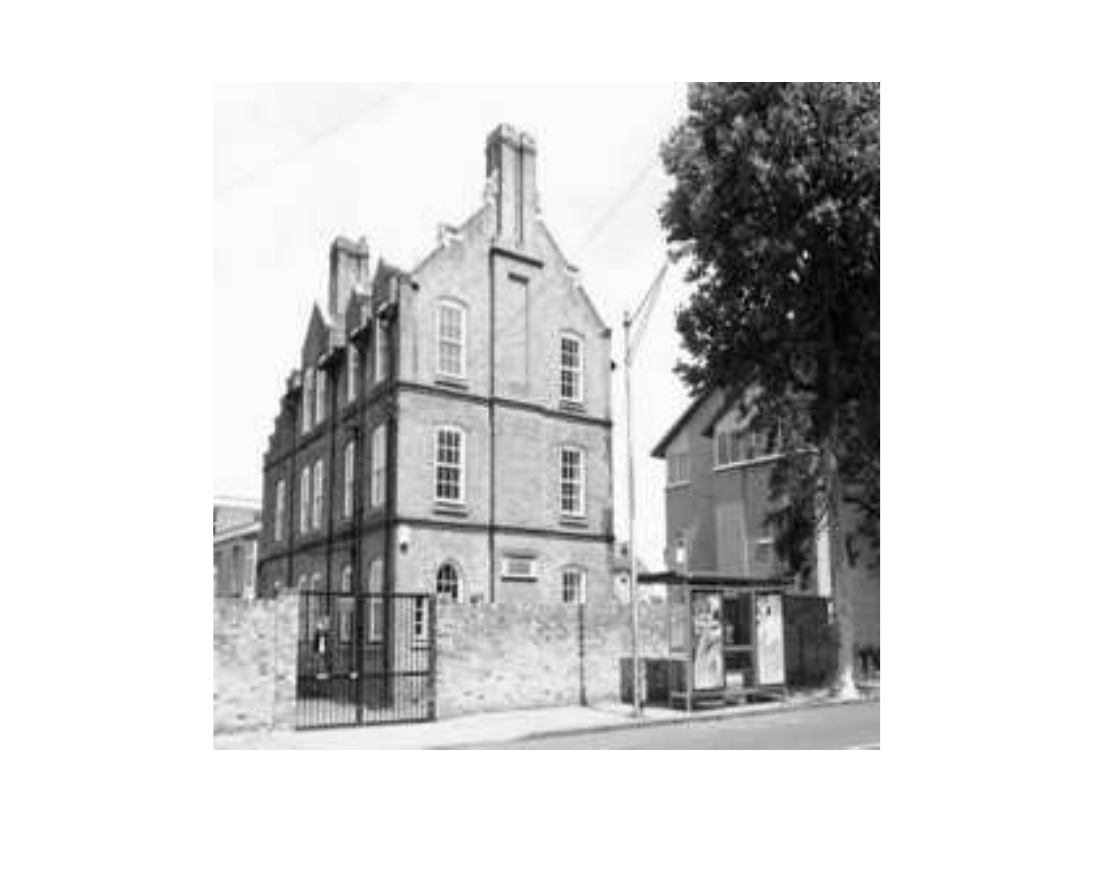}}
\subfigure{\includegraphics[width = 0.1\textwidth]{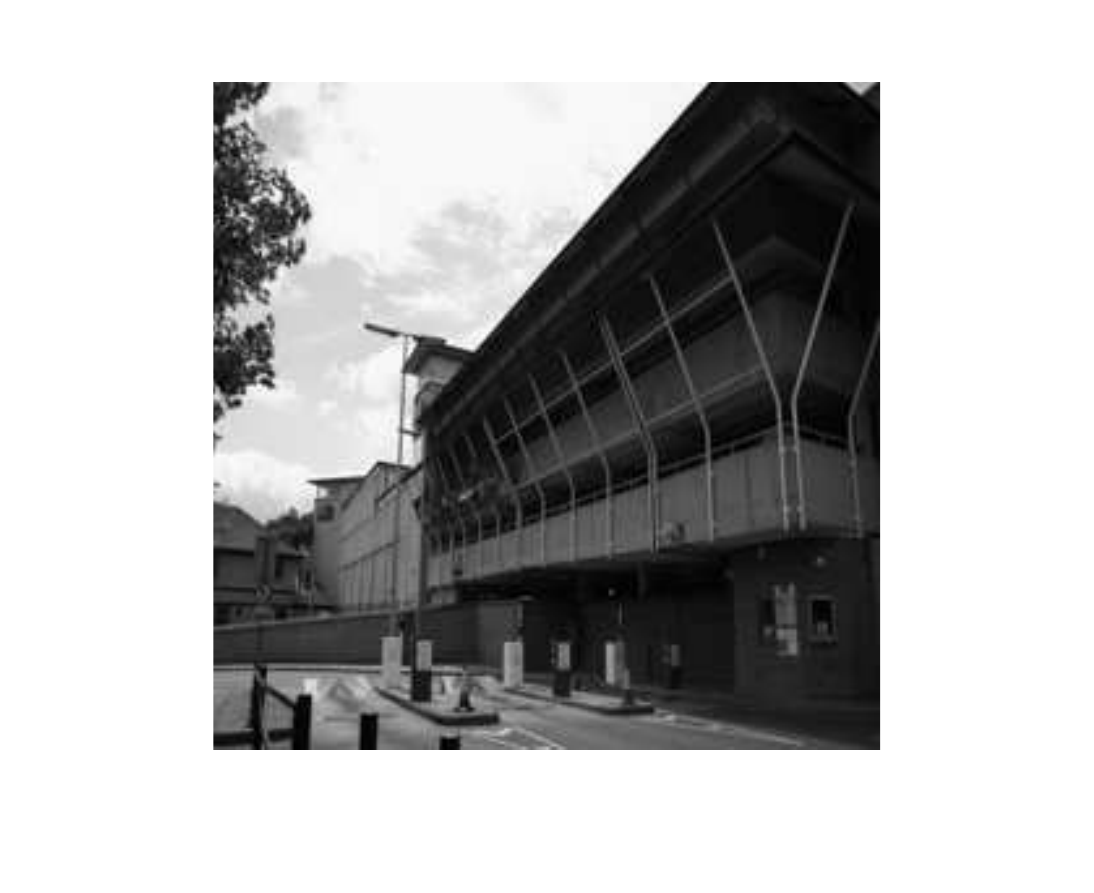}}
\subfigure{\includegraphics[width = 0.1\textwidth]{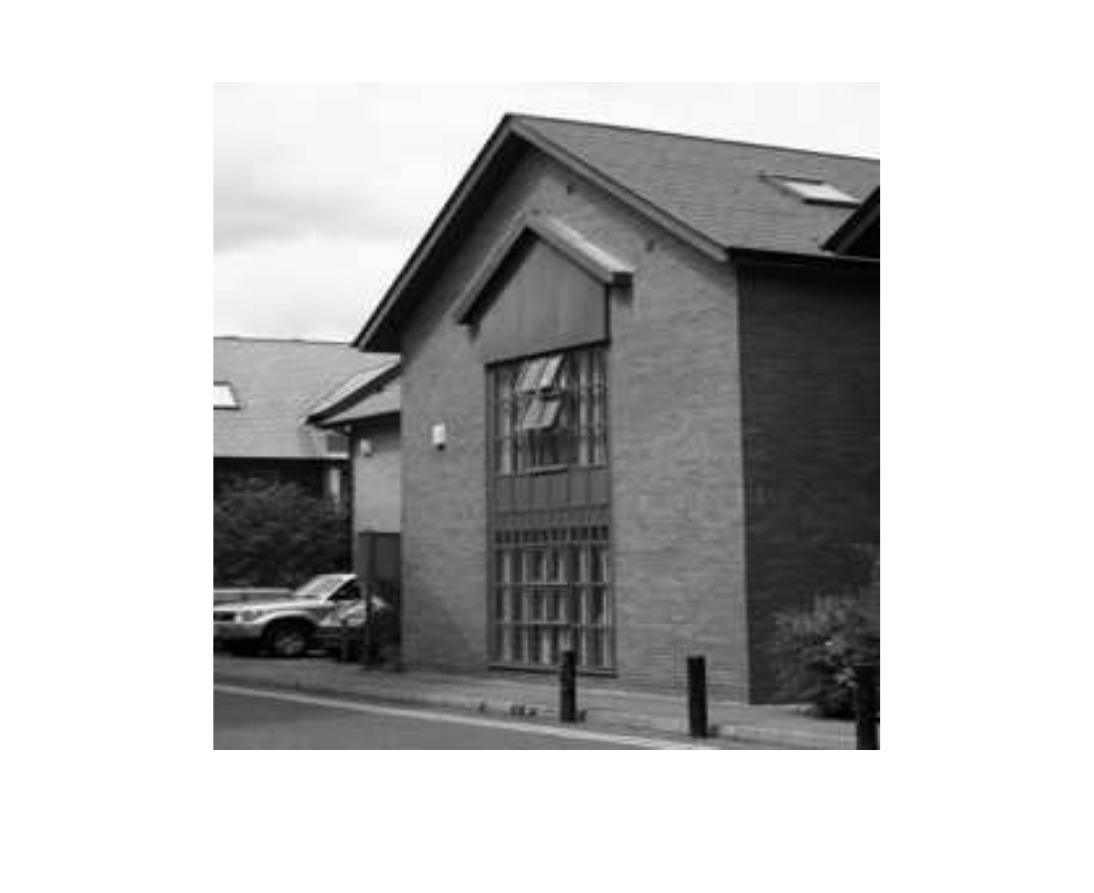}}
\subfigure{\includegraphics[width = 0.1\textwidth]{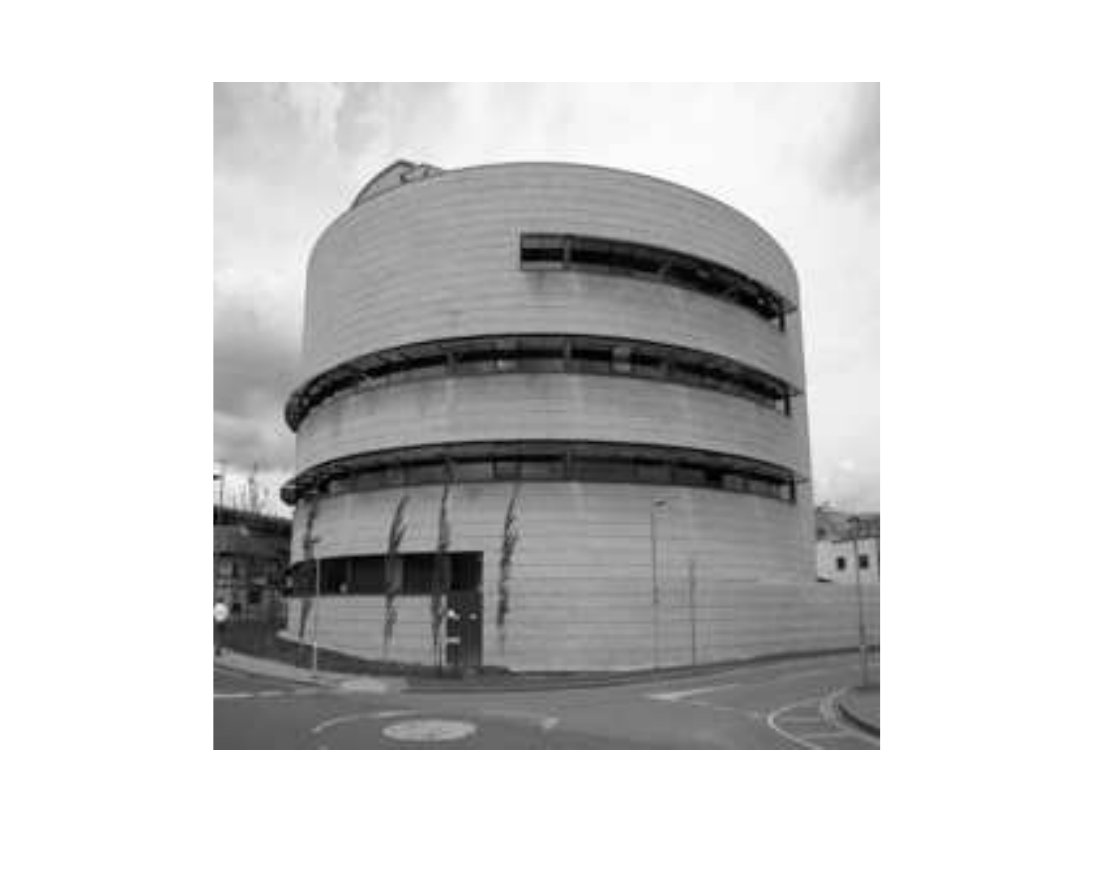}}
\subfigure{\includegraphics[width = 0.1\textwidth]{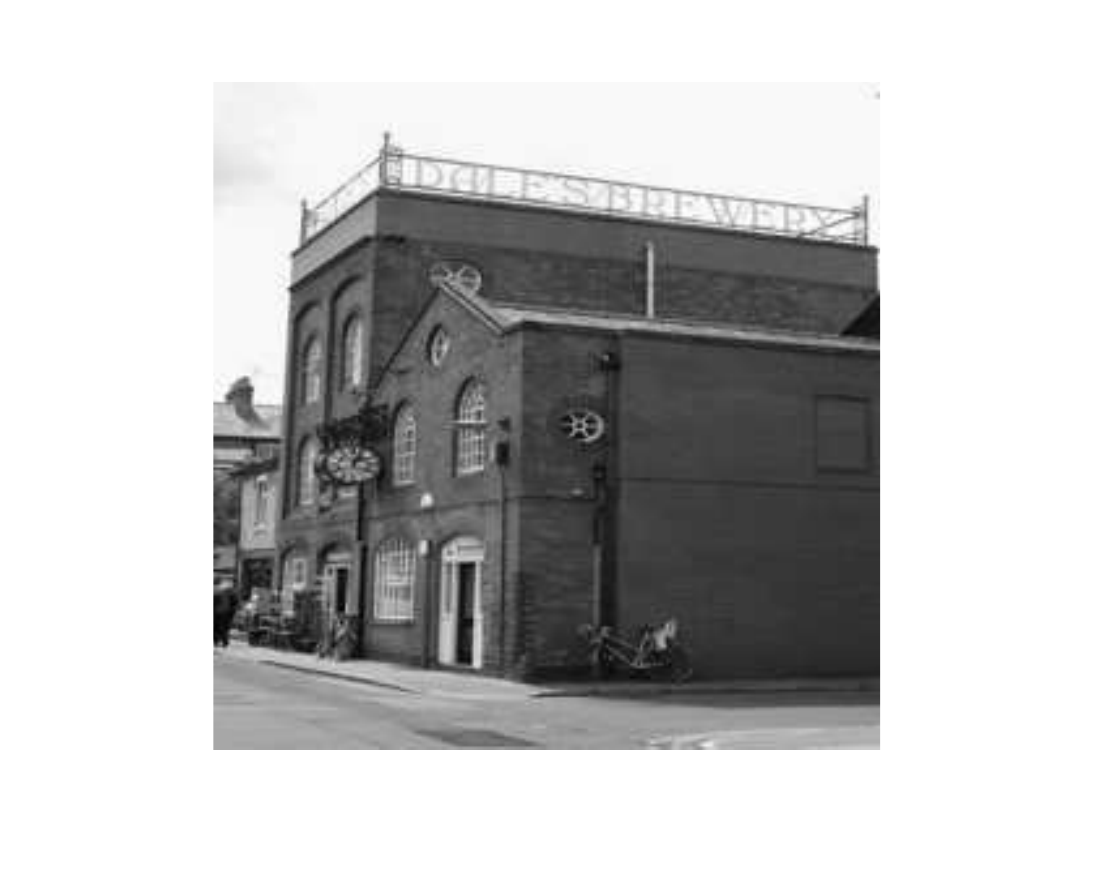}}
\subfigure{\includegraphics[width = 0.1\textwidth]{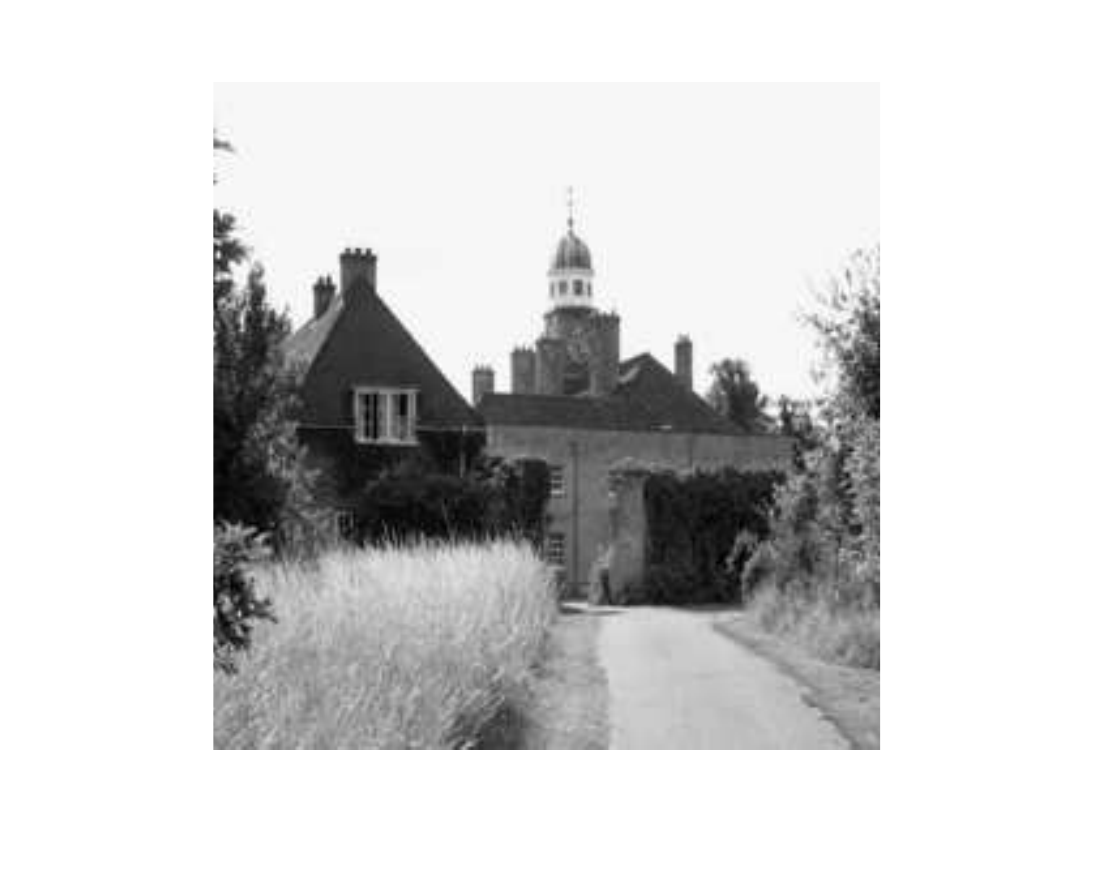}}
\subfigure{\includegraphics[width = 0.1\textwidth]{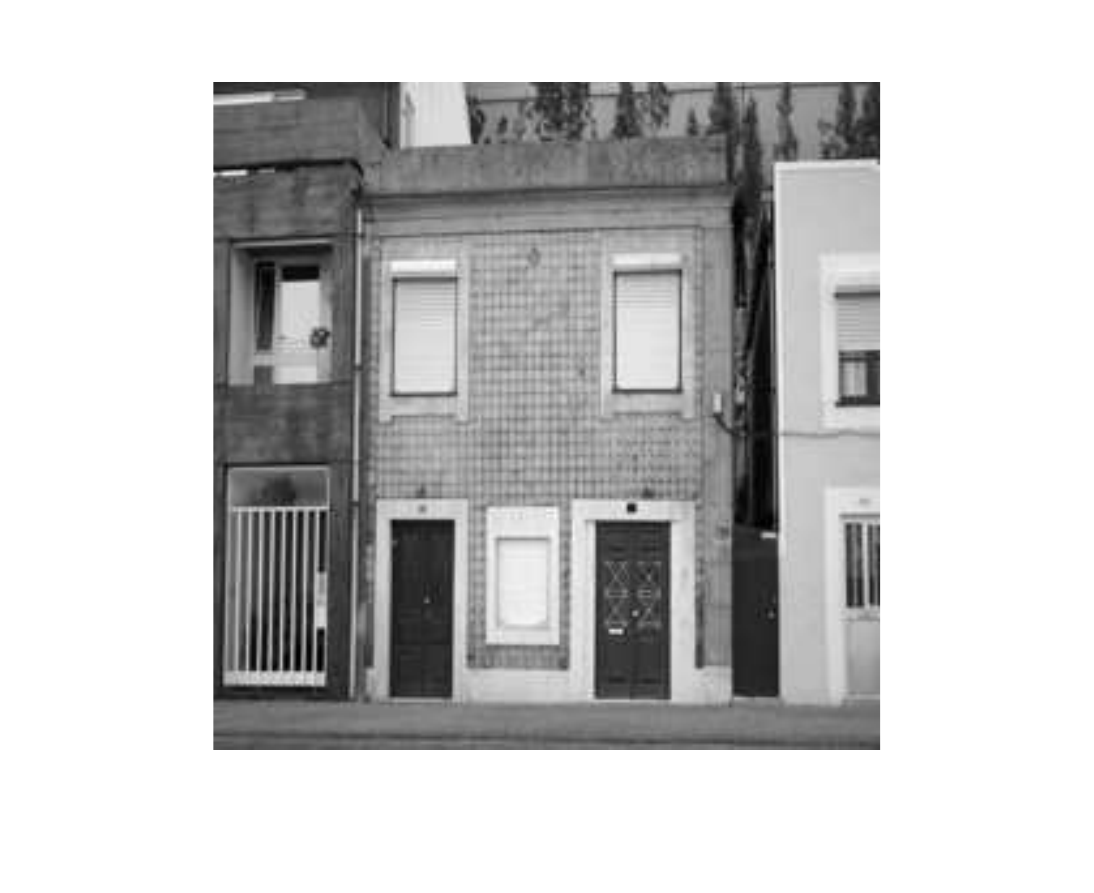}}
\subfigure{\includegraphics[width = 0.1\textwidth]{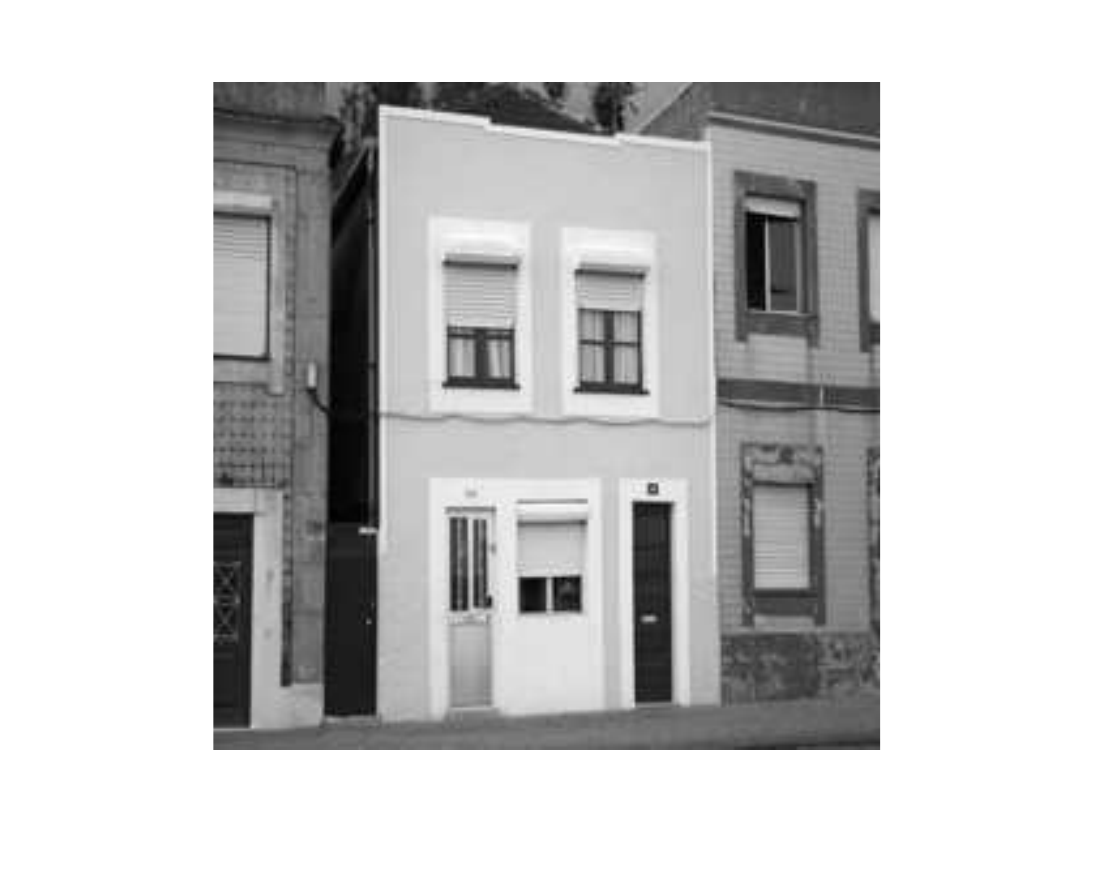}}
\subfigure{\includegraphics[width = 0.1\textwidth]{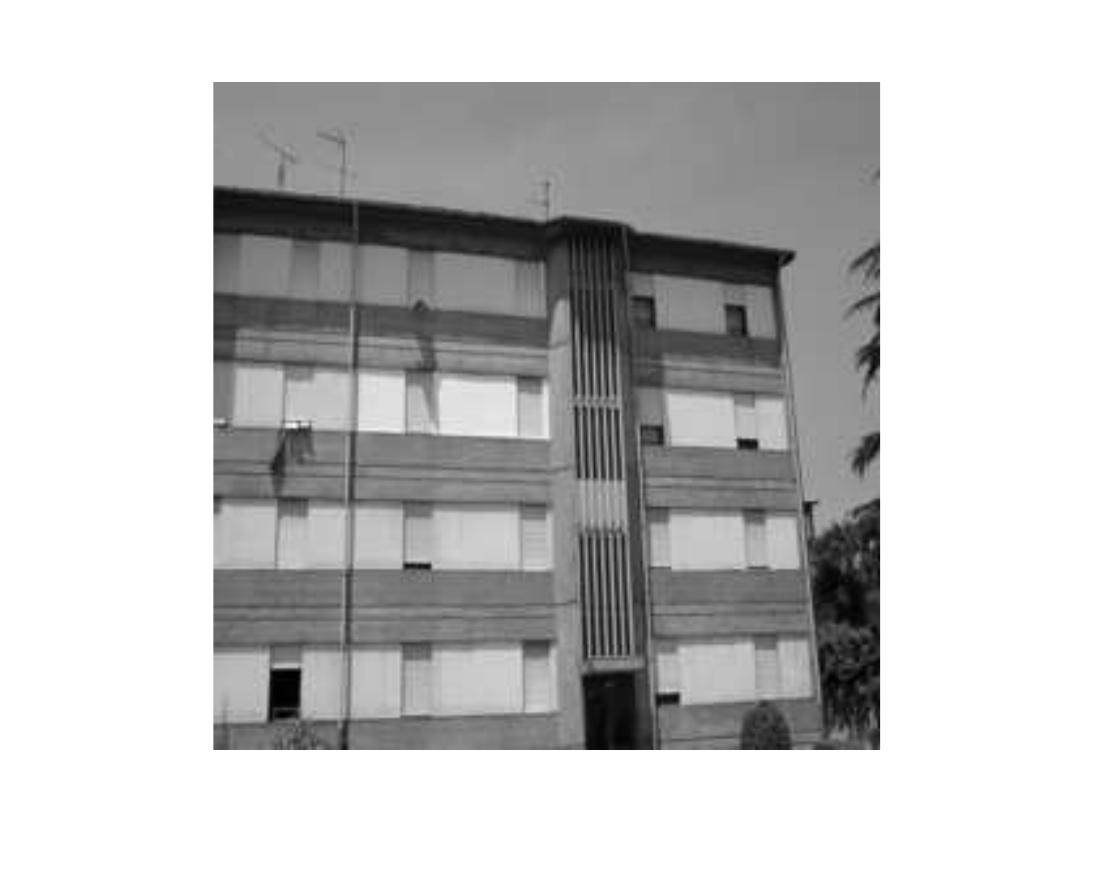}}
\subfigure{\includegraphics[width = 0.1\textwidth]{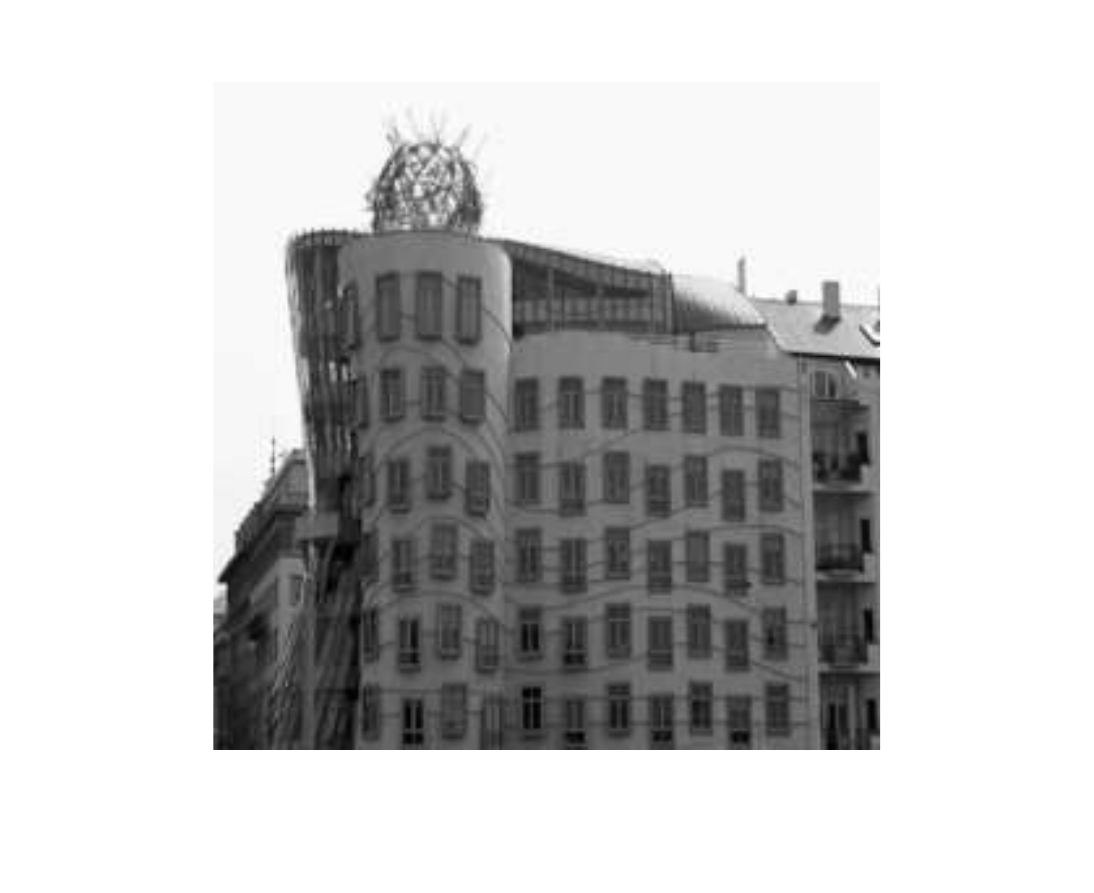}}
}
\centerline{
\subfigure{\includegraphics[width = 0.1\textwidth]{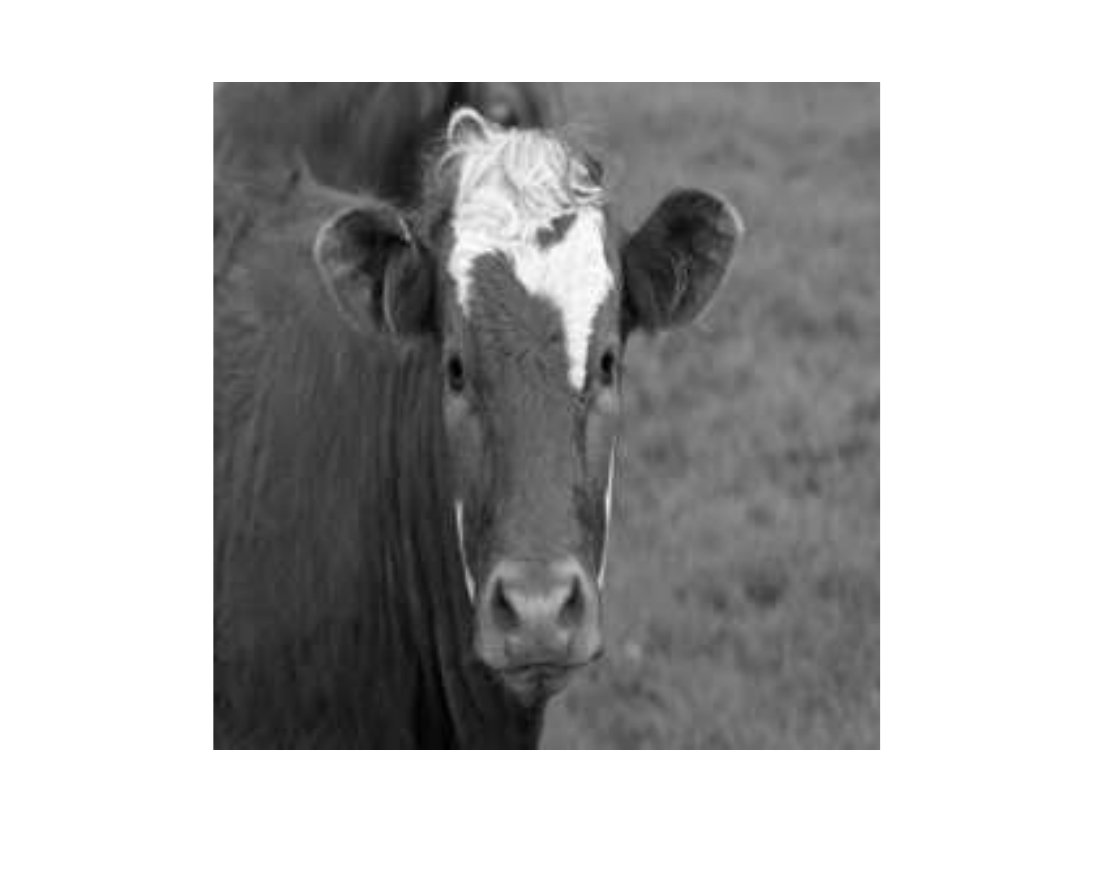}}
\subfigure{\includegraphics[width = 0.1\textwidth]{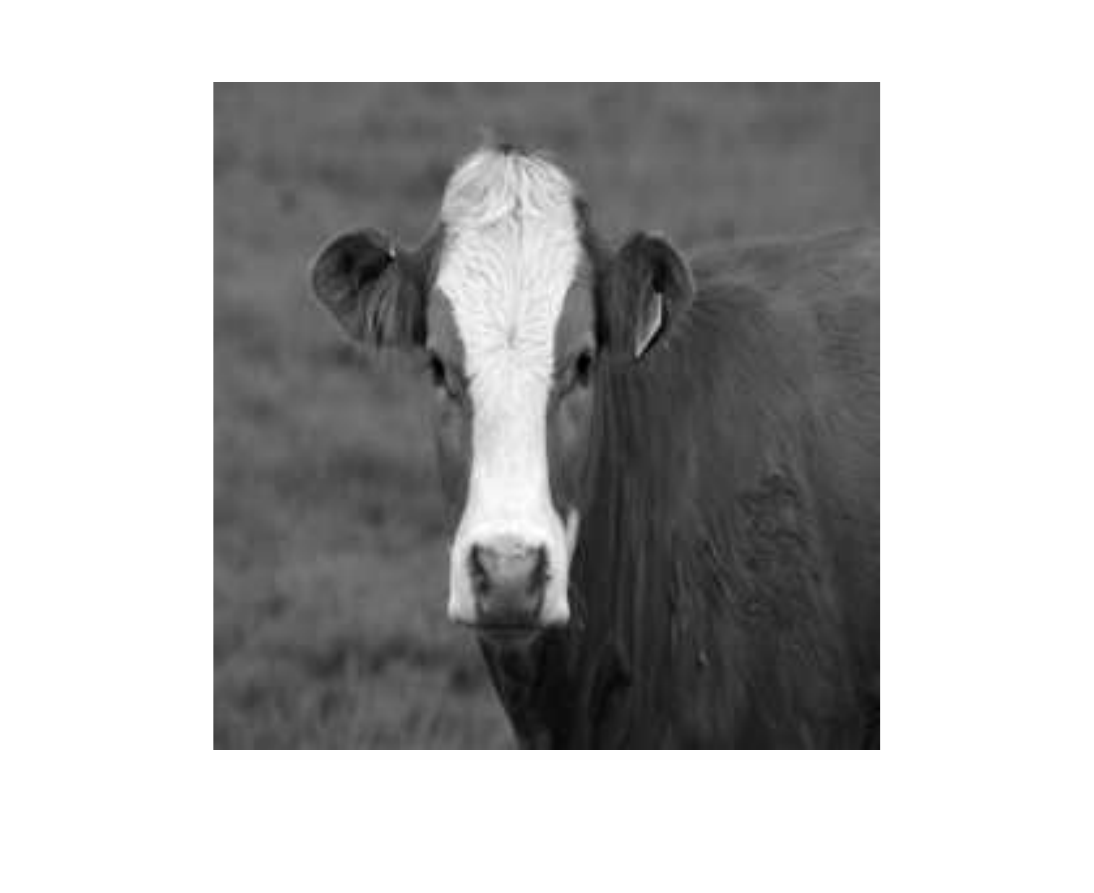}}
\subfigure{\includegraphics[width = 0.1\textwidth]{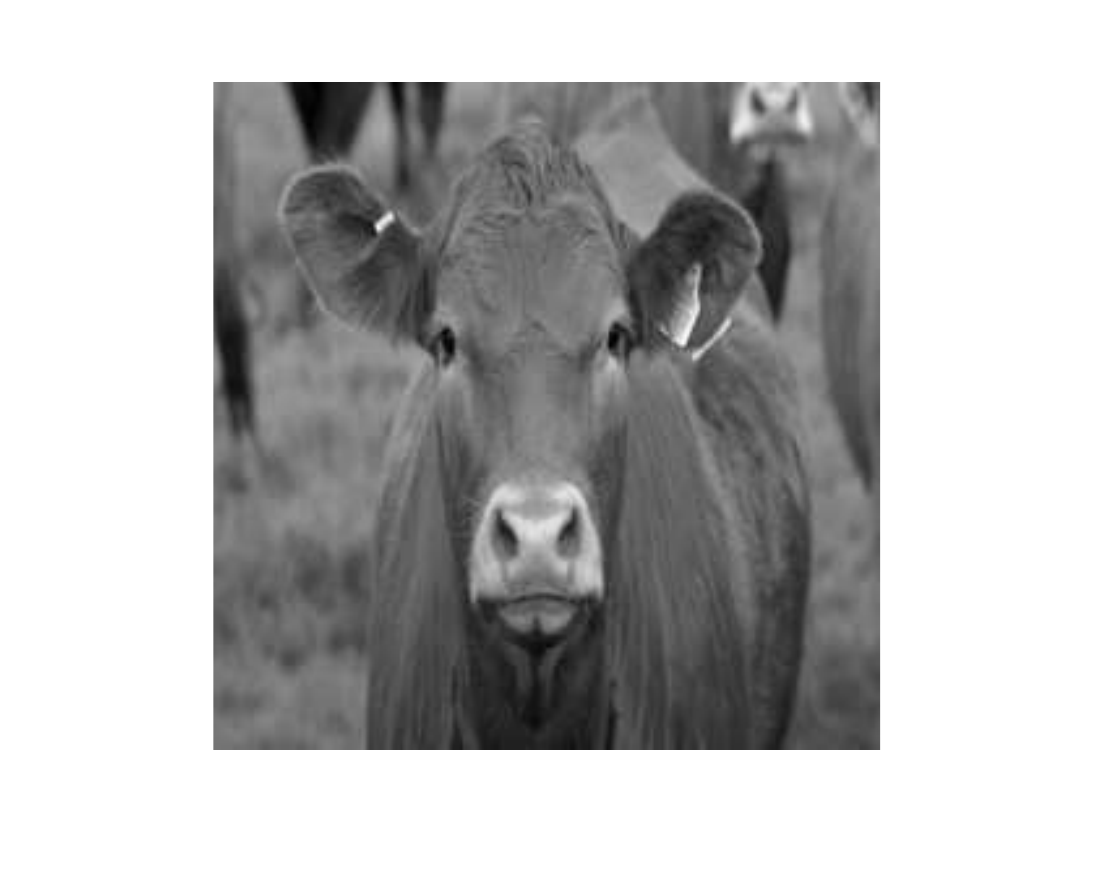}}
\subfigure{\includegraphics[width = 0.1\textwidth]{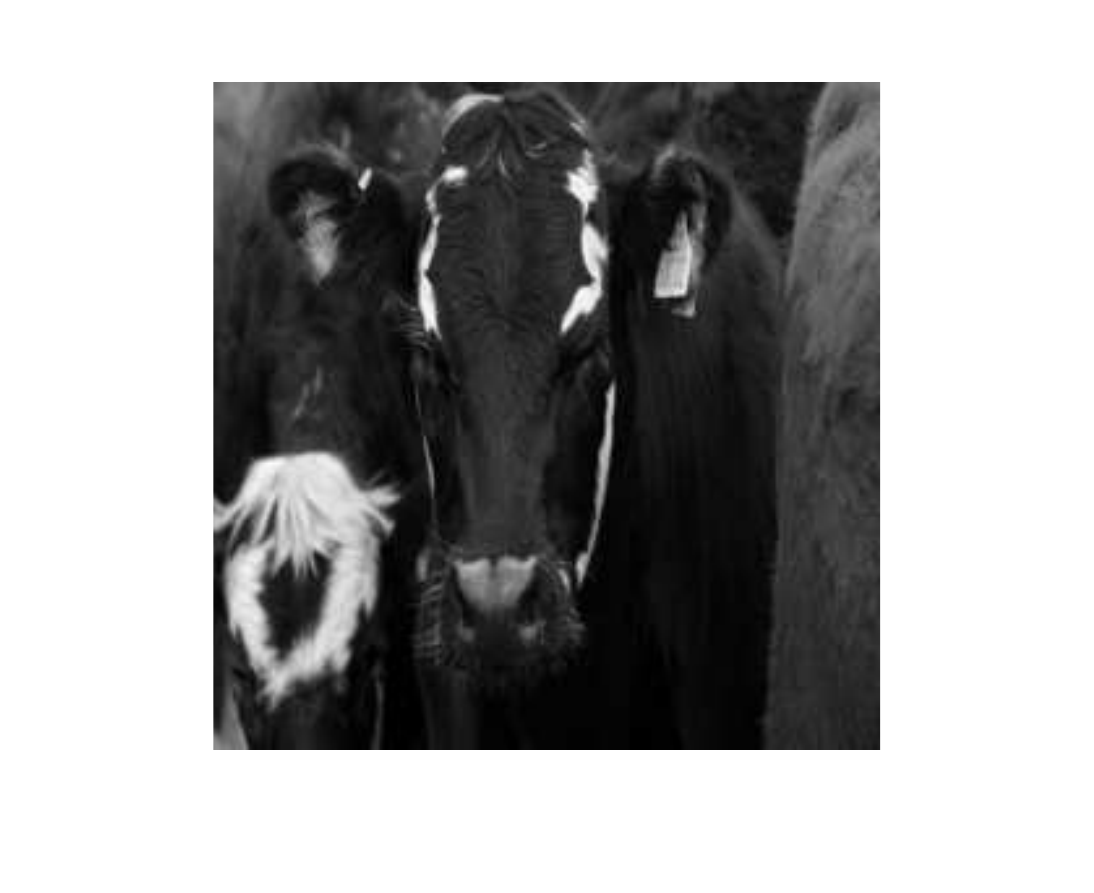}}
\subfigure{\includegraphics[width = 0.1\textwidth]{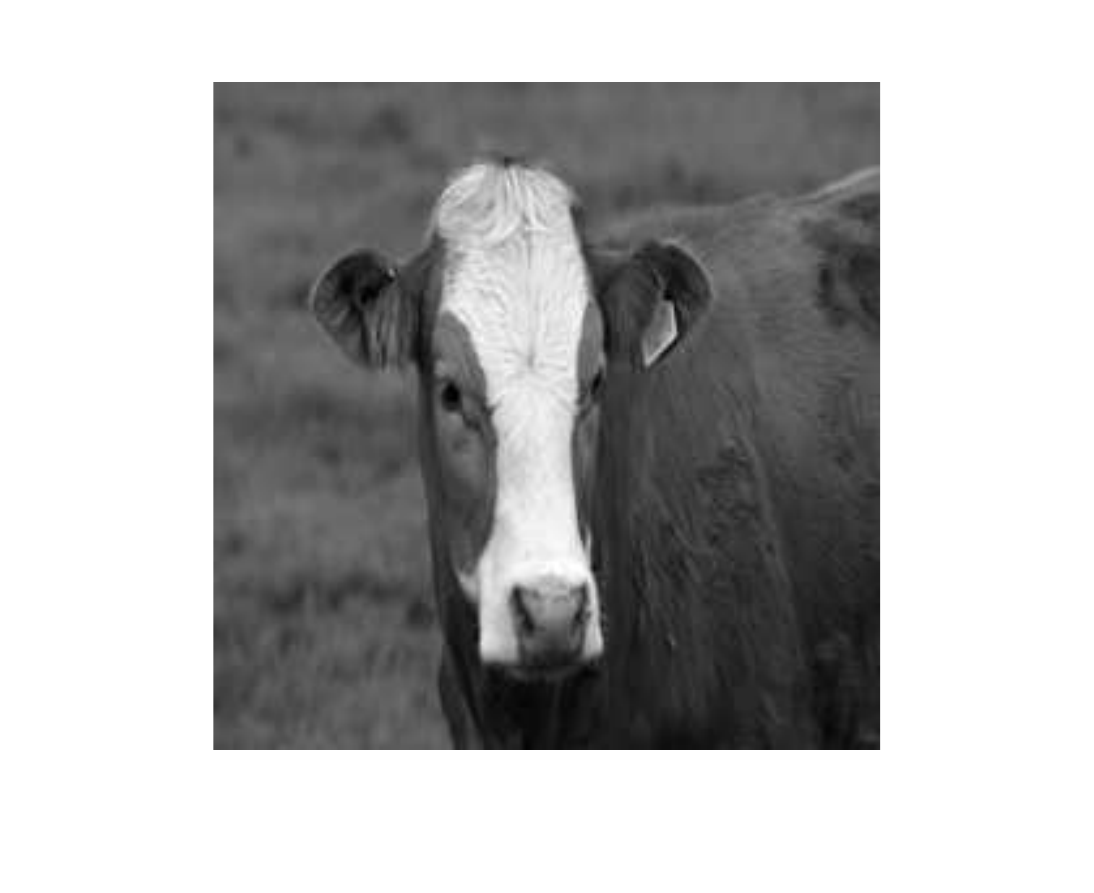}}
\subfigure{\includegraphics[width = 0.1\textwidth]{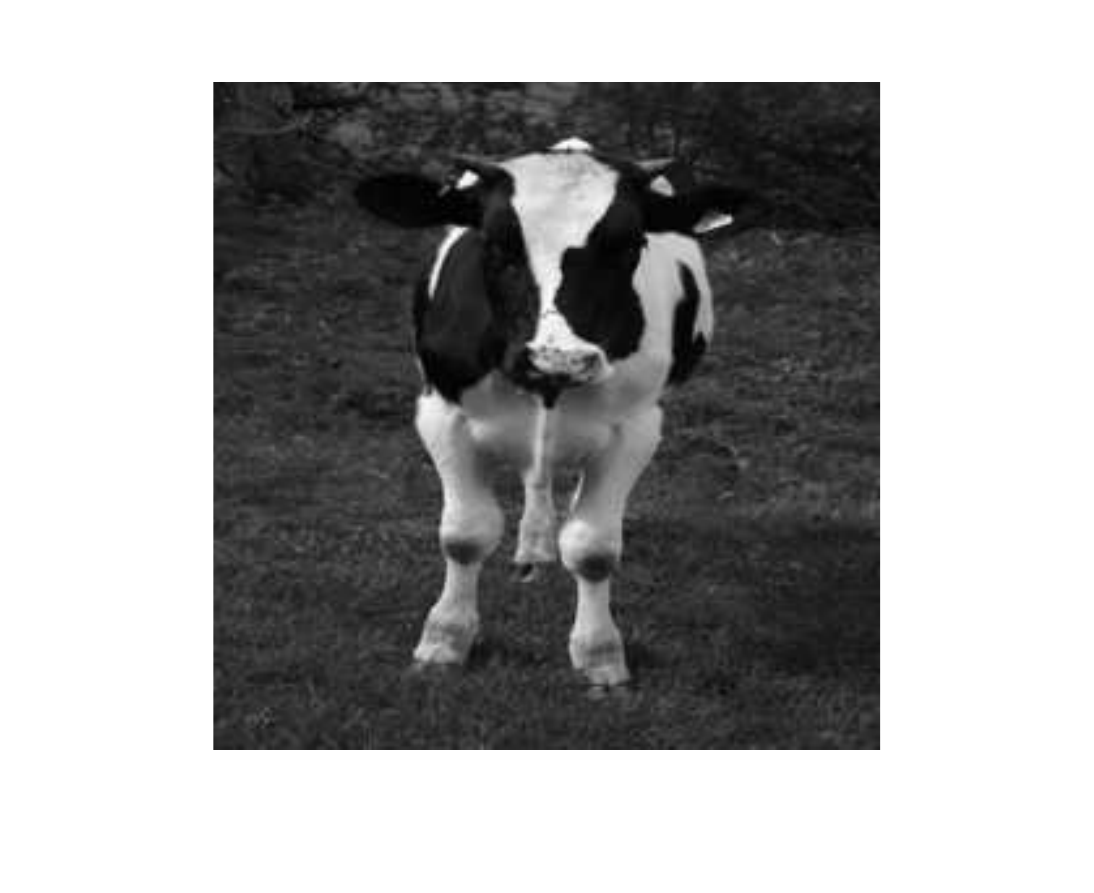}}
\subfigure{\includegraphics[width = 0.1\textwidth]{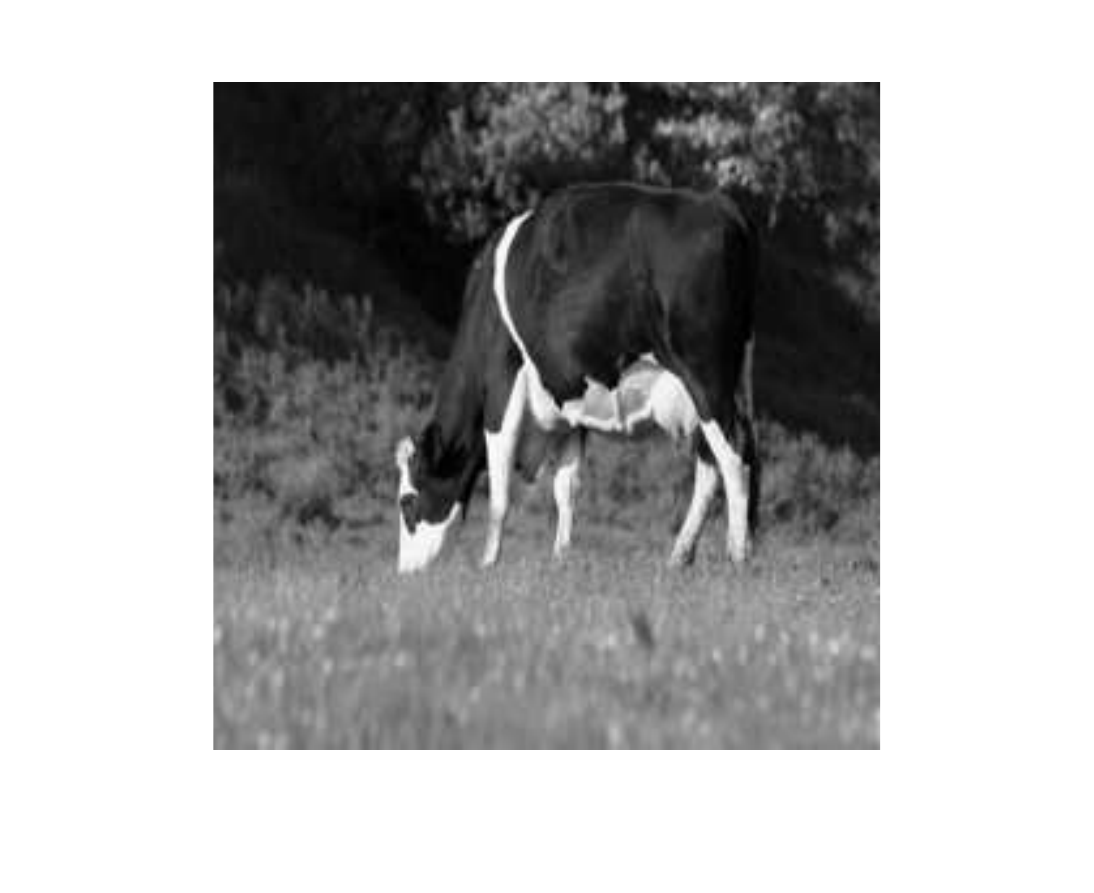}}
\subfigure{\includegraphics[width = 0.1\textwidth]{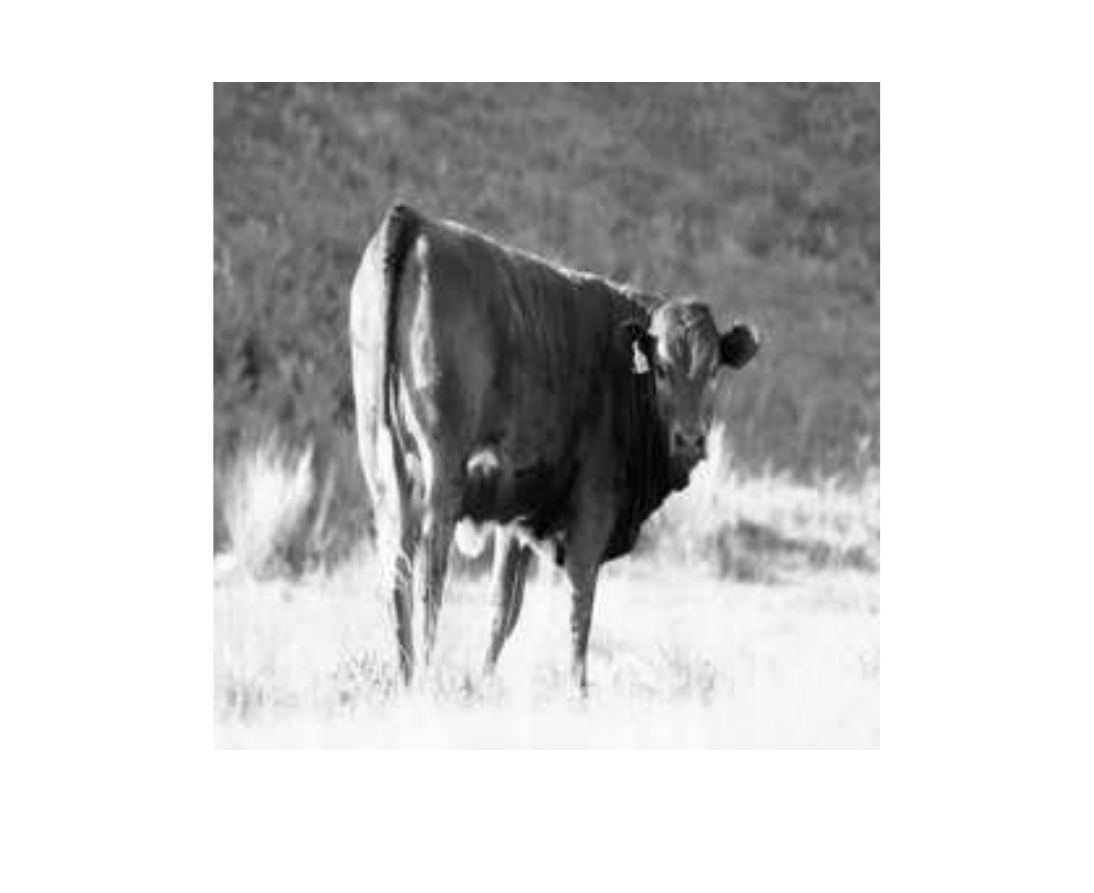}}
\subfigure{\includegraphics[width = 0.1\textwidth]{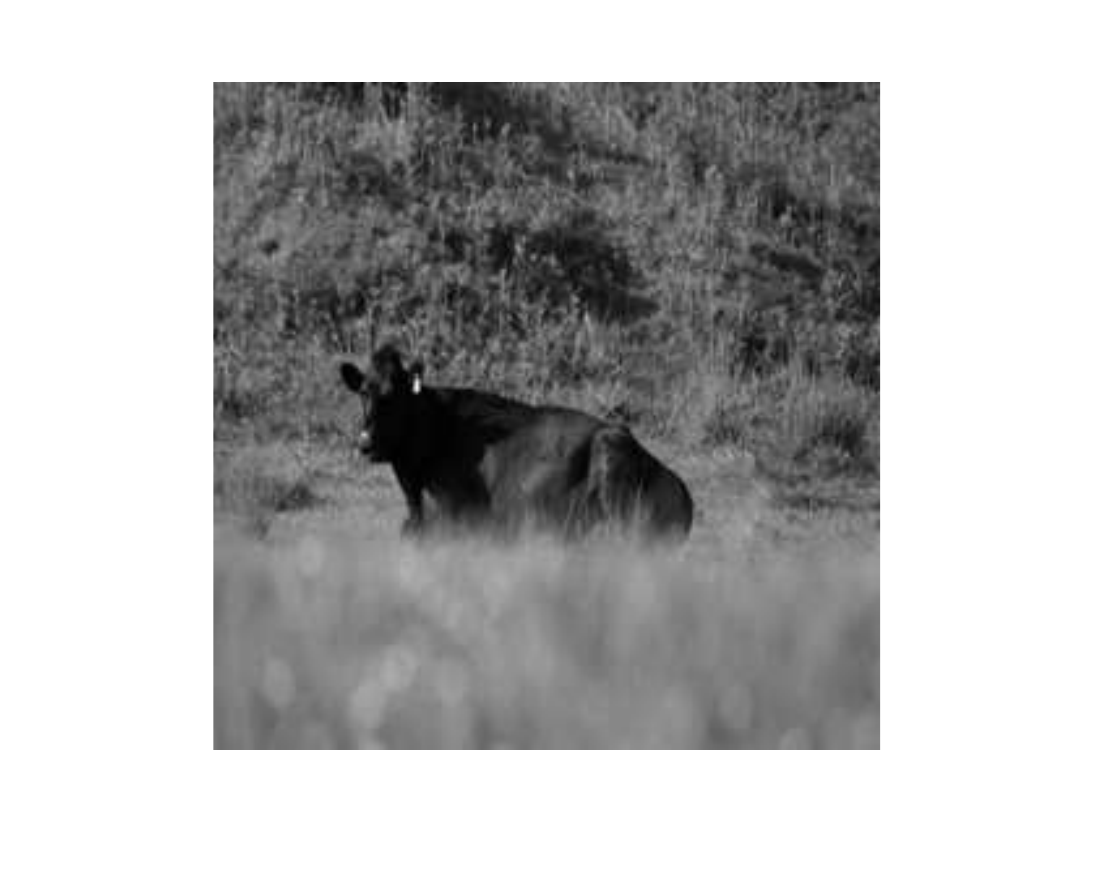}}
\subfigure{\includegraphics[width = 0.1\textwidth]{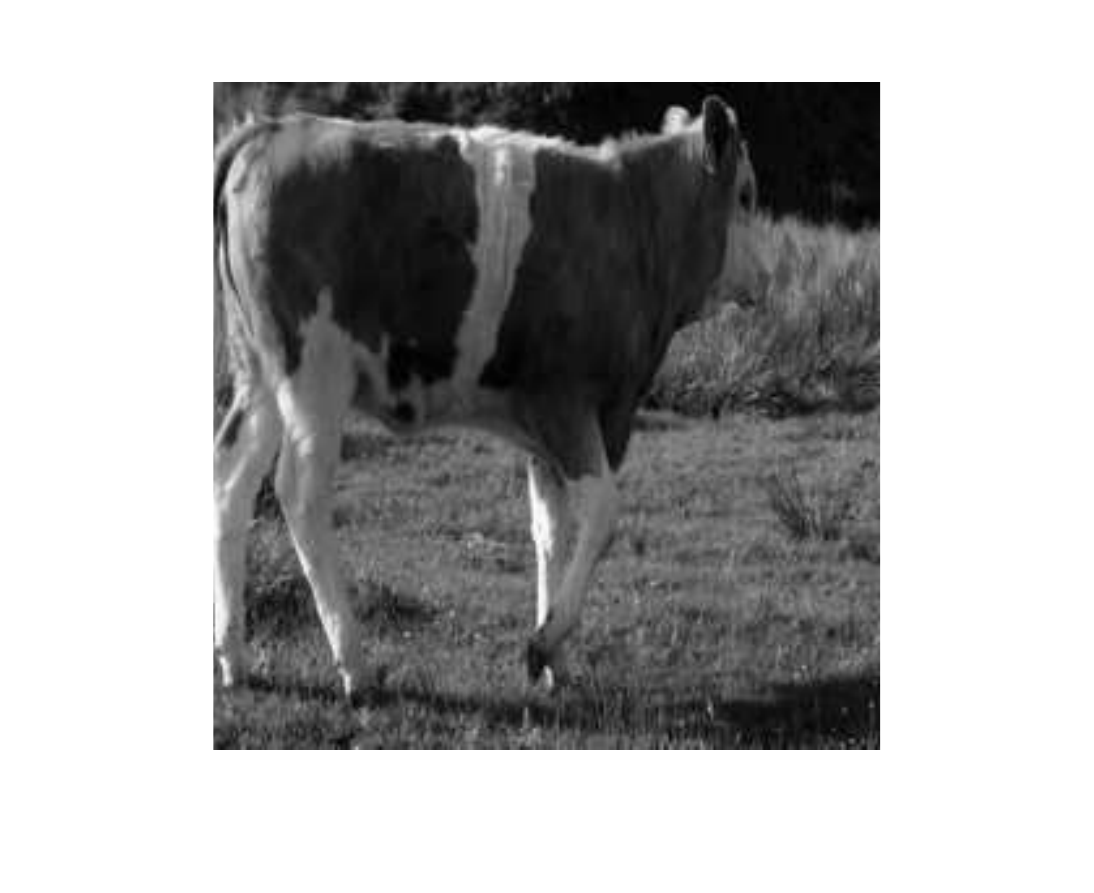}}
}
\centerline{
\subfigure{\includegraphics[width = 0.1\textwidth]{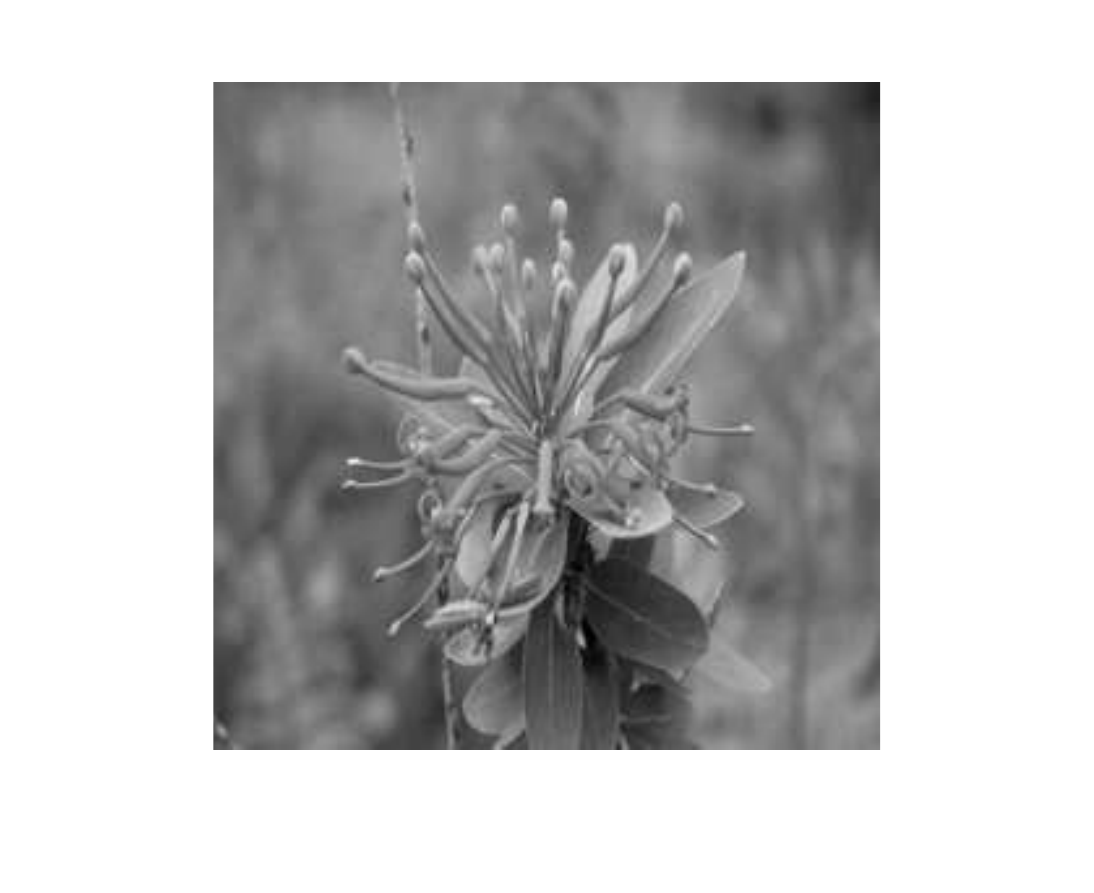}}
\subfigure{\includegraphics[width = 0.1\textwidth]{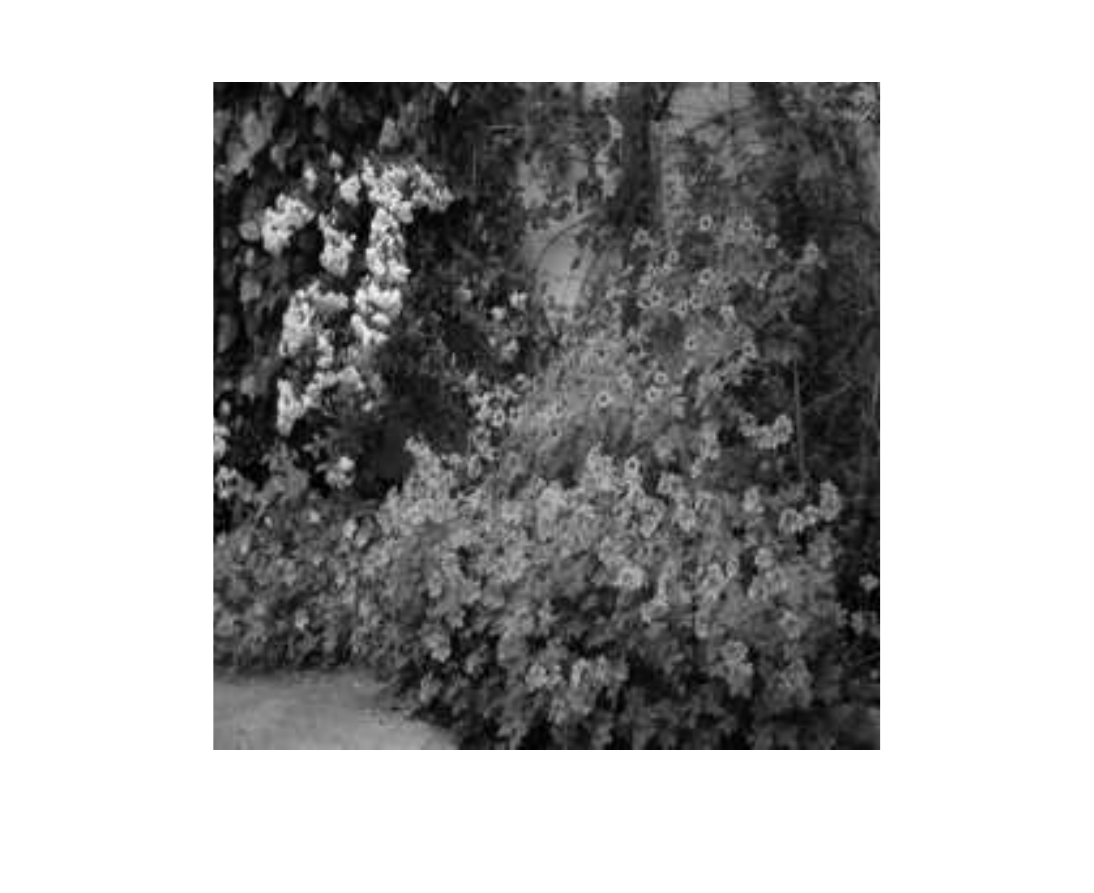}}
\subfigure{\includegraphics[width = 0.1\textwidth]{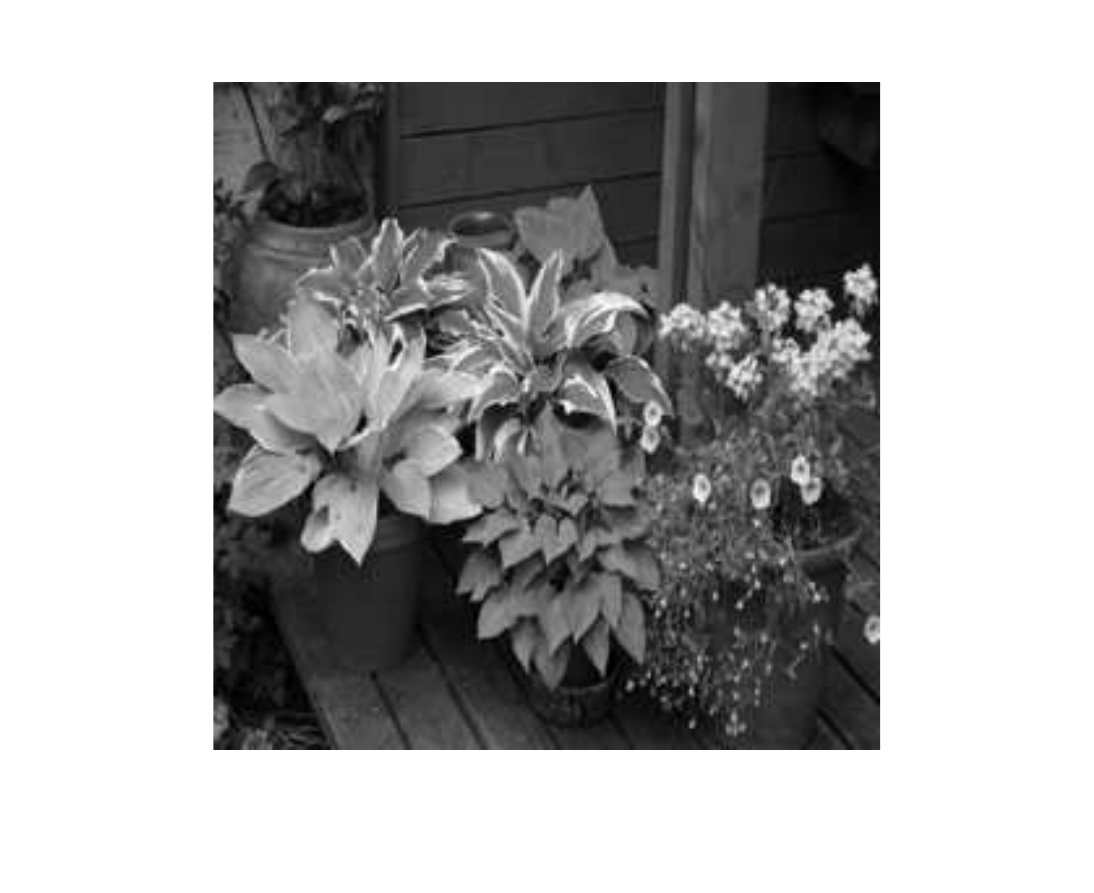}}
\subfigure{\includegraphics[width = 0.1\textwidth]{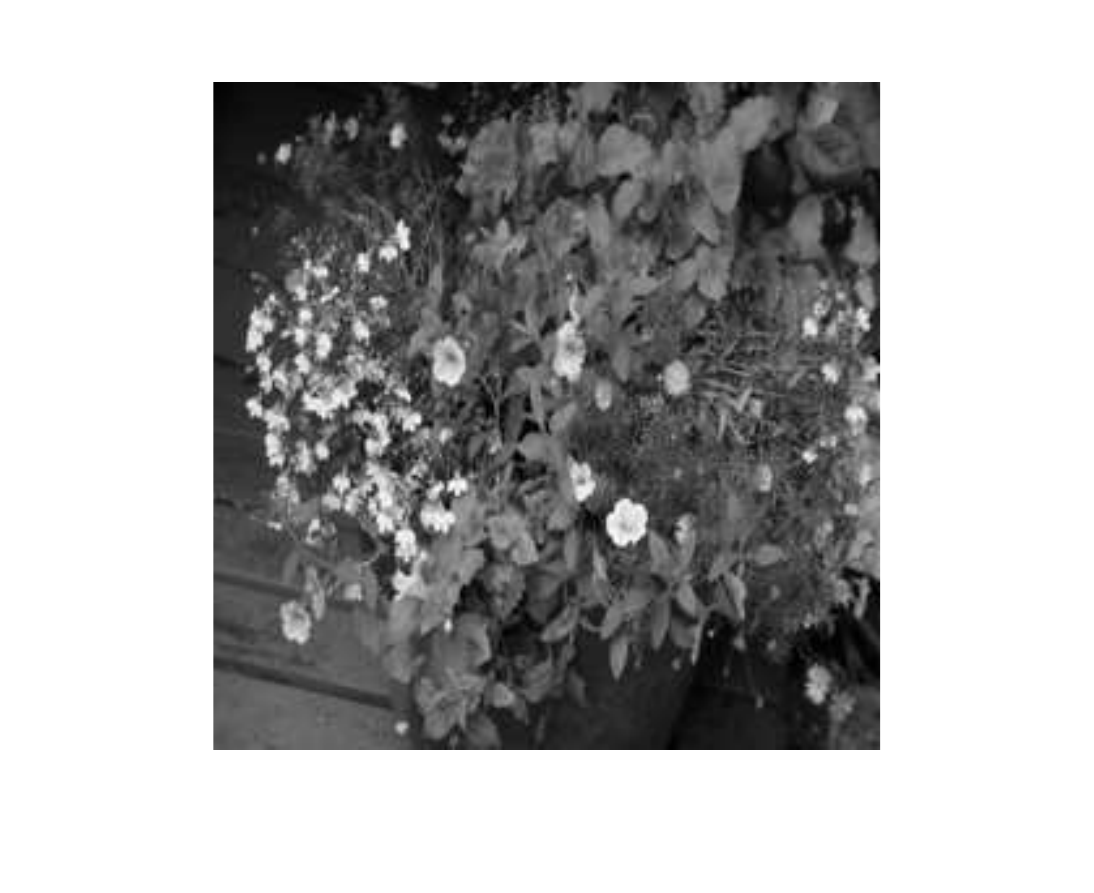}}
\subfigure{\includegraphics[width = 0.1\textwidth]{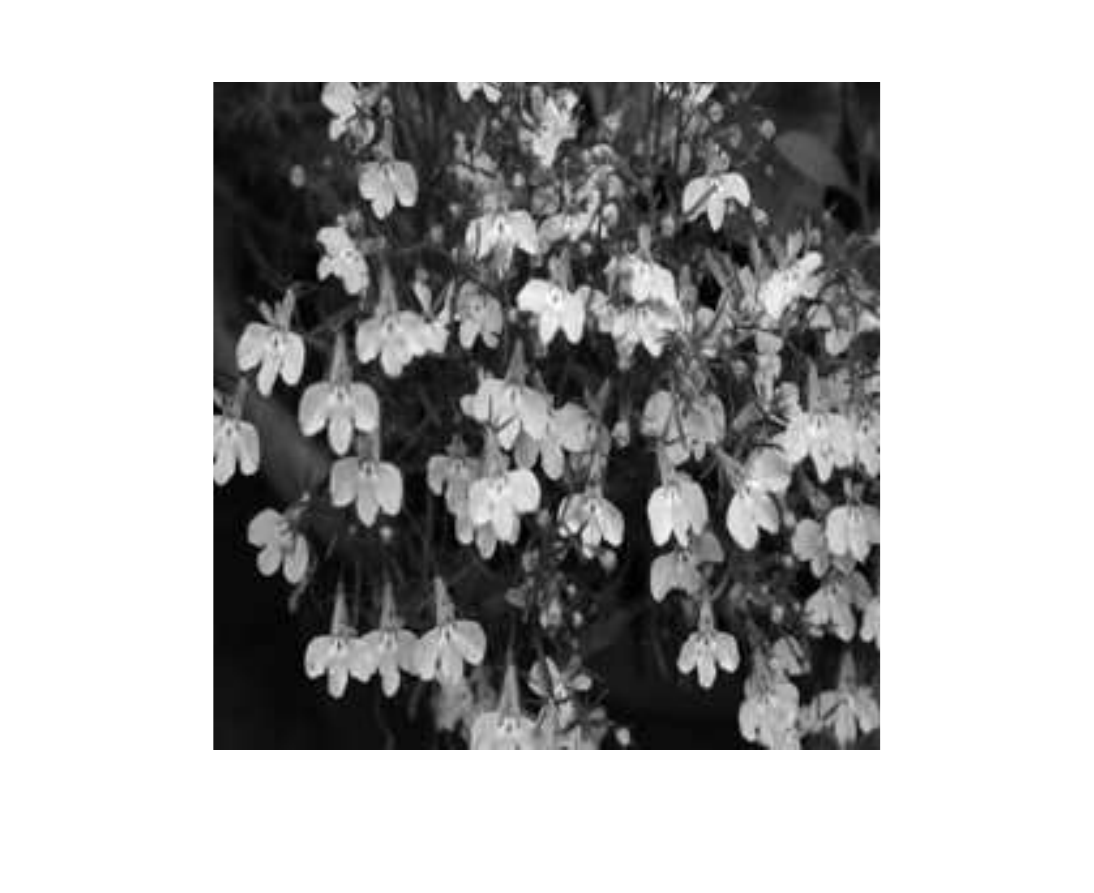}}
\subfigure{\includegraphics[width = 0.1\textwidth]{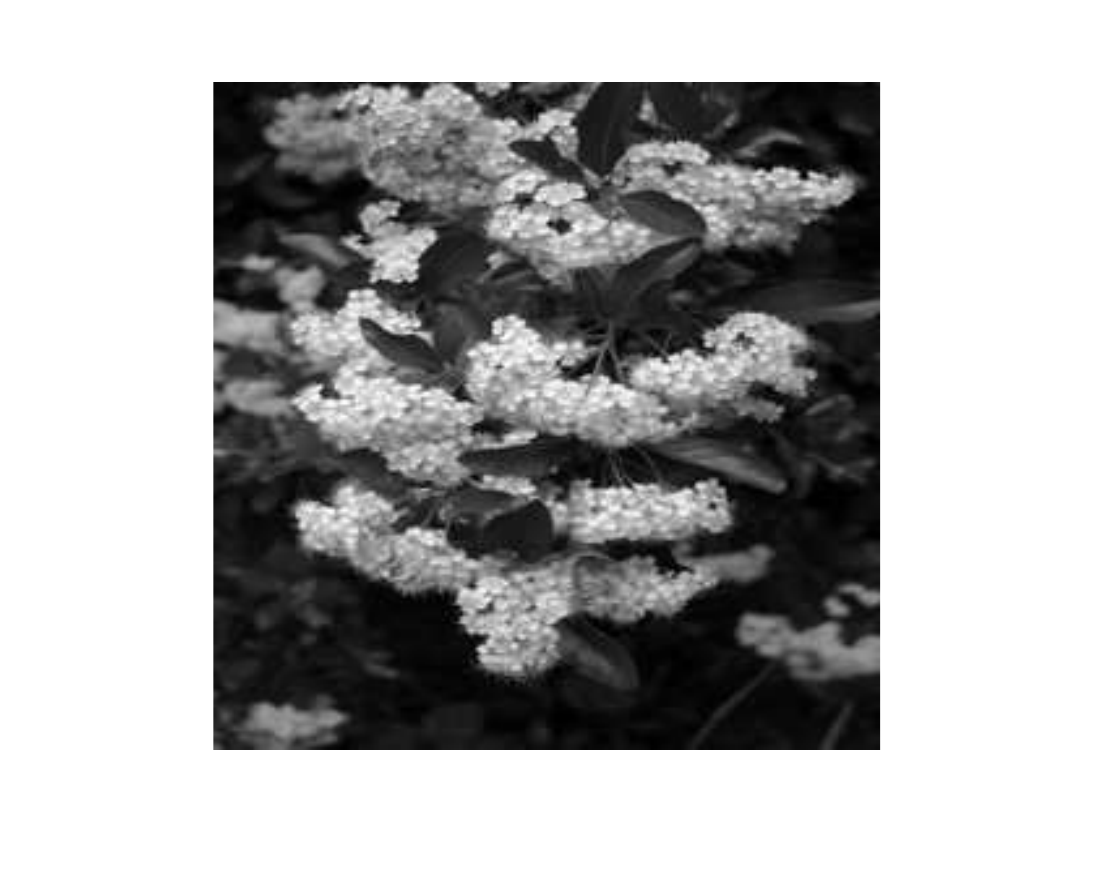}}
\subfigure{\includegraphics[width = 0.1\textwidth]{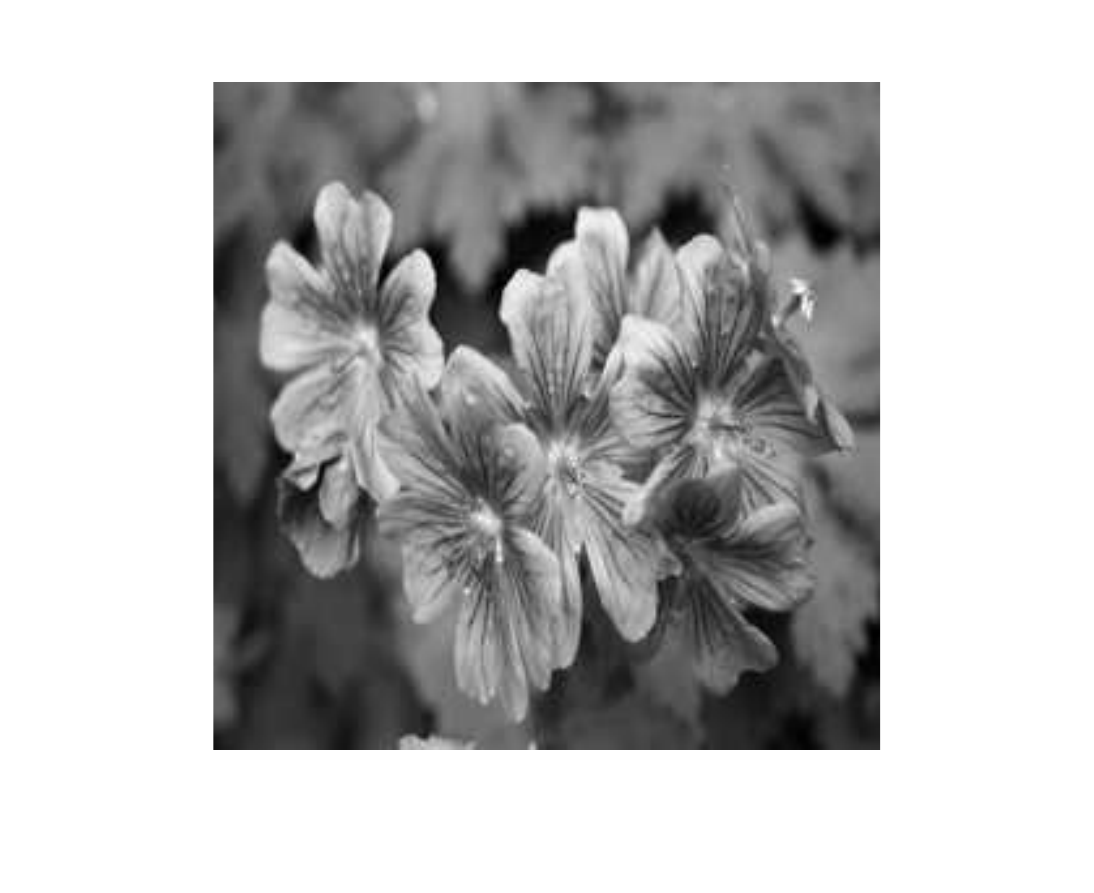}}
\subfigure{\includegraphics[width = 0.1\textwidth]{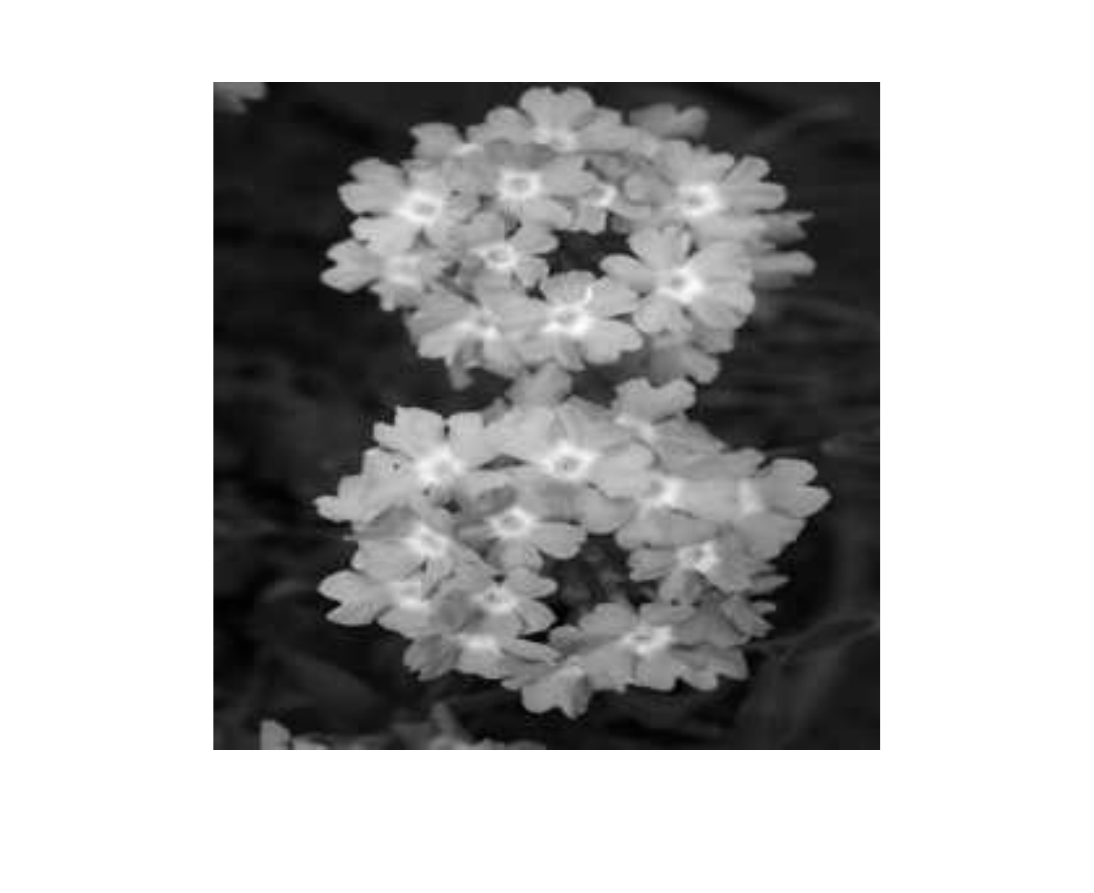}}
\subfigure{\includegraphics[width = 0.1\textwidth]{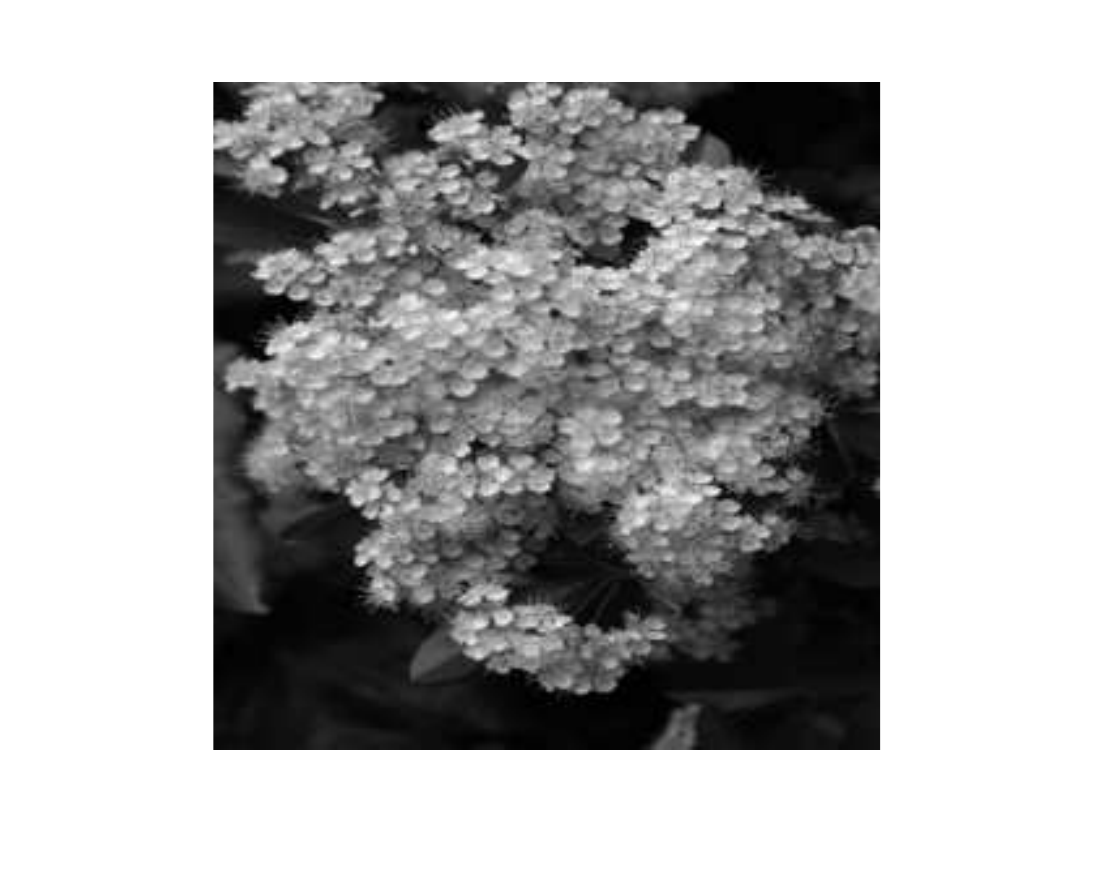}}
\subfigure{\includegraphics[width = 0.1\textwidth]{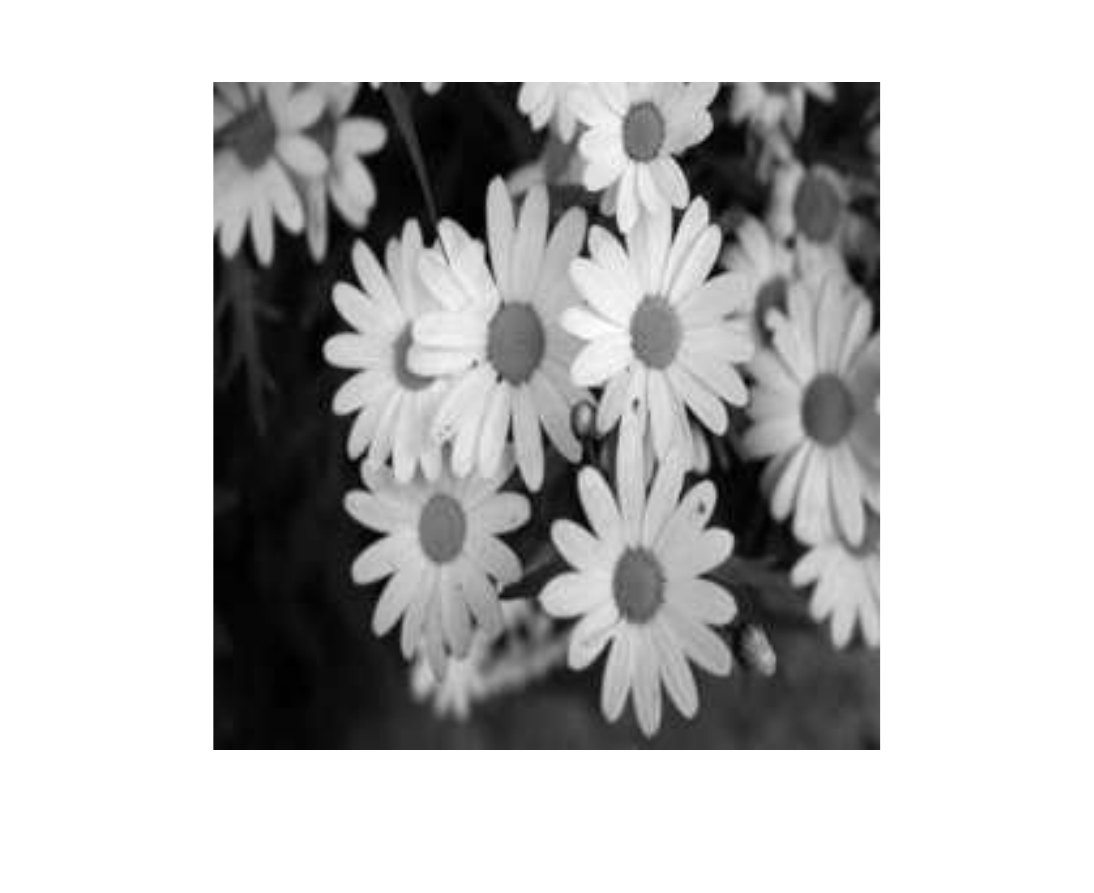}}
}
\caption{Randomly selected natural images from three different classes used for test. The first row are ``buildings'', the second row are ``cows'' and the third row are ``flowers''.}
\label{fig:TestImage}
\end{figure*}

For experiments on natural images we used the MSR Cambridge dataset \cite{MSRimage}. Ten randomly selected test images belonging to three classes of this dataset are used for experiments. The images are shown in Fig. \ref{fig:TestImage}. We have used $64\times 64$ images. Each image is divided into $8\times 8$ blocks. After reconstructing all blocks of an image in the decoder, the $NMSE$ for the reconstructed image is calculated. The task is to simultaneously encode 4 blocks ($L=4$) of an image and reconstruct them in the decoder. This means that $\mathbf{S}$ in \eqref{eq:Intro4} has 4 columns each one having $N=64$ entries. We used $50\%$ measurements, i.e., $\mathbf{Y}$ in $\eqref{eq:Intro4}$ have 4 columns each one having $M=32$ entries. 

\begin{figure}[t]
\centerline{
\subfigure{\includegraphics[width = 0.5\textwidth]{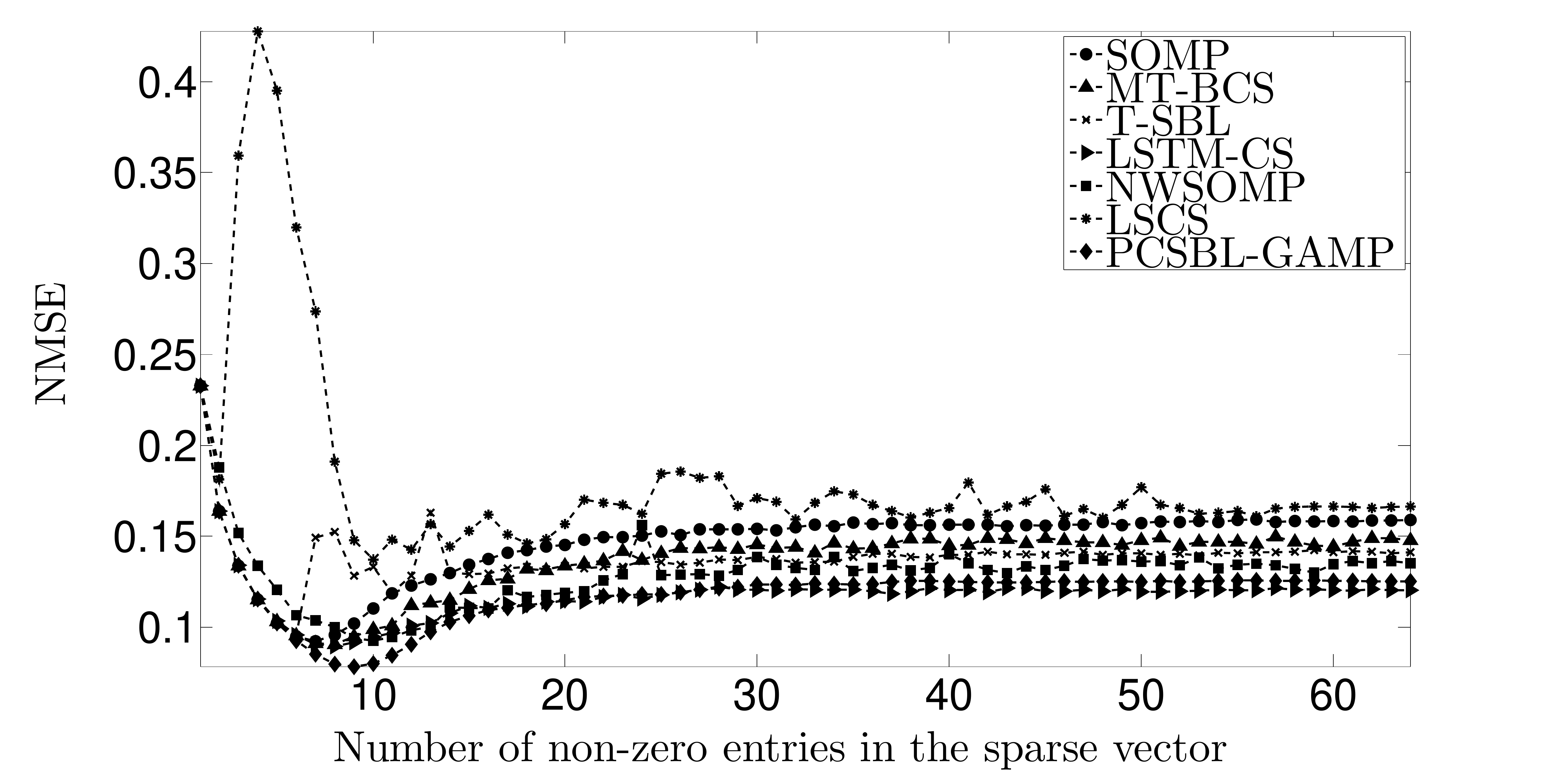}}
}
\centerline{
\subfigure{\includegraphics[width = 0.5\textwidth]{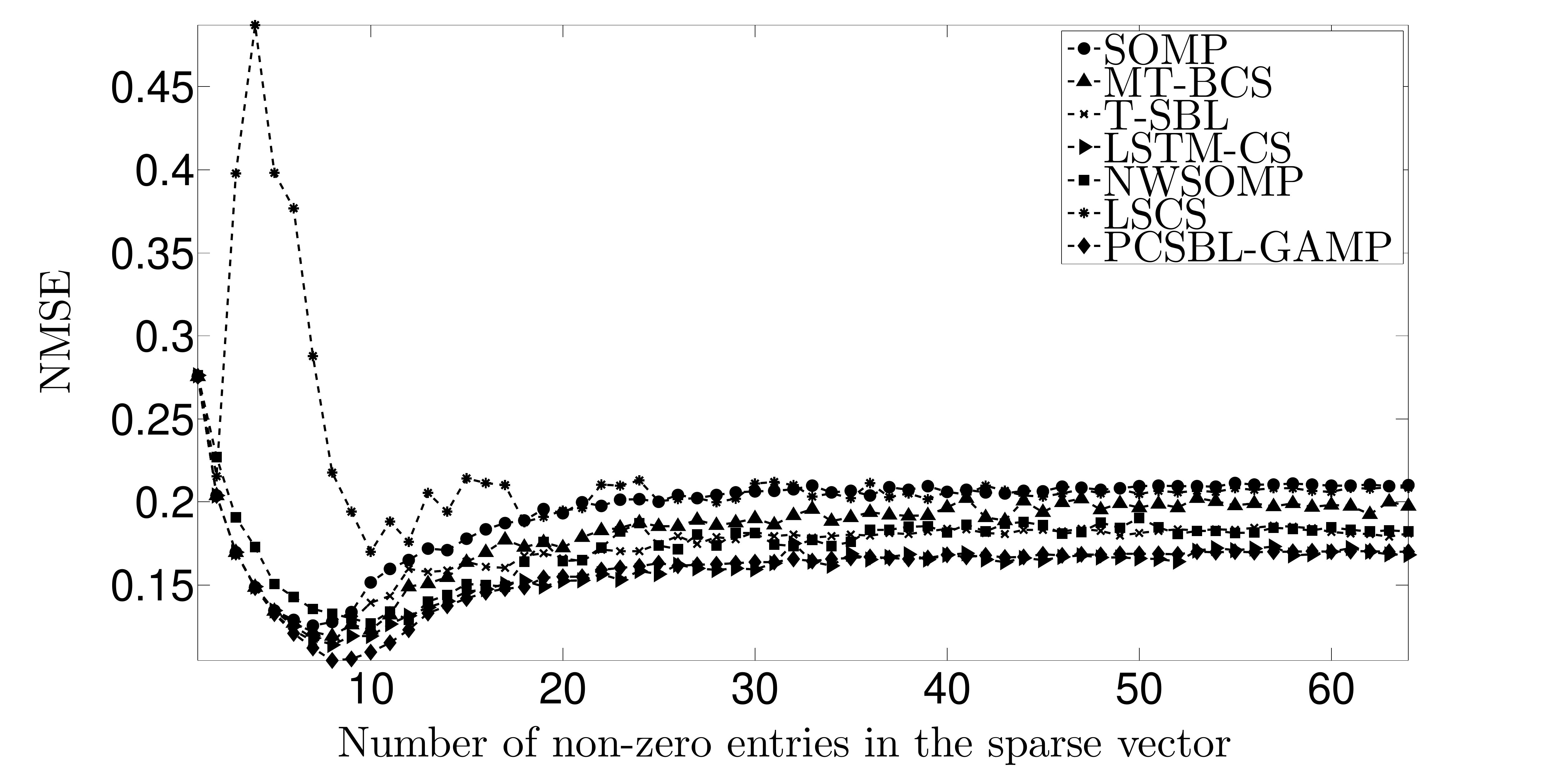}}
}
\centerline{
\subfigure{\includegraphics[width = 0.5\textwidth]{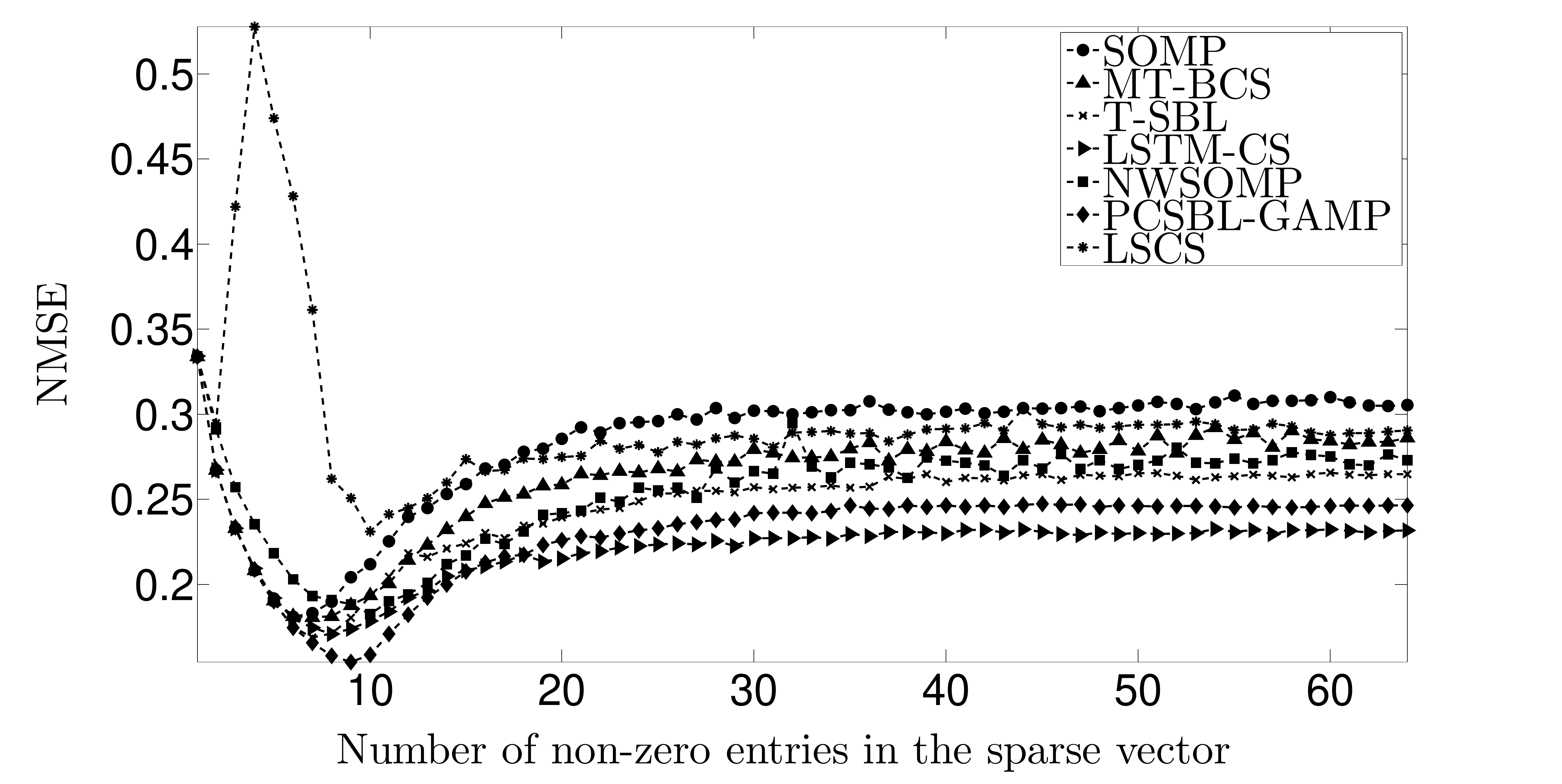}}
}
\caption{Comparison of different MMV reconstruction algorithms for natural image dataset using DCT transform and just one layer for LSTM model in LSTM-CS. Image classes from top to bottom respectively: buildings, cows and flowers.}
\label{fig:DCTresults}
\end{figure}

\begin{figure}[t]
\centerline{
\subfigure{\includegraphics[width = 0.5\textwidth]{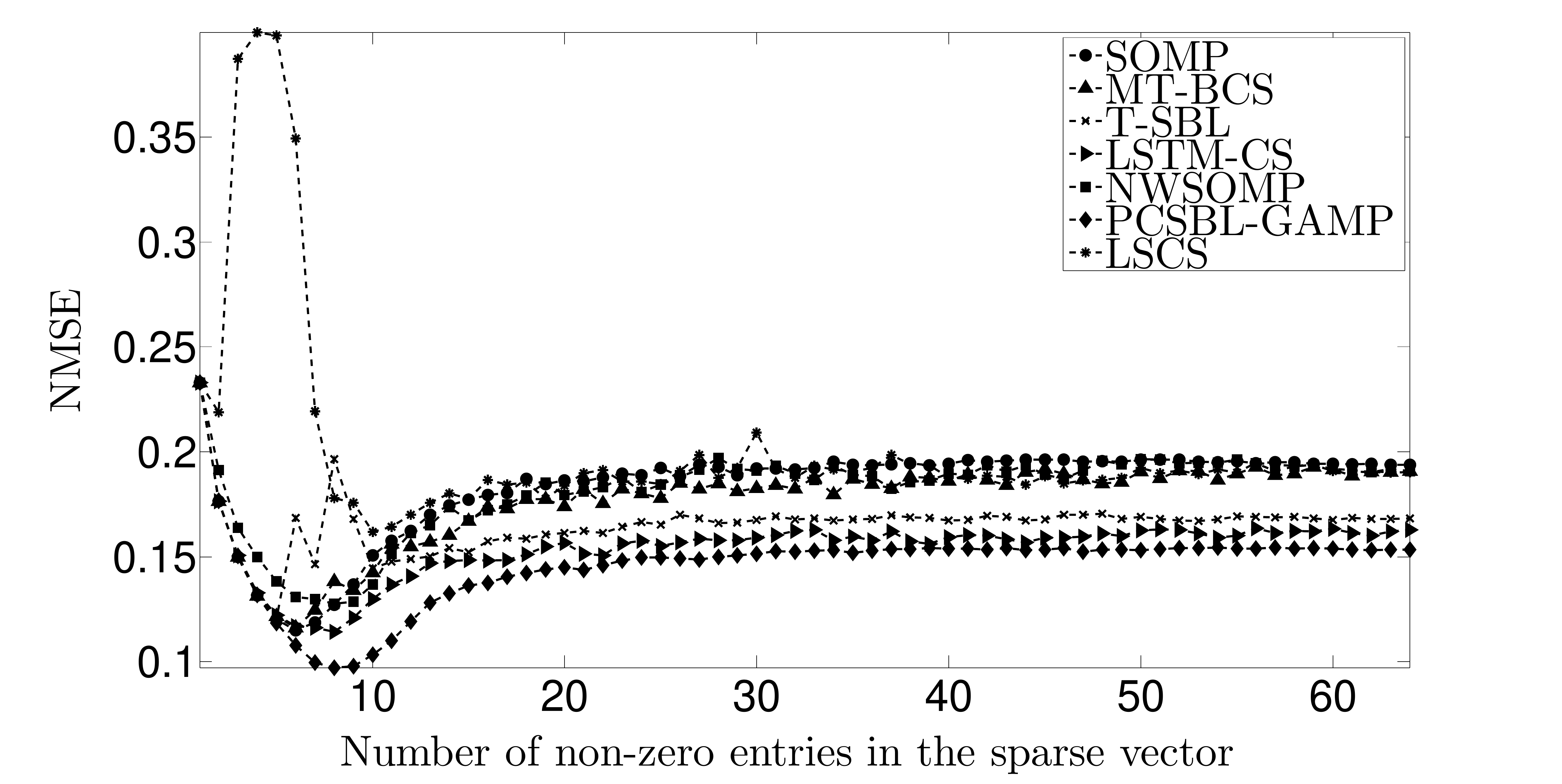}}
}
\centerline{
\subfigure{\includegraphics[width = 0.5\textwidth]{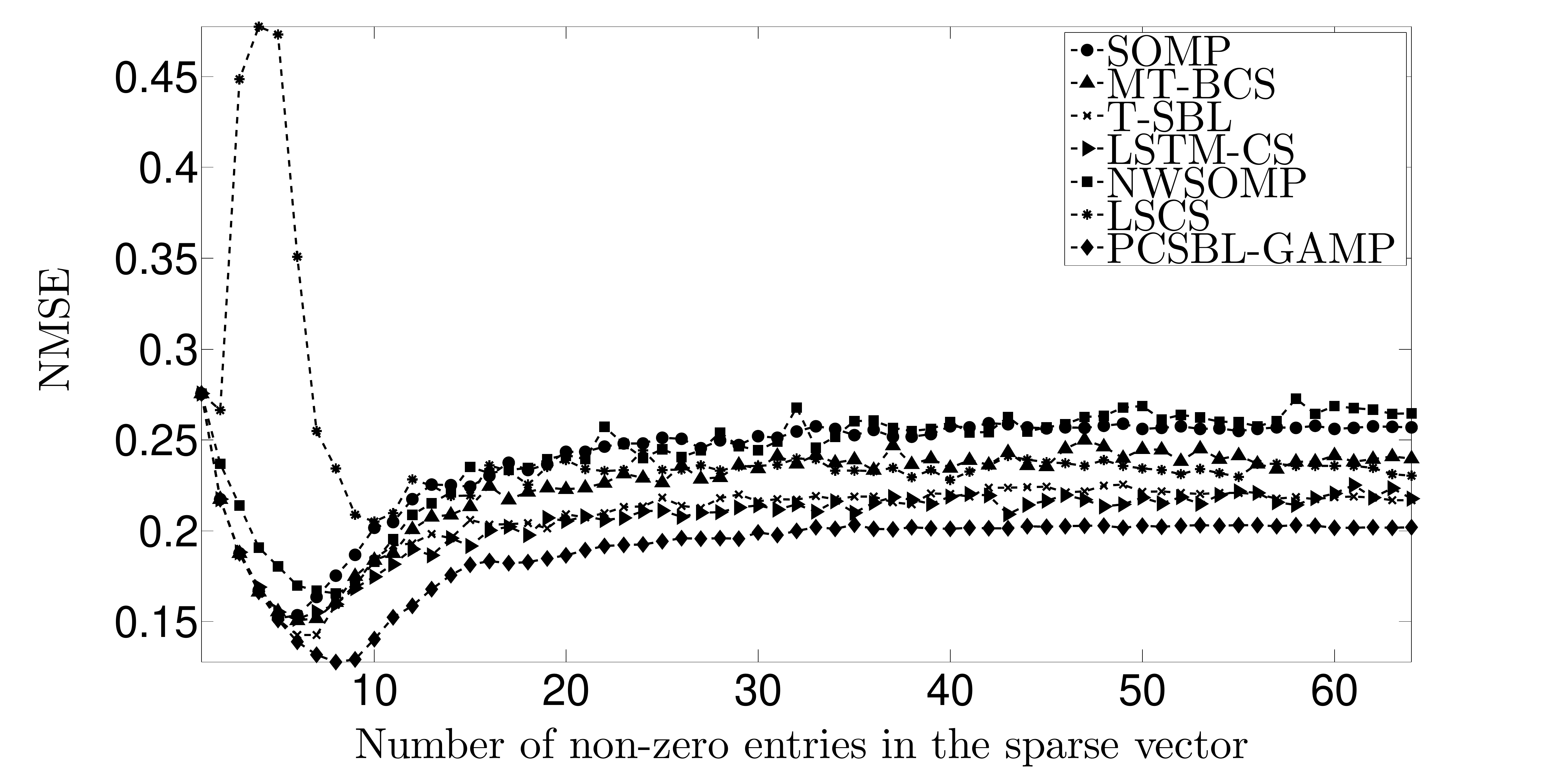}}
}
\centerline{
\subfigure{\includegraphics[width = 0.5\textwidth]{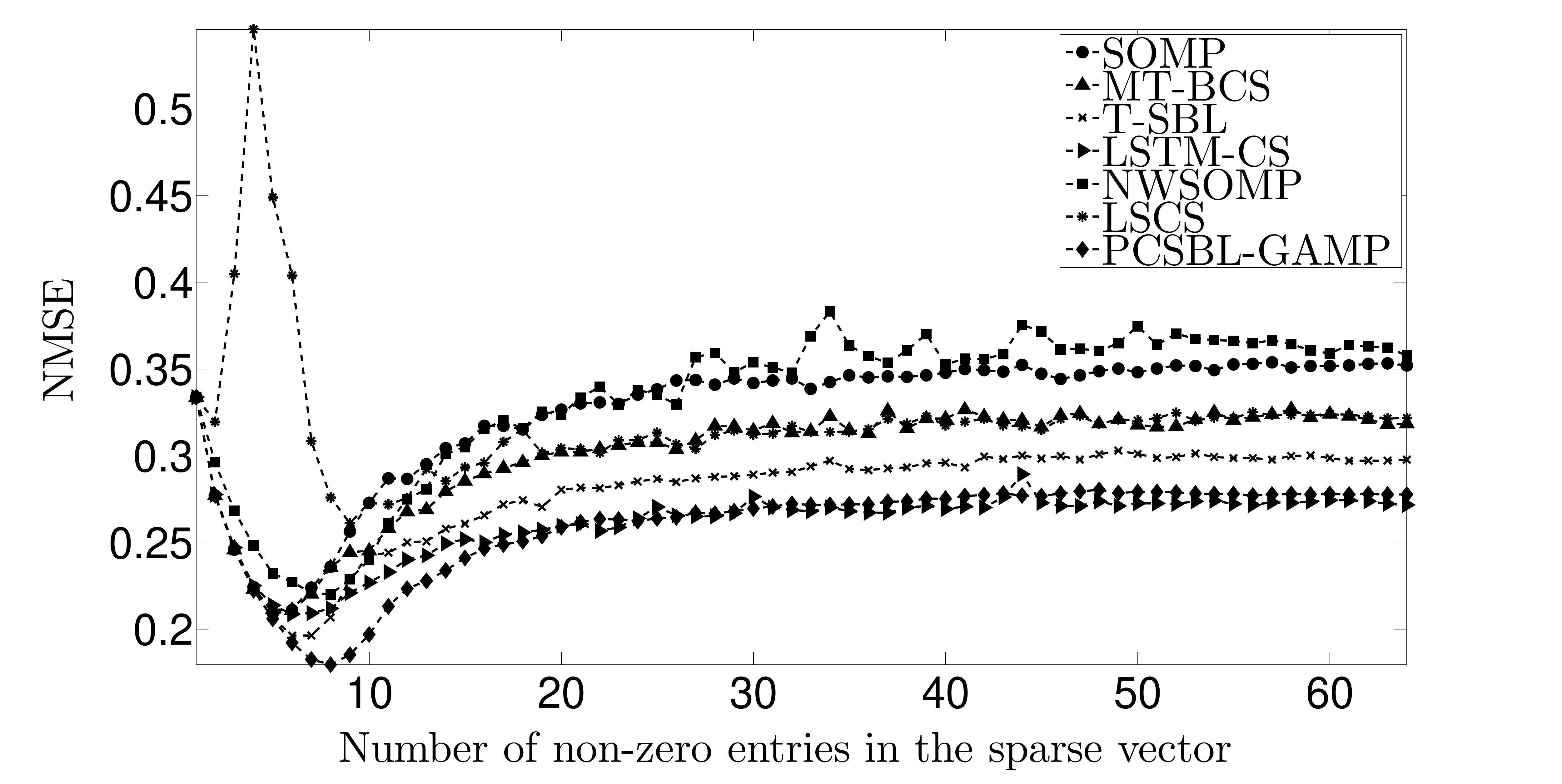}}
}
\caption{Comparison of different MMV reconstruction algorithms for natural image dataset using Wavelet transform and just one layer for LSTM model in LSTM-CS. Image classes from top to bottom respectively: buildings, cows and flowers.}
\label{fig:Waveletresults}
\end{figure}

We have compared the performance of the proposed algorithm, LSTM-CS, with SOMP, T-SBL, MT-BCS and NWSOMP. We have not included results of applying BCS per channel due its weak performance compared to other methods (this is shown in the experiments for MNIST dataset). 
We have used the same setting as the settings for the MNIST dataset for different methods which is explained in the previous section. The only differences here are: (i) For each class of images, we have used just 55 images for training set and 5 images for validation set which do not include any of 10 images used for test. (ii) We have used 15 epochs for training LSTM-CS which is enough for this dataset, compared to 25 epochs for the MNIST dataset. The experiments were performed for two popular transforms, DCT and Wavelet, for all aforementioned reconstruction algorithms. For the wavelet transform we used Haar wavelet transform with 3 levels of decomposition. Results for DCT transform are presented in Fig. \ref{fig:DCTresults}. Results for wavelet transform are presented in Fig. \ref{fig:Waveletresults}. 

\begin{figure}[t]
\centerline{
\subfigure{\includegraphics[width = 0.5\textwidth]{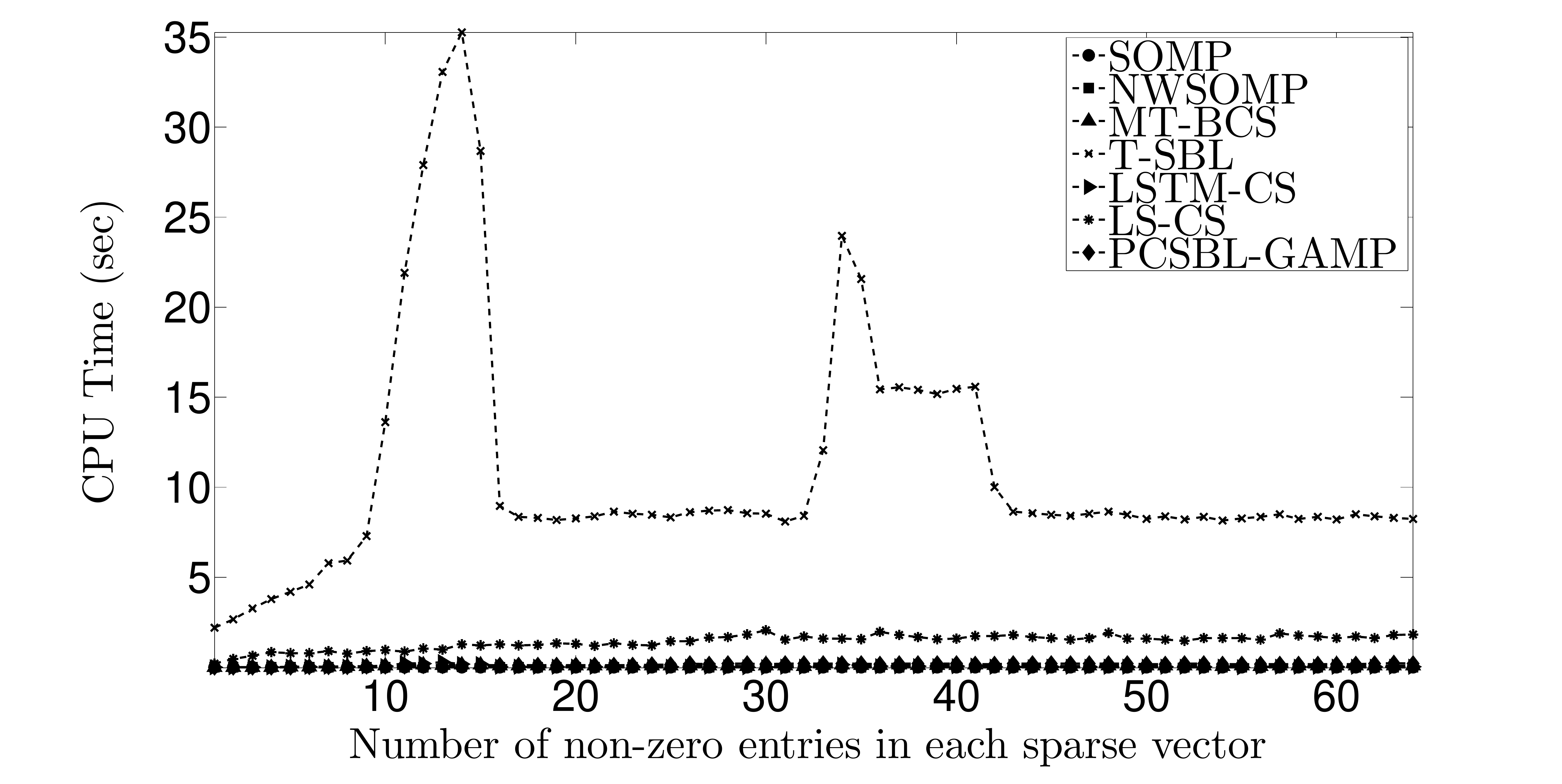}}
}
\centerline{
\subfigure{\includegraphics[width = 0.5\textwidth]{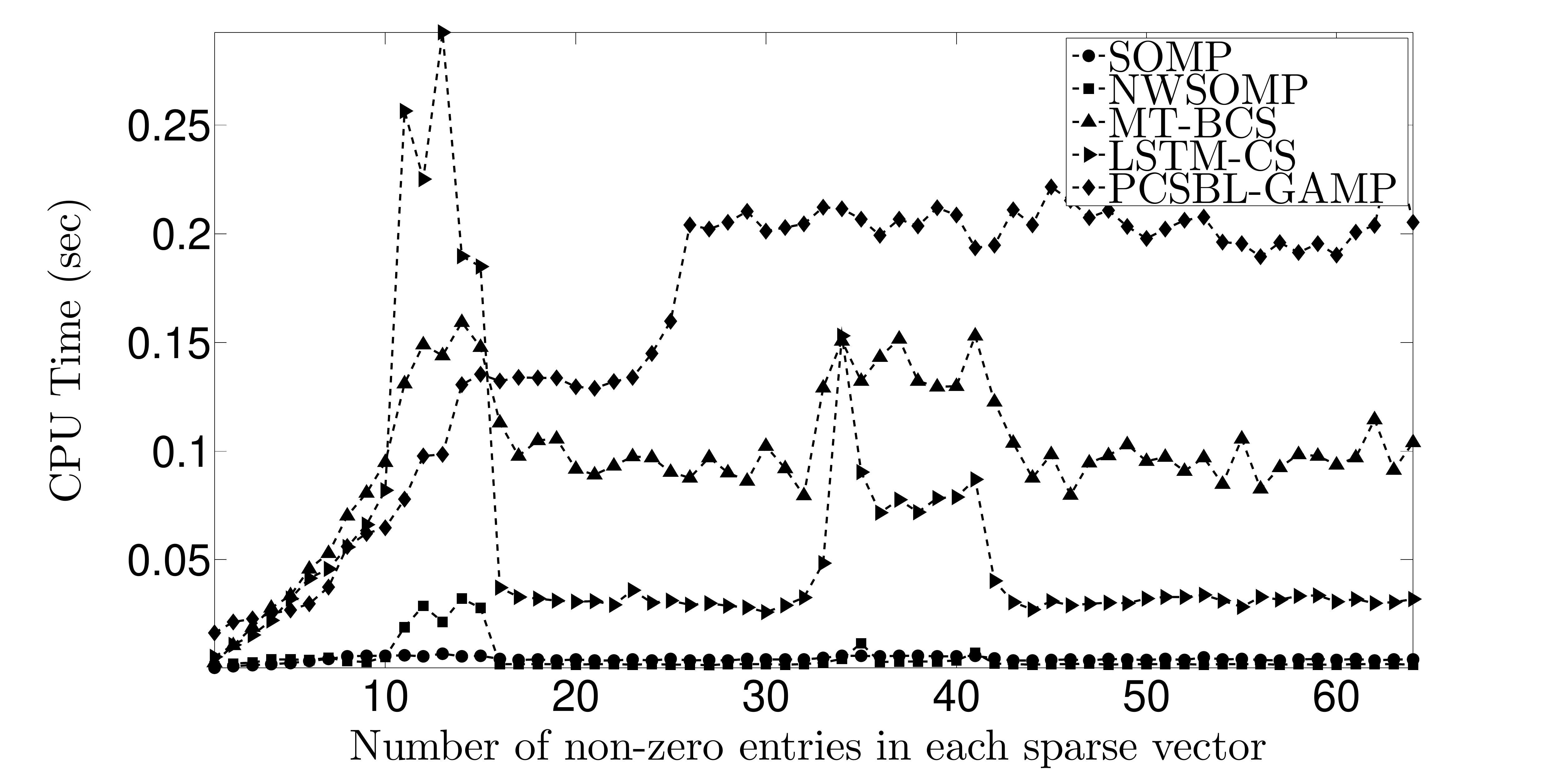}}
}
\caption{CPU time for different MMV reconstruction algorithms. These times are for the experiment using DCT transform for 10 test images from the building class. The bottom figure is the same as top figure but without T-SBL and LS-CS to make the difference among different methods more clear.}
\label{fig:CPUtime}
\end{figure}

To conclude the experiments section, the CPU time for different reconstruction algorithms for the MMV problem discussed in this paper are presented in Fig. \ref{fig:CPUtime}. Each point on the curves in Fig. \ref{fig:CPUtime} is the time spent to reconstruct each sparse vector averaged over all the $8\times 8$ blocks in 10 test images. We observe from this figure that the proposed algorithm is almost as fast as greedy algorithms. Please note that there is a faster version of T-SBL that is known as TMSBL. It will improve the CPU time of T-SBL but it is still slower than other reconstruction methods. 

\section{Conclusions and Future Work}
\label{sec:conclusion}
This paper presents a method to reconstruct sparse vectors for the MMV problem. The proposed method learns the structure of sparse vectors and does not rely on the commonly used joint sparsity assumption. Through experiments on two real world datasets, we showed that the proposed method outperforms the general MMV baseline SOMP as well as a number of Bayesian model based methods for the MMV problem. Please note that we have not used multiple layers of LSTM or the advanced deep learning methods for training, e.g., regularization using drop out which can improve the performance of LSTM-CS. This paper is a proof of concept that deep learning methods and specifically sequence modelling methods, e.g., LSTM, can improve the performance of the MMV solvers significantly. This is specially the case when the sparsity patterns are more complicated than that of obtained by the DCT or Wavelet transforms. We showed this on the MNIST dataset. Please note that if collecting training samples is expensive or enough training samples are not available, using other sparse reconstruction methods is recommended. Our future work includes:  1) Extending the LSTM-CS to bidirectional LSTM-CS. 2) Extending the proposed method to non-linear distributed compressive sensing. 3) Using the proposed method for video compressive sensing where there is correlation amongst the video frames, and compressive sensing of EEG signals where there is correlation amongst the different EEG channels. 

\section{Acknowledgement}
\label{sec:acknowledge}
We want to thank the authors of \cite{Rao3}, \cite{Carin1}, \cite{Carin2},\cite{LSCS} and \cite{AfterRev2} for making the code of their work available. This was important in performing comparisons. 
For reproducibility of the results, please contact the authors for the MATLAB codes of the proposed LSTM-CS method. We also want to thank WestGrid and Compute Canada Calcul Canada for providing computational resources for part of this work.

\appendices
\section{Expressions for the Gradients}
\label{sec:appGradientExpress}
In this appendix we present the final gradient expressions that are necessary to use for training the proposed model for the MMV problem. Due to lack of space, we omit the presentation of full derivations of these gradients.

Starting with the cost function in \eqref{eq:cost}, we use the Nesterov method described in \eqref{eq:Nesterov} to update LSTM-CS model parameters. Here, $\mathbf{\Lambda}$ is one of the weight matrices or bias vectors $\{\mathbf{W}_1,\mathbf{W}_2,\mathbf{W}_3,\mathbf{W}_4,\mathbf{W}_{rec1},\mathbf{W}_{rec2},\mathbf{W}_{rec3},\mathbf{W}_{rec4}$\\ $,\mathbf{W}_{p1}, \mathbf{W}_{p2}, \mathbf{W}_{p3}, \mathbf{b}_1, \mathbf{b}_2, \mathbf{b}_3, \mathbf{b}_4\}$ in the LSTM-CS architecture. The general format of the gradient of the cost function,  $\nabla L(\mathbf{\Lambda})$, is the same as \eqref{eq:costGrad}. To calculate $\frac{\partial L_{r,i,\tau}(\mathbf{\Lambda})}{\partial\mathbf{\Lambda}}$ from \eqref{eq:cost} we have:
\begin{equation}
\label{eq:appendix1}
\frac{\partial L_{r,i,\tau}(\mathbf{\Lambda})}{\partial\mathbf{\Lambda}} = -\sum_{j=1}^N s_{0,r,i,\tau}(j)\frac{\partial log(s_{r,i,\tau}(j))}{\partial\mathbf{\Lambda}}
\end{equation}
After a straightforward derivation of derivatives we will have:
\begin{equation}
\label{eq:appendix2}
\frac{\partial L_{r,i,\tau}(\mathbf{\Lambda})}{\partial\mathbf{\Lambda}} = ( \beta \mathbf{s}_{r,i,\tau} - \mathbf{s}_{0,r,i,\tau} )\frac{\partial \mathbf{z}_{\tau}}{\partial \mathbf{\Lambda}}
\end{equation}
where $\mathbf{z}_{\tau}$ is the vector $\mathbf{z}$ for $\tau$-th channel in Fig. \ref{Fig:BlockDiagram} and $\beta$ is a scalar defined as:
\begin{equation}
\label{eq:appendix3}
\beta = \sum_{j=1}^N s_{0,r,i,\tau}(j)
\end{equation}
Since during training data generation we have generated one hot vectors for $\mathbf{s}_0$, $\beta$ always equals to 1. Since we are looking at different channels as a sequence, for a more clear presentation we show any vector corresponding to $t$-th channel with $(t)$ instead of index $\tau$. For example, $\mathbf{z}_{\tau}$ is represented by $\mathbf{z}(t)$. 

Since $\mathbf{z}(t) = \mathbf{U}\mathbf{v}(t)$ we have:
\begin{equation}
\label{eq:appendix4}
\frac{\partial \mathbf{z}(t)}{\partial \mathbf{\Lambda}} = \mathbf{U}^T \frac{\partial \mathbf{v}(t)}{\partial \mathbf{\Lambda}}
\end{equation}
Combining \eqref{eq:appendix2}, \eqref{eq:appendix3} and \eqref{eq:appendix4} we will have:
\begin{equation}
\label{eq:appendix5}
\frac{\partial L_{r,i,t}(\mathbf{\Lambda})}{\partial\mathbf{\Lambda}} = \mathbf{U}^T( \mathbf{s}_{r,i}(t) - \mathbf{s}_{0,r,i}(t) )\frac{\partial \mathbf{v}(t)}{\partial \mathbf{\Lambda}}
\end{equation}
Starting from ``$t=L$''-th channel, we define $\mathbf{e}(t)$ as:
\begin{equation}
\label{eq:appendix6}
\mathbf{e}(t) = \mathbf{U}^T( \mathbf{s}_{r,i}(t) - \mathbf{s}_{0,r,i}(t) )
\end{equation}
The expressions for the gradients for different parameters of LSTM-CS model are presented in the subsequent sections. We omit the subscripts $r$ and $i$ for simplicity of presentation. Please note that the final value of the gradient is sum of gradient values over the mini-batch samples and number of channels as represented by summations in \eqref{eq:costGrad}.

\subsection{Output Weights $\mathbf{U}$}
\label{sec:U}
\begin{equation}
\label{eq:updateU}
\frac{\partial L_t}{\partial\mathbf{U}} = ( \mathbf{s}(t) - \mathbf{s}_0(t) ). \mathbf{v}(t)^T
\end{equation}
\subsection{Output Gate}
\label{sec:o_g}
For recurrent connections we have:
\begin{equation}
\label{eq:LSTM8c}
\frac{\partial L_t}{\partial\mathbf{W}_{rec1}} = \mathbf{\delta}^{rec1}(t).\mathbf{v}(t-1)^T
\end{equation}
where
\begin{equation}
\label{eq:LSTM9c}
\mathbf{\delta}^{rec1}(t) = \mathbf{o}(t)\circ(1-\mathbf{o}(t))\circ h(\mathbf{c}(t))\circ \mathbf{e}(t)
\end{equation}
For input connections, $\mathbf{W}_1$, and peephole connections, $\mathbf{W}_{p1}$, we will have:
\begin{equation}
\label{eq:LSTM10}
\frac{\partial L_t}{\partial\mathbf{W}_1} = \mathbf{\delta}^{rec1}(t).\mathbf{r}(t)^T
\end{equation}
\begin{equation}
\label{eq:LSTM11}
\frac{\partial L_t}{\partial\mathbf{W}_{p1}} = \mathbf{\delta}^{rec1}(t).\mathbf{c}(t)^T
\end{equation}
The derivative for output gate bias values will be:
\begin{equation}
\label{eq:LSTM11Bias}
\frac{\partial L_t}{\partial\mathbf{b}_{1}} = \mathbf{\delta}^{rec1}(t)
\end{equation}

\subsection{Input Gate}
\label{sec:o_i}
For the recurrent connections we have:
\begin{align}
\label{eq:LSTM20c}
&\frac{\partial L_t}{\partial \mathbf{W}_{rec3}} = diag(\mathbf{\delta}^{rec3}(t)).\frac{\partial \mathbf{c}(t)}{\partial \mathbf{W}_{rec3}}
\end{align}
where
\begin{align}
\label{eq:LSTM21c}
&\mathbf{\delta}^{rec3}(t) = (1-h(\mathbf{c}(t)))\circ (1+h(\mathbf{c}(t)))\circ \mathbf{o}(t) \circ \mathbf{e}(t)\nn\\
&\frac{\partial \mathbf{c}(t)}{\partial \mathbf{W}_{rec3}} = diag(\mathbf{f}(t)).\frac{\partial \mathbf{c}(t-1)}{\partial \mathbf{W}_{rec3}} + \mathbf{b}_{i}(t).\mathbf{v}(t-1)^T\nn\\
&\mathbf{b}_{i}(t) = \mathbf{y}_{g}(t)\circ \mathbf{i}(t) \circ (1-\mathbf{i}(t))
\end{align}
For the input connections we will have the following:
\begin{align}
\label{eq:LSTM22c}
&\frac{\partial L_t}{\partial \mathbf{W}_{3}} = diag(\mathbf{\delta}^{rec3}(t)).\frac{\partial \mathbf{c}(t)}{\partial \mathbf{W}_{3}}
\end{align}
where
\begin{equation}
\label{eq:LSTM23c}
\frac{\partial \mathbf{c}(t)}{\partial \mathbf{W}_{3}} = diag(\mathbf{f}(t)).\frac{\partial \mathbf{c}(t-1)}{\partial \mathbf{W}_{3}} + \mathbf{b}_{i}(t).\mathbf{r}(t)^T
\end{equation}
For the peephole connections we will have:
\begin{align}
\label{eq:LSTM24c}
&\frac{\partial L_t}{\partial \mathbf{W}_{p3}} = diag(\mathbf{\delta}_{y}^{rec3}(t)).\frac{\partial \mathbf{c}(t)}{\partial \mathbf{W}_{p3}}
\end{align}
where
\begin{equation}
\label{eq:LSTM25c}
\frac{\partial \mathbf{c}(t)}{\partial \mathbf{W}_{p3}} = diag(\mathbf{f}(t)).\frac{\partial \mathbf{c}(t-1)}{\partial \mathbf{W}_{p3}} + \mathbf{b}_{i}(t).\mathbf{c}(t-1)^T
\end{equation}
For bias values, $\mathbf{b}_3$, we will have:
\begin{align}
\label{eq:LSTM26c}
&\frac{\partial L_t}{\partial \mathbf{b}_{3}} = diag(\mathbf{\delta}^{rec3}(t)).\frac{\partial \mathbf{c}(t)}{\partial \mathbf{b}_{3}}
\end{align}
where
\begin{equation}
\label{eq:LSTM27c}
\frac{\partial \mathbf{c}(t)}{\partial \mathbf{b}_{3}} = diag(\mathbf{f}(t)).\frac{\partial \mathbf{c}(t-1)}{\partial \mathbf{b}_{3}} + \mathbf{b}_{i}(t)
\end{equation}
\subsection{Forget Gate}
\label{sec:o_f}
For the recurrent connections we will have:
\begin{align}
\label{eq:LSTM30c}
&\frac{\partial L_t}{\partial \mathbf{W}_{rec2}} = diag(\mathbf{\delta}^{rec2}(t)).\frac{\partial \mathbf{c}(t)}{\partial \mathbf{W}_{rec2}}
\end{align}
where
\begin{align}
\label{eq:LSTM31c}
&\mathbf{\delta}^{rec2}(t) = (1-h(\mathbf{c}(t)))\circ (1+h(\mathbf{c}(t)))\circ \mathbf{o}(t) \circ \mathbf{e}(t)\nn\\
&\frac{\partial \mathbf{c}(t)}{\partial \mathbf{W}_{rec2}} = diag(\mathbf{f}(t)).\frac{\partial \mathbf{c}(t-1)}{\partial \mathbf{W}_{rec2}} + \mathbf{b}_{f}(t).\mathbf{v}(t-1)^T\nn\\
&\mathbf{b}_{f}(t) = \mathbf{c}(t-1)\circ \mathbf{f}(t) \circ (1-\mathbf{f}(t))
\end{align}
For input connections to forget gate we will have:
\begin{align}
\label{eq:LSTM32c}
&\frac{\partial L_t}{\partial \mathbf{W}_{2}} = diag(\mathbf{\delta}^{rec2}(t)).\frac{\partial \mathbf{c}(t)}{\partial \mathbf{W}_{2}}
\end{align}
where
\begin{equation}
\label{eq:LSTM33c}
\frac{\partial \mathbf{c}(t)}{\partial \mathbf{W}_{2}} = diag(\mathbf{f}(t)).\frac{\partial \mathbf{c}(t-1)}{\partial \mathbf{W}_{2}} + \mathbf{b}_{f}(t).\mathbf{r}(t)^T
\end{equation}
For peephole connections we have:
\begin{align}
\label{eq:LSTM34c}
&\frac{\partial L_t}{\partial \mathbf{W}_{p2}} = diag(\mathbf{\delta}^{rec2}(t)).\frac{\partial \mathbf{c}(t)}{\partial \mathbf{W}_{p2}}
\end{align}
where
\begin{equation}
\label{eq:LSTM35c}
\frac{\partial \mathbf{c}(t)}{\partial \mathbf{W}_{p2}} = diag(\mathbf{f}(t)).\frac{\partial \mathbf{c}(t-1)}{\partial \mathbf{W}_{p2}} + \mathbf{b}_{f}(t).\mathbf{c}(t-1)^T
\end{equation}
For forget gate's bias values we will have:
\begin{align}
\label{eq:LSTM36c}
&\frac{\partial L_t}{\partial \mathbf{b}_{2}} = diag(\mathbf{\delta}^{rec2}(t)).\frac{\partial \mathbf{c}(t)}{\partial \mathbf{b}_{2}}
\end{align}
where
\begin{equation}
\label{eq:LSTM37c}
\frac{\partial \mathbf{c}(t)}{\partial \mathbf{b}_{2}} = diag(\mathbf{f}(t)).\frac{\partial \mathbf{c}(t-1)}{\partial \mathbf{b}_{3}} + \mathbf{b}_{f}(t)
\end{equation}

\subsection{Input without Gating ($\mathbf{y}_g(t)$)}
\label{sec:o_yg}
For recurrent connections we will have:
\begin{align}
\label{eq:LSTM40c}
&\frac{\partial L_t	}{\partial \mathbf{W}_{rec4}} = diag(\mathbf{\delta}^{rec4}(t)).\frac{\partial \mathbf{c}(t)}{\partial \mathbf{W}_{rec4}}
\end{align}
where
\begin{align}
\label{eq:LSTM41c}
&\mathbf{\delta}^{rec4}(t) = (1-h(\mathbf{c}(t)))\circ (1+h(\mathbf{c}(t)))\circ \mathbf{o}(t) \circ \mathbf{e}(t)\nn\\
&\frac{\partial \mathbf{c}(t)}{\partial \mathbf{W}_{rec4}} = diag(\mathbf{f}(t)).\frac{\partial \mathbf{c}(t-1)}{\partial \mathbf{W}_{rec4}} + \mathbf{b}_{g}(t).\mathbf{v}(t-1)^T\nn\\
&\mathbf{b}_{g}(t) = \mathbf{i}(t) \circ (1-\mathbf{y}_{g}(t))\circ (1+\mathbf{y}_{g}(t))
\end{align}
For input connections we have:
\begin{align}
\label{eq:LSTM42c}
&\frac{\partial L_t}{\partial \mathbf{W}_{4}} = diag(\mathbf{\delta}^{rec4}(t)).\frac{\partial \mathbf{c}(t)}{\partial \mathbf{W}_{4}}
\end{align}
where
\begin{equation}
\label{eq:LSTM43c}
\frac{\partial \mathbf{c}(t)}{\partial \mathbf{W}_{4}} = diag(\mathbf{f}(t)).\frac{\partial \mathbf{c}(t-1)}{\partial \mathbf{W}_{4}} + \mathbf{b}_{g}(t).\mathbf{r}(t)^T
\end{equation}
For bias values we will have:
\begin{align}
\label{eq:LSTM44c}
&\frac{\partial L_t}{\partial \mathbf{b}_{4}} = diag(\mathbf{\delta}^{rec4}(t)).\frac{\partial \mathbf{c}(t)}{\partial \mathbf{b}_{4}}
\end{align}
where
\begin{equation}
\label{eq:LSTM45c}
\frac{\partial \mathbf{c}(t)}{\partial \mathbf{b}_{4}} = diag(\mathbf{f}(t)).\frac{\partial \mathbf{c}(t-1)}{\partial \mathbf{b}_{4}} + \mathbf{b}_{g}(t)
\end{equation}

\subsection{Error signal backpropagation}
\label{sec:errBPP}
Error signals are back propagated through time using following equations:
\begin{align}
\label{eq:LSTM46}
&\mathbf{\delta}^{rec1}(t-1) = [\mathbf{o}(t-1)\circ (1-\mathbf{o}(t-1))\circ h(\mathbf{c}(t-1))]\nn\\
&\circ [\mathbf{W}_{rec1}^T.\mathbf{\delta}^{rec1}(t) + \mathbf{e}(t-1)]
\end{align}
\begin{align}
\label{eq:LSTM47}
&\mathbf{\delta}^{rec_i}(t-1) = [(1-h(\mathbf{c}(t-1)))\circ (1+h(\mathbf{c}(t-1)))\nn\\
&\circ \mathbf{o}(t-1)]\circ [ \mathbf{W}_{rec_i}^T.\mathbf{\delta}^{rec_i}(t) + \mathbf{e}(t-1)],\nn\\
&\;\;\;\;for \;\;\; i\in \{2,3,4\}
\end{align}

\ifCLASSOPTIONcaptionsoff
  \newpage
\fi

\bibliographystyle{IEEEtran}
\bibliography{refs}

\begin{IEEEbiography}[{\includegraphics[width=1in,height=1.25in,clip,keepaspectratio]{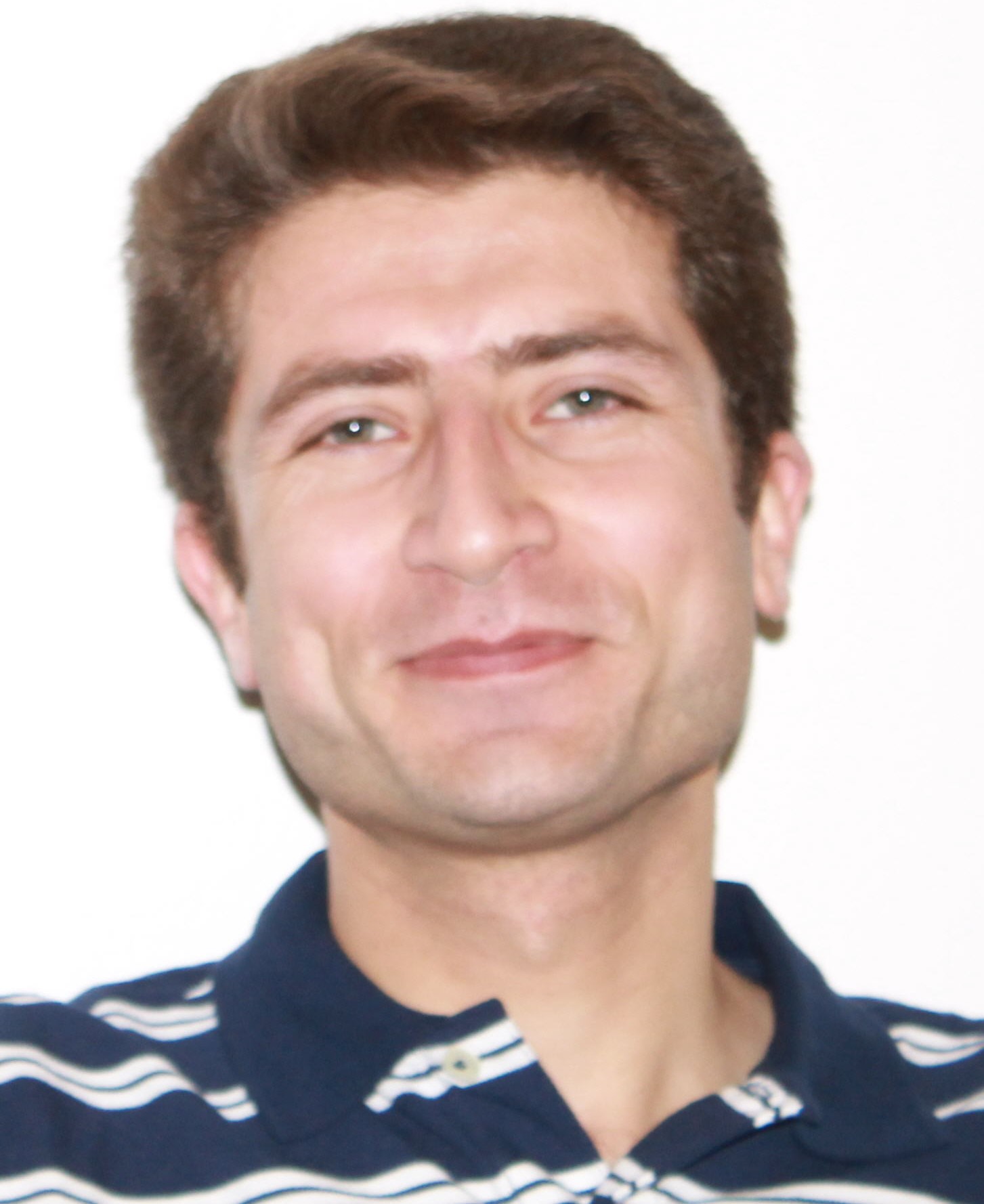}}]{Hamid Palangi}

(S'12) is a Ph.D. Candidate in the Electrical and Computer Engineering Department at the University of British Columbia (UBC), Canada. Before joining UBC in Jan. 2012, he received his M.Sc. degree in 2010 from Sharif University of Technology, Iran and B.Sc. degree in 2007 from Shahid Rajaee University, Iran, both in Electrical Engineering. Since Jan. 2012, he has been a member of Image and Signal Processing Lab at UBC. His main research interests are Machine Learning, Deep Learning and Neural Networks, Linear Inverse Problems and Compressive Sensing, with applications in Natural Language and Image data.

\end{IEEEbiography}

\begin{IEEEbiography}[{\includegraphics[width=1in,height=1.25in,clip,keepaspectratio]{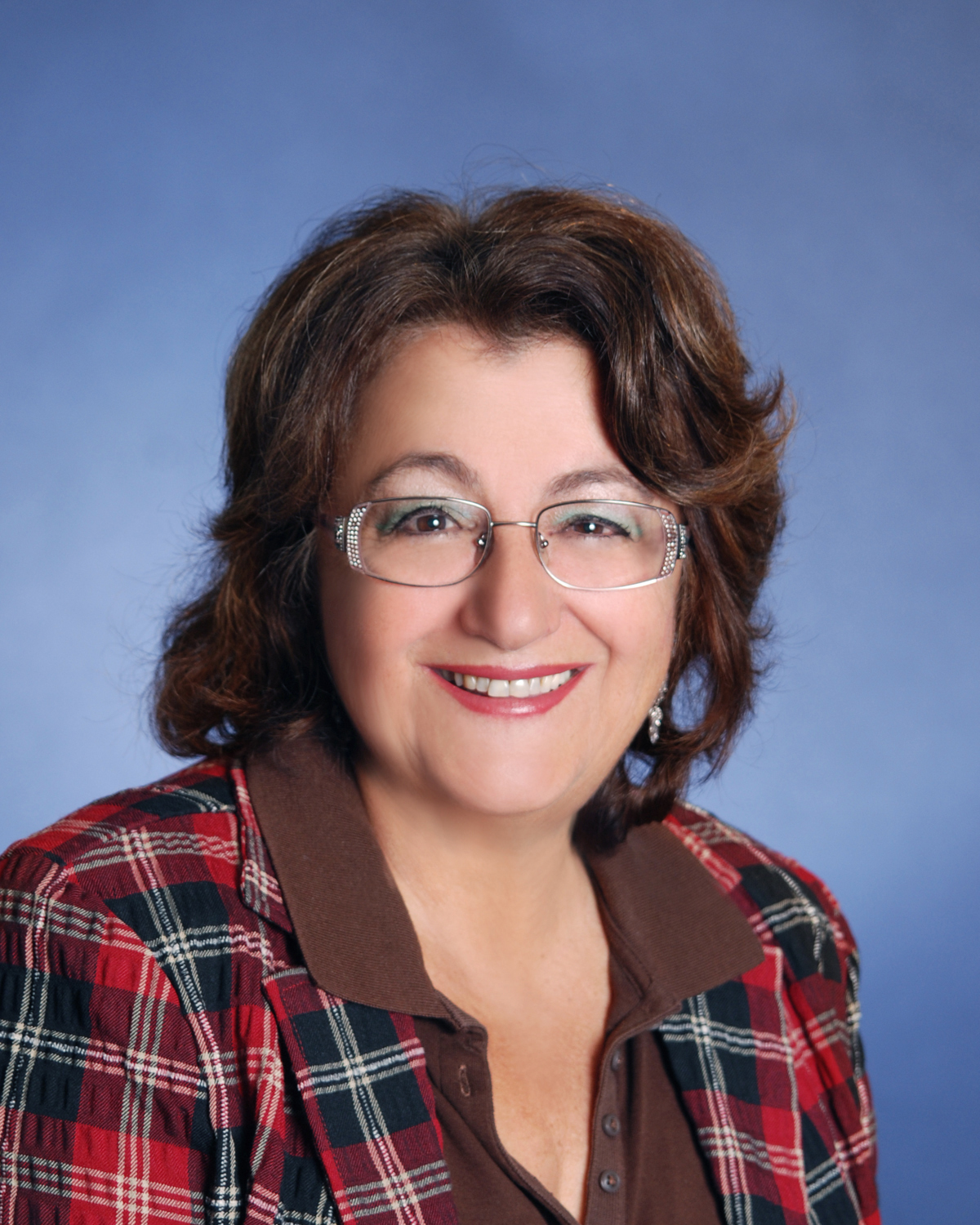}}]{Rabab Ward}

is a Professor  Emeritus in the Electrical and Computer Engineering Department at the University of British Columbia (UBC), Canada.. Her research interests are mainly in the areas of signal, image and video processing. She has made contributions in the areas of signal detection, image encoding, image recognition, restoration and enhancement, and their applications to multimedia and medical imaging, face recognition , infant cry signals and brain computer interfaces. She has published around 500 refereed journal and conference papers and holds six patents related to cable television, picture monitoring, measurement and noise reduction.
She is a Fellow of  the Royal Society of Canada , the IEEE , the Canadian Academy of Engineers and the Engineering Institute of Canada.
She has received many  top awards such as  the  "Society Award” of  the IEEE Signal Processing Society, the Career Achievement Award of CUFA BC,The Paradigm Shifter Award from The Society for Canadian Women in Science and Technology and British Columbia's APEGBC top engineering award "The RA McLachlan Memorial Award" and UBC Killam Research Prize and Killam Senior Mentoring Award.                                                                                                  
She is presently  the President of the IEEE Signal Processing Society. She was the General Chair of IEEE ICIP 2000 and Co-Chair of  IEEE ICASSP 2013.

\end{IEEEbiography}

\begin{IEEEbiography}[{\includegraphics[width=1in,height=1.25in,clip,keepaspectratio]{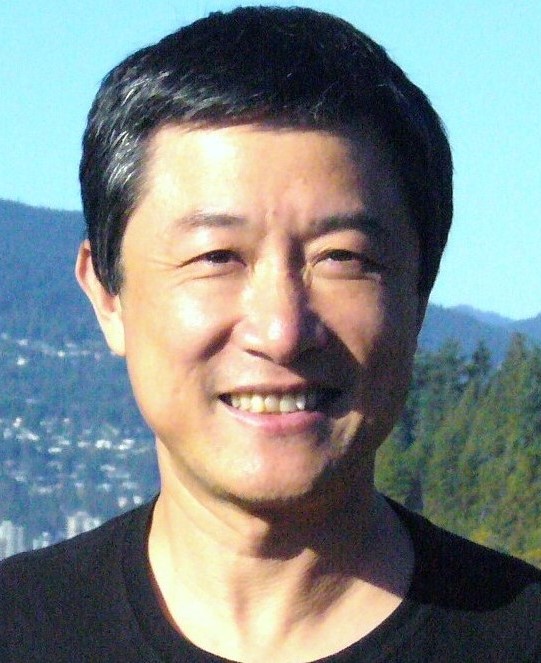}}]{Li Deng}

received a Ph.D. from the University of Wisconsin-Madison. He was an assistant and then tenured full professor at the University of Waterloo, Ontario, Canada during 1989-1999. Immediately afterward he joined Microsoft Research, Redmond, USA as a Principal Researcher, where he currently directs the R \& D of its Deep Learning Technology Center he founded in early 2014. Dr. Deng’s current activities are centered on business-critical applications involving big data analytics, natural language text, semantic modeling, speech, image, and multimodal signals. Outside his main responsibilities, Dr. Deng’s research interests lie in solving fundamental problems of machine learning, artificial and human intelligence, cognitive and neural computation with their biological connections, and multimodal signal/information processing. In addition to over 70 granted patents and over 300 scientific publications in leading journals and conferences, Dr. Deng has authored or co-authored 5 books including 2 latest books: Deep Learning: Methods and Applications (NOW Publishers, 2014) and Automatic Speech Recognition: A Deep-Learning Approach (Springer, 2015), both with English and Chinese editions. Dr. Deng is a Fellow of the IEEE, the Acoustical Society of America, and the ISCA. He served on the Board of Governors of the IEEE Signal Processing Society. More recently, he was the Editor-In-Chief for the IEEE Signal Processing Magazine and for the IEEE/ACM Transactions on Audio, Speech, and Language Processing; he also served as a general chair of ICASSP and area chair of NIPS. Dr. Deng’s technical work in industry-scale deep learning and AI has impacted various areas of information processing, especially Microsoft speech products and text- and big-data related products/services. His work helped initiate the resurgence of (deep) neural networks in the modern big-data, big-compute era, and has been recognized by several awards, including the 2013 IEEE SPS Best Paper Award and the 2015 IEEE SPS Technical Achievement Award “for outstanding contributions to deep learning and to automatic speech recognition.”

\end{IEEEbiography}

\end{document}